 \documentclass[tablecaption=bottom,wcp]{jmlr} 

\usepackage{booktabs}
\usepackage[load-configurations=version-1]{siunitx} 

\newcommand{\pzk}{p(x, z_{\phi, k})}

\newcommand{\qzk}{q_\phi(z_{\phi, k})}

\newcommand{\qzkstl}{q_\psi(z_{\phi, k})}

\theorembodyfont{\upshape}
\theoremheaderfont{\scshape}
\theorempostheader{:}
\theoremsep{\newline}

\usepackage{nicefrac}

\jmlrproceedings{AABI 2020}{3rd Symposium on Advances in Approximate Bayesian Inference, 2020}

\title[Evaluation of Biased Alpha Divergence Minimization]{Empirical Evaluation of Biased Methods for Alpha Divergence Minimization}

\author{\Name{Tomas Geffner} \Email{tgeffner@cs.umass.edu}\and
 \Name{Justin Domke} \Email{domke@cs.umass.edu}\\
 \addr University of Massachusetts, Amherst}

\begin{document}

\maketitle



\vspace{-1cm}

\section{Introduction}

Traditional variational inference (VI) minimizes the ``exclusive'' KL divergence $\mathrm{KL}(q\Vert p)$ between the approximating distribution $q$ and the target $p$. There has been great recent interest in methods to minimize other alpha-divergences, such as the ``inclusive'' KL divergence, $\mathrm{KL}(p\Vert q)$. Some methods employ unbiased gradient estimators \citep{dieng2017chivi, kuleshov2017neural}. These estimators often suffer from a high variance, difficulting optimization \citep{geffner2020difficulty}. Another class of methods estimate a gradient using self-normalized importance sampling \citep{rws, OnIWAE, renyiVI}. While these estimators may control variance, they do so at the cost of some bias. While some positive results have been observed for biased methods (e.g. higher log-likelihoods \citep{renyiVI, dieng2017chivi}), the magnitude of the bias and the effect it has on the distributions they return are not well understood.

In this paper we empirically evaluate biased methods for alpha-divergence minimization.
In particular, we focus on how the bias affects the solutions found, and how this depends on the dimensionality of the problem. Our two main takeaways are (i) solutions returned by these methods appear to be strongly biased towards minimizers of the traditional ``exclusive'' KL-divergence, $\mathrm{KL}(q\Vert p)$. And (ii) in high dimensions, an impractically large amount of computation is needed to mitigate this bias and obtain solutions that actually minimize the alpha-divergence of interest.

Finally, we relate these results to the curse of dimensionality. In high dimensions, it is well known that self-normalized importance sampling often suffers from ``weight degeneracy'' (unless the number of samples used is exponential in the dimensionality of the problem \citep{bugallo2017adaptive, bengtsson2008curse}), resulting in estimates with high bias. We empirically show that weight degeneracy does indeed occur with these estimators in cases where they return highly biased solutions.

\subsection{Estimators considered} \label{sec:estimators}

\textbf{Notation:} $q_\phi$ denotes the variational distribution parameterized by $\phi$. $z_{\phi, k}$ denotes a sample from $q_\phi$ obtained via reparameterization \citep{vaes_welling, doublystochastic_titsias}. $\psi$ denotes the parameters $\phi$ ``protected under differentiation'' (i.e. $\psi = \texttt{stop\_gradient}(\phi)$).

\begin{itemize}
    \item For the Renyi alpha-divergence, $\mathrm{R}_\alpha(q||p)$, \citet{renyiVI} proposed the estimator
    \[ g_{R_\alpha} = -\sum_{k=1}^K \frac{w_{\alpha, k}}{\sum_{j=1}^K w_{\alpha, j}} \, \nabla_\phi \log \frac{\pzk}{\qzk}, \qquad \mbox{where} \qquad w_{\alpha, k} = \left( \frac{\pzk}{\qzk} \right)^{1 - \alpha}. \]
    
    This is defined for $\alpha>0$. We use $\alpha=0.5$ in our experiments.
    
    \item For the ``inclusive'' divergence $\mathrm{KL}(p||q)$, the reweighted wake-sleep estimator \citep{rws} (also used in Edward \citep{edward2016}) is given by
    \[ g_\mathrm{rws} = \sum_{k=1}^K \frac{w_k}{\sum_{j=1}^K w_j} \, \nabla_\phi \log q_\phi(z_{\psi, k}), \qquad \mbox{where} \qquad w_k = \frac{p(x, z_{\phi, k})}{q_\phi(z_{\phi, k})}.\]
    
    For the same divergence, the ``sticking the landing'' estimator \citep{stickingthelanding} is given by\footnote{This estimator was originally proposed as an estimator for importance weighted variational inference \citep{IWVAE}. \citet{OnIWAE} introduced the view of it being a self-normalized importance sampling estimator for the gradient of $\mathrm{KL}(p||q)$.}
    
    \[ g_\mathrm{stl} = -\sum_{k=1}^K \frac{w_k}{\sum_{j=1}^K w_j} \, \nabla_\phi \log \frac{\pzk}{\qzkstl}, \qquad \mbox{where} \qquad w_k = \frac{p(x, z_{\phi, k})}{q_\phi(z_{\phi, k})}.\]
    
    \item For the chi divergence, $\chi^2(p||q)$, the CHIVI algorithm \citep{dieng2017chivi} uses the estimator 
    \[ g_\mathrm{chivi} = -\sum_{k=1}^K \left(\frac{w_k}{\max_j w_j}\right)^2  \nabla_\phi \log \frac{\pzk}{\qzk}, \qquad \mbox{where} \qquad w_k = \frac{p(x, z_{\phi, k})}{q_\phi(z_{\phi, k})}. \]
    
    (This estimator was used by \citet{dieng2017chivi} in their experiments, but not in their analysis.) For the same divergence, the doubly reparameterized estimator\footnote{It is known that importance weighted VI is equivalent to minimizing the $\chi^2$ divergence in the limit \citep{maddison2017filtering, domke2018importance}. The doubly reparameterized estimator for importance weighted VI was introduced by \citet{doublyrep}, and \citet{OnIWAE} introduced the view of it being a self-normalized importance sampling estimator for the gradient of $\chi^2(p||q)$.} \citep{doublyrep, OnIWAE} is given by
    \[ g_\mathrm{drep} = -\sum_{k=1}^K \left(\frac{w_k}{\sum_{j=1}^K w_j}\right)^2 \nabla_\phi \log \frac{\pzk}{\qzkstl}, \qquad \mbox{where} \qquad w_k = \frac{p(x, z_{\phi, k})}{q_\phi(z_{\phi, k})}. \]
\end{itemize}

All of these estimators are asymptotically unbiased in the limit of $K\rightarrow \infty$ except for $g_\mathrm{chivi}$. However, the bias for finite $K$ is not well understood.

\section{Empirical Evaluation}

We now present an empirical evaluation of the estimators described above. We consider two scenarios for the model $p$:  a simple Gaussian distribution and logistic regression. In both cases we use Adam \citep{adam} with each of the gradient estimators to minimize the corresponding alpha-divergence, and compare the results obtained against the theoretically optimal ones.

\subsection{Evaluation I: Gaussian Model} \label{sec:gaussian}


\noindent\textbf{Model:} Similarly to \citet{neal2011mcmc}, we set the target $p$ to be a diagonal $d$-dimensional Gaussian with mean zero and variances $\sigma_{p_i}^2 = 0.2 + 9.8 \frac{i}{d}$. So, the variance of the components of $p$ grows linearly from $\sigma_{p_1}^2=0.2$ to $\sigma_{p_d}^2=10$. We ran simulations for dimensionalities $d \in \{10, 100, 1000\}$\vspace{0.1cm}

\noindent\textbf{Variational distribution:} We set $q$ to be a mean-zero isotropic Gaussian with covariance $\sigma_q^2 I$. So, $q$ has a single parameter $\sigma_q$, which we initialize to $\sigma_q^2 = 9$.\vspace{0.1cm}

\noindent\textbf{Optimization details:} We attempt to optimize alpha-divergences by running Adam (step-size $\eta = 0.01$) for $2000$ steps using each of the gradient estimators introduced in Section \ref{sec:estimators}. We repeat this for estimators obtained using $K$ samples, with $K \in \{10, 100, 1000\}$.\vspace{0.1cm}


\noindent\textbf{Baselines:} In this scenario we can compute the optimal $\sigma_q$ to exactly minimize each of $\mathrm{KL}(q\Vert p)$, $\mathrm{KL}(p\Vert q)$, $R_\alpha(q\Vert p)$ and $\chi^2(p\Vert q)$. This gives us a clear way of visualizing the bias induced by each estimator.\vspace{0.1cm}

\noindent\textbf{Results:} Fig.~\ref{fig:target_method_gauss} shows how the parameter $\sigma_q$ evolves as optimization proceeds when using the ``sticking the landing'' estimator $g_{stl}$, which targets the divergence $\mathrm{KL}(p\Vert q)$. For low dimensions ($d = 10$), the optimal value $\sigma_q^*$ is recovered almost exactly as long as $K \geq 100$ samples are used to estimate the gradients. For higher dimensions, the solution is increasingly biased towards the minimizer of $\mathrm{KL}(q||p)$. While this bias can in theory be mitigated by increasing the number of samples $K$ used to estimate the gradients, the number required becomes impractically large in high dimensions.

\begin{figure}[htbp]
\floatconts
  {fig:target_method_gauss}
  {\caption{\textbf{For high dimensions an impractically large number of samples $K$ is needed to mitigate the estimator's bias.} Optimization results when minimizing $\mathrm{KL}(p||q)$ for the synthetic Gaussian model using the biased gradient estimator $g_{stl}$ obtained using $K$ samples.}}
  {\includegraphics[scale=0.3, trim = {0 0 0 0}, clip]{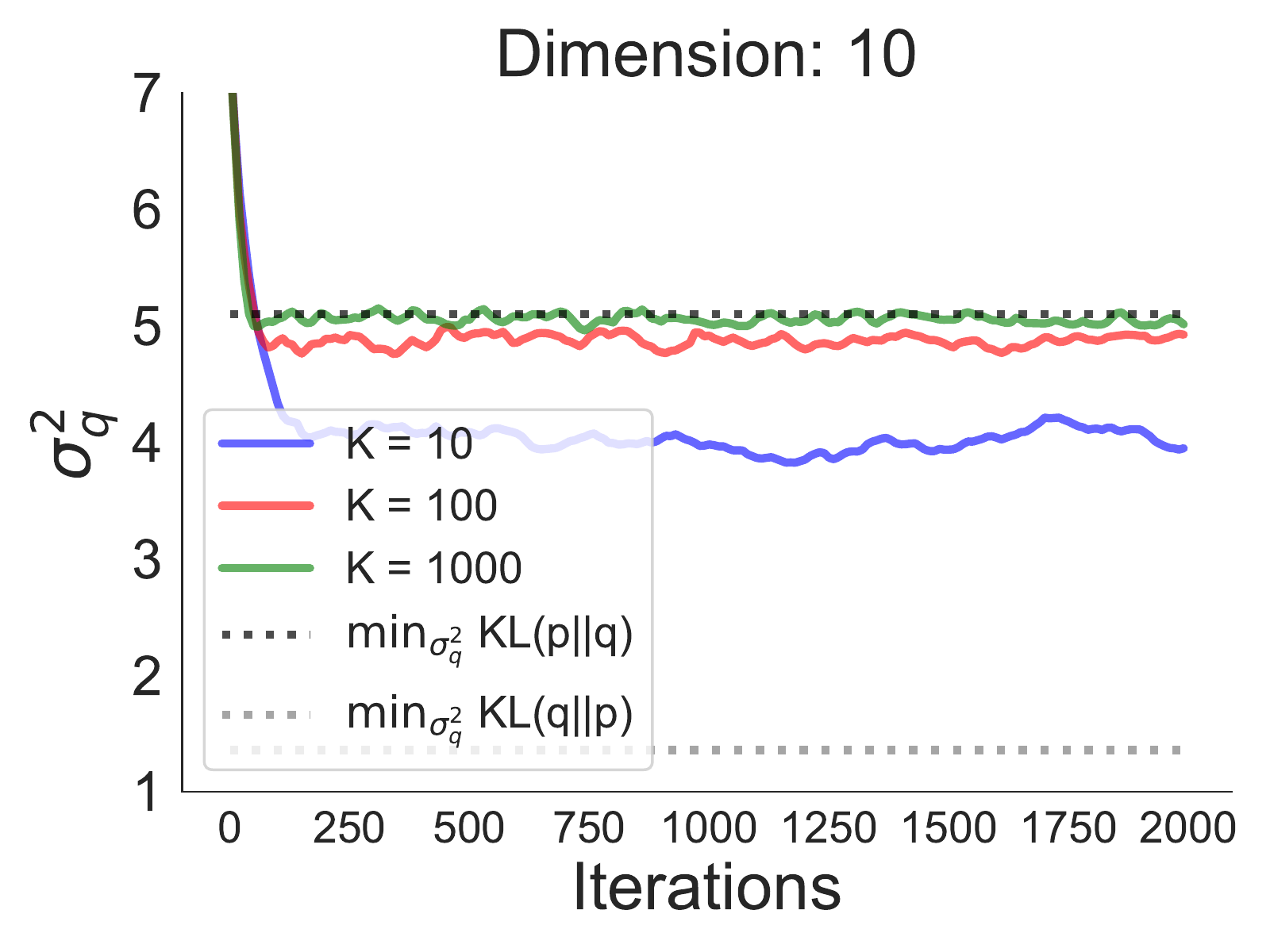}\hspace{0.5cm}
  \includegraphics[scale=0.3, trim = {2.1cm 0 0 0}, clip]{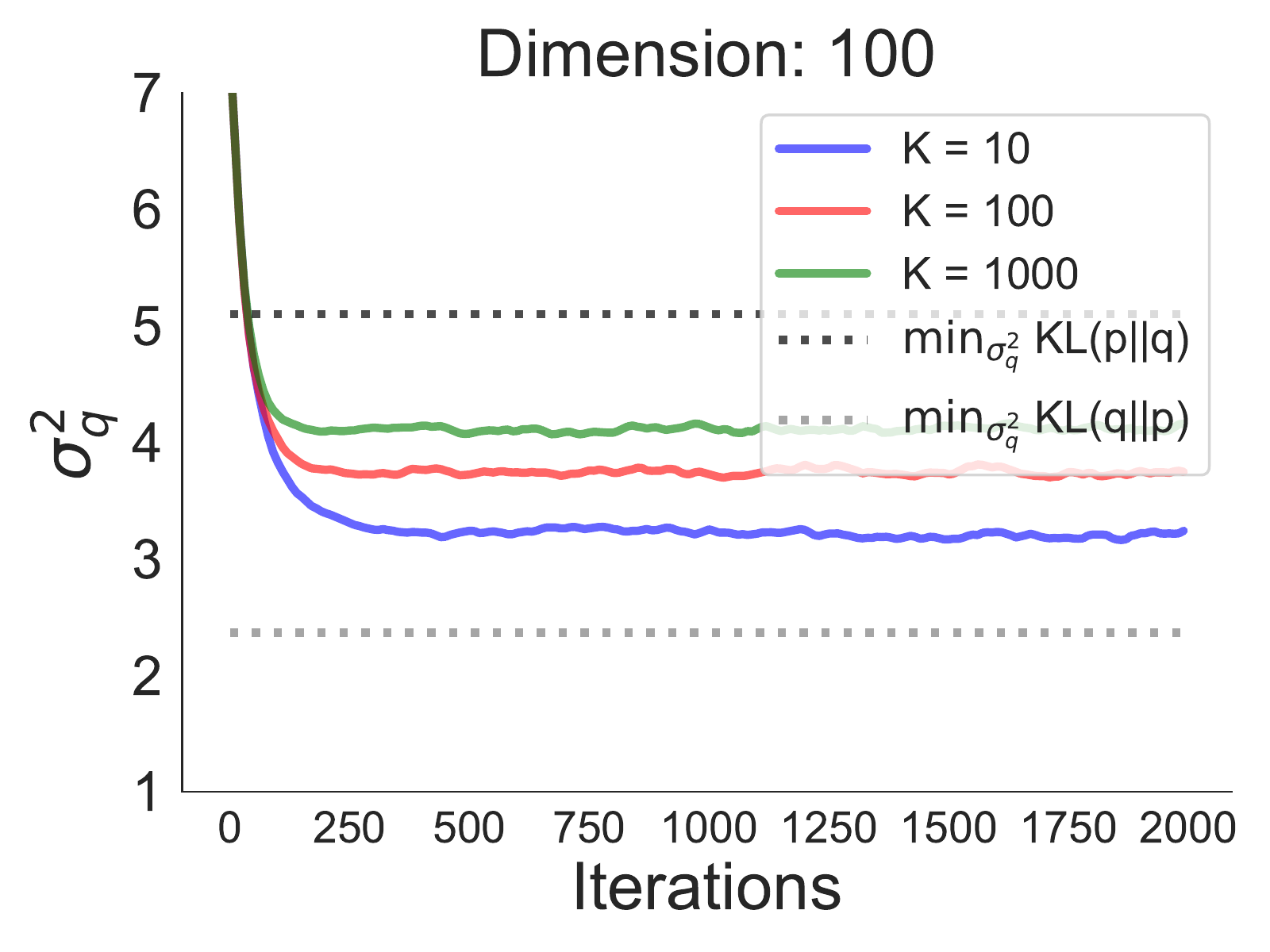}\hspace{0.5cm}
  \includegraphics[scale=0.3, trim = {2.1cm 0 0 0}, clip]{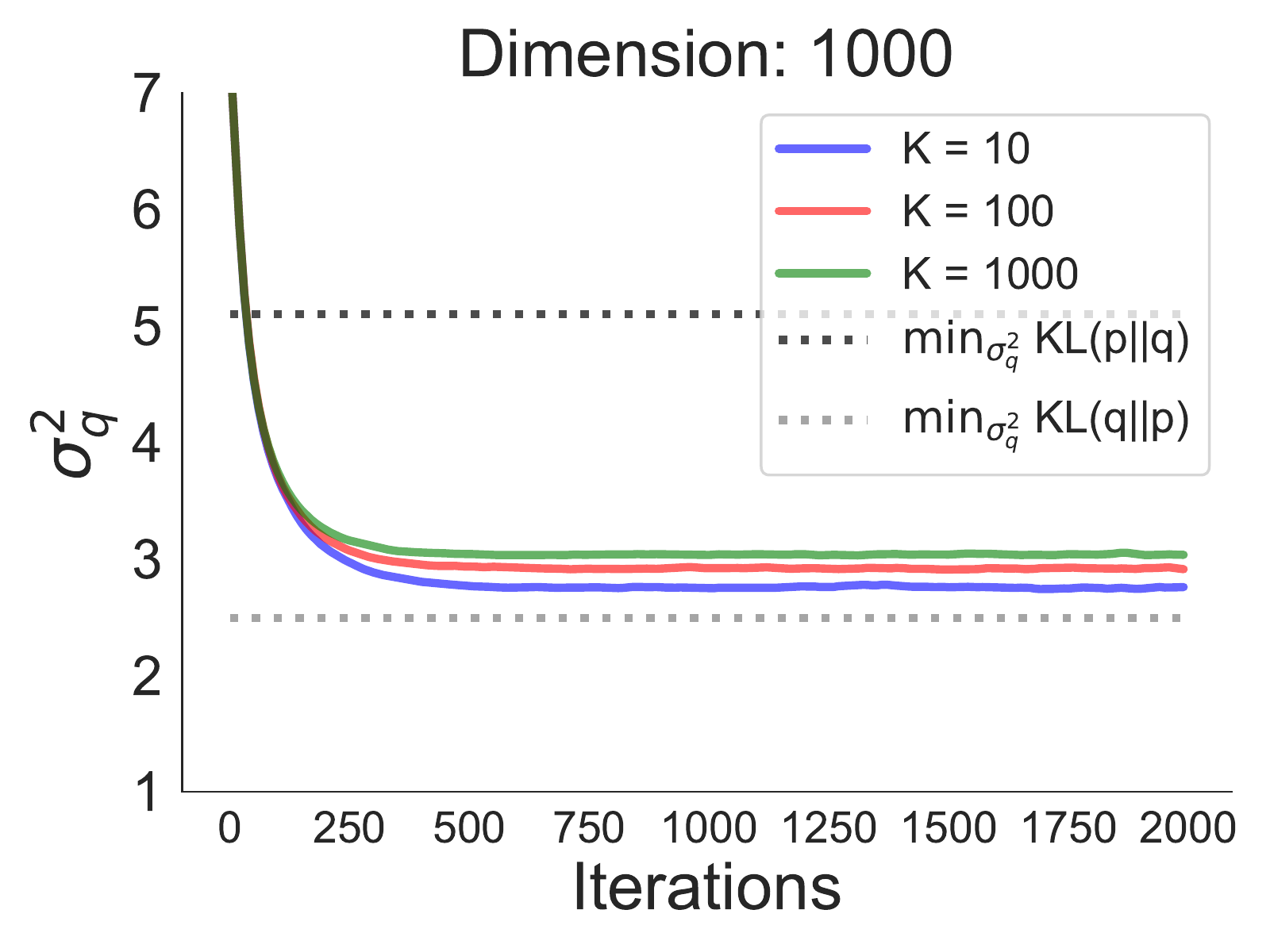}
  }
\end{figure}

Fig.~\ref{fig:var_dim_gauss} shows that a similar phenomena occurs with all other estimators introduced in Section \ref{sec:estimators}. The plots do not show optimization traces; they show the final $\sigma_q^2$ after 2000 optimization steps as a function of the problem's dimension. (We show raw optimization results for all estimators in Appendix \ref{apd:1}). The same conclusion as the one described above applies for all estimators (except $\g_\mathrm{chivi}$): The methods tend to work well in low dimensions, but return suboptimal solutions that are strongly biased towards minimizers of $\mathrm{KL}(q||p)$ in higher dimensions. Again, while this bias can be mitigated by increasing the number of samples $K$ used to estimate gradients, the value of $K$ required becomes impractically large in high dimensions. ($\g_\mathrm{chivi}$ also yields suboptimal solutions in low dimensions. This is likely because this estimator uses atypical weight normalization and so is not asymptotically unbiased.)

\begin{figure}[htbp]
\floatconts
  {fig:var_dim_gauss}
  {\caption{\textbf{In high dimensions solutions are strongly biased towards minimizers of $\mathrm{KL}(q||p)$.} Optimization results for all estimators for the synthetic Gaussian model, as a function of the dimensionality of the problem and the number of samples $K$ used to estimate gradients.}}
  {\includegraphics[scale=0.3, trim = {0 0 0 0}, clip]{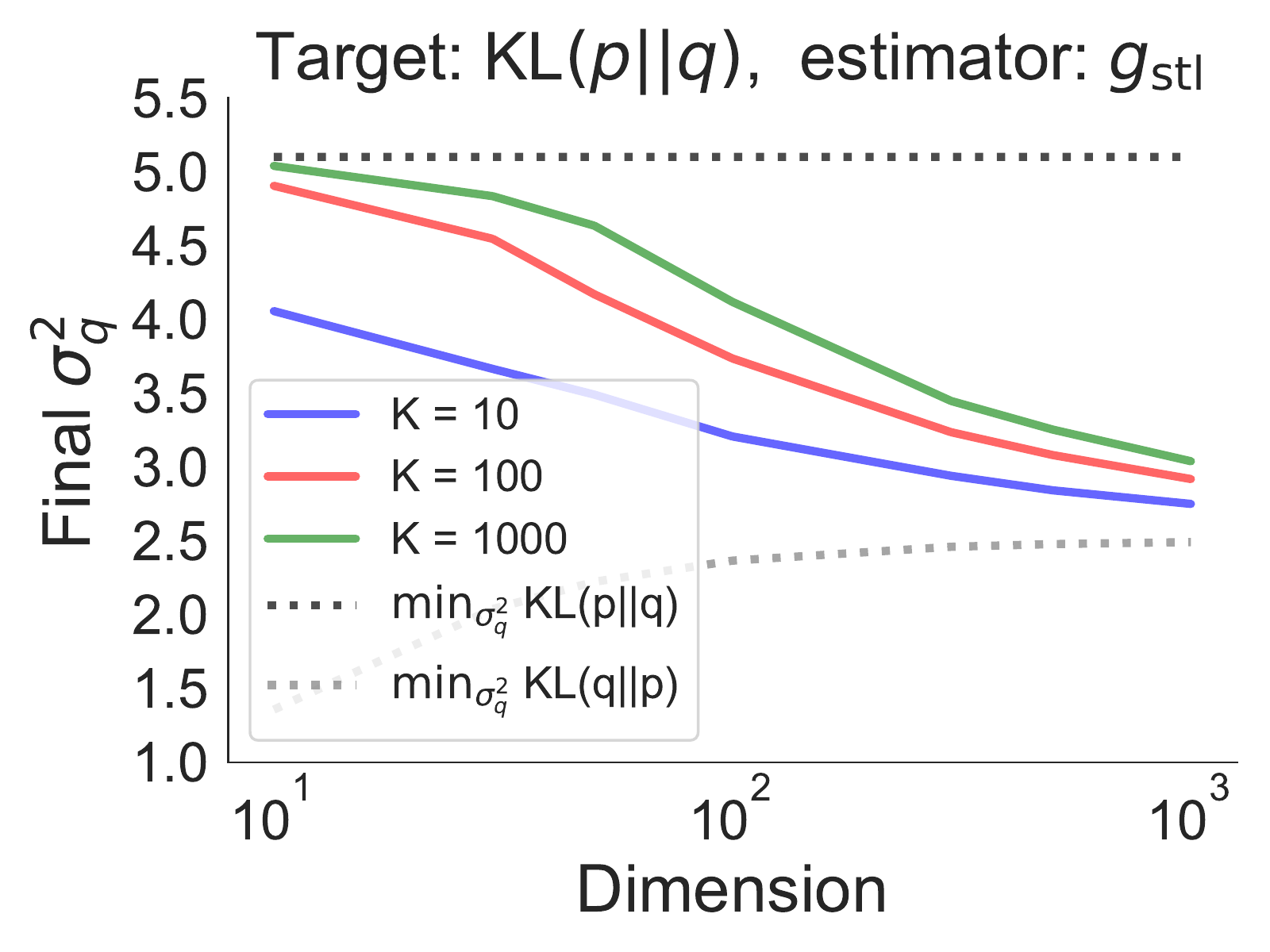}\hspace{0.5cm}
  \includegraphics[scale=0.3, trim = {2.7cm 0 0 0}, clip]{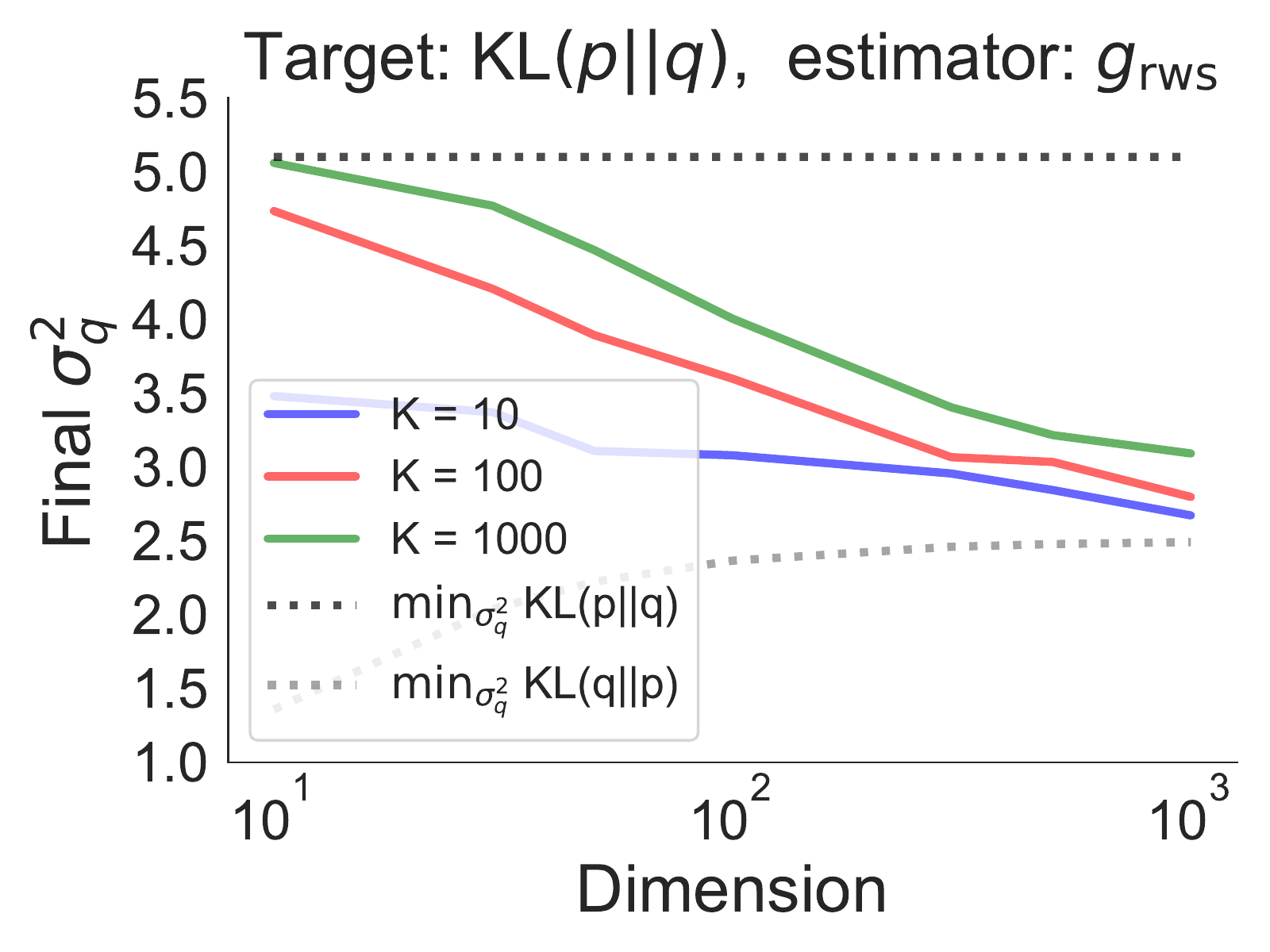}\hspace{0.5cm}
  \includegraphics[scale=0.3, trim = {2.7cm 0 0 0}, clip]{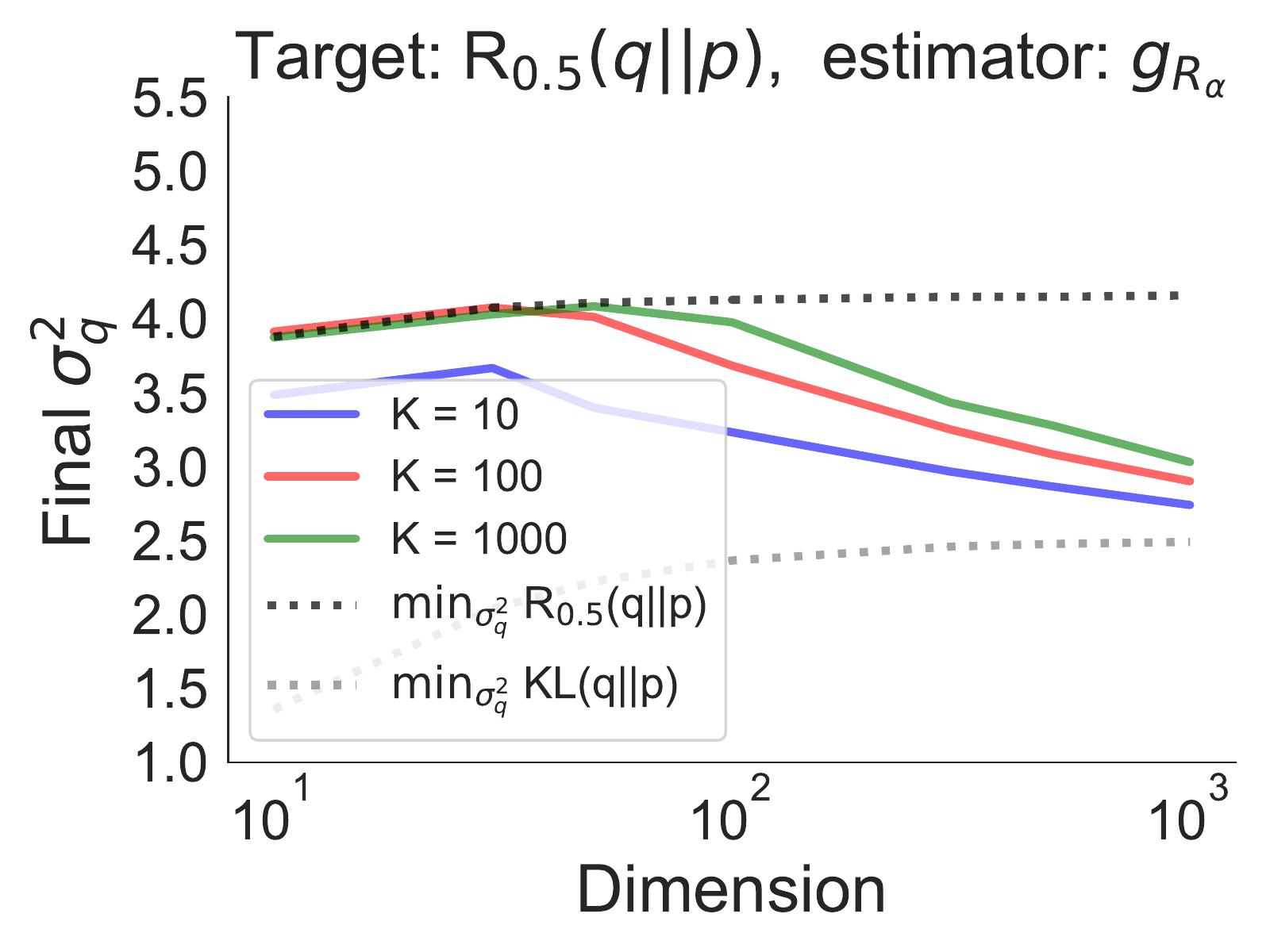}\vspace{0.5cm}
  
  \includegraphics[scale=0.3, trim = {0 0 0 0}, clip]{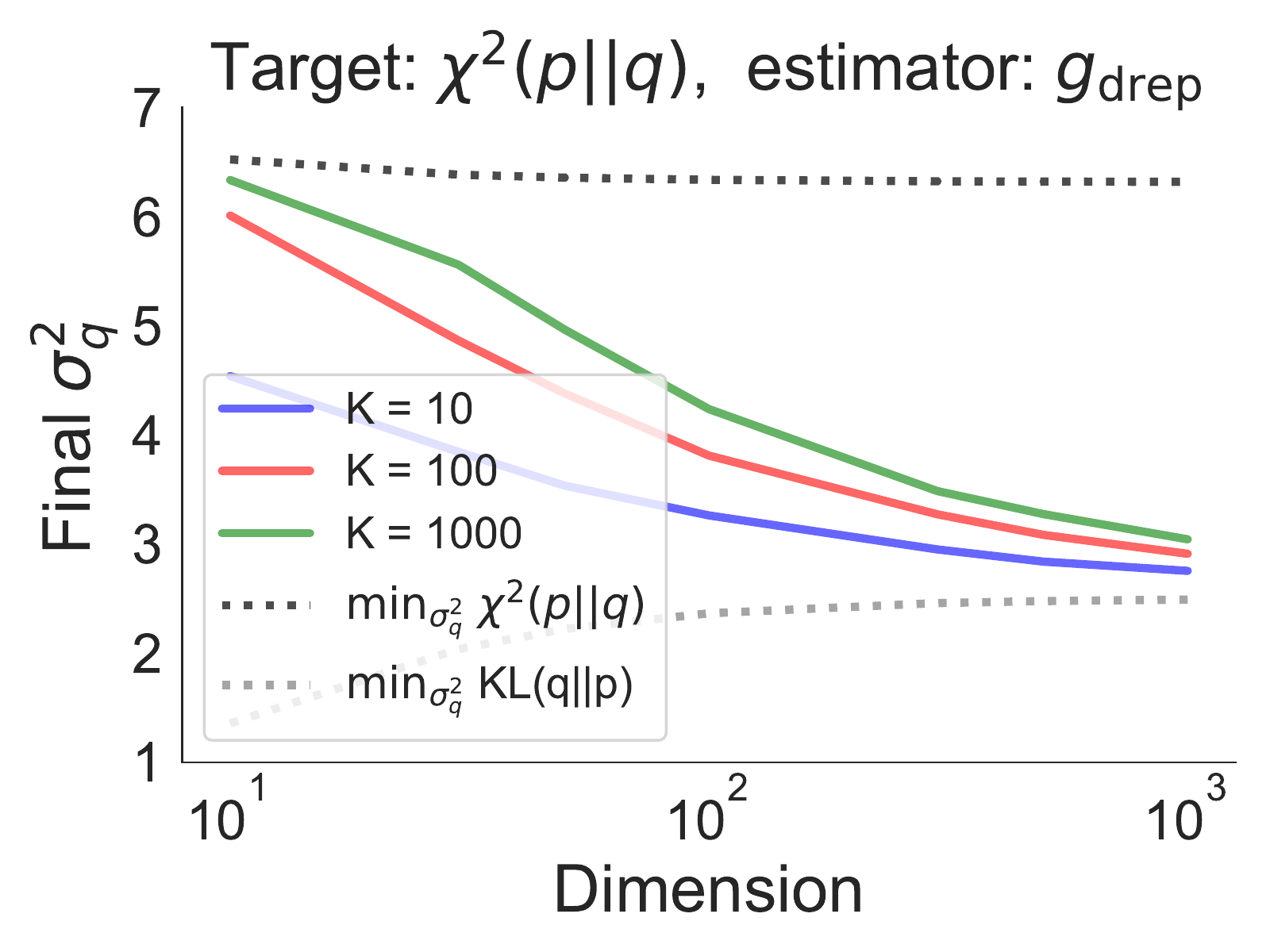}\hspace{0.5cm}
  \includegraphics[scale=0.3, trim = {2.2cm 0 0 0}, clip]{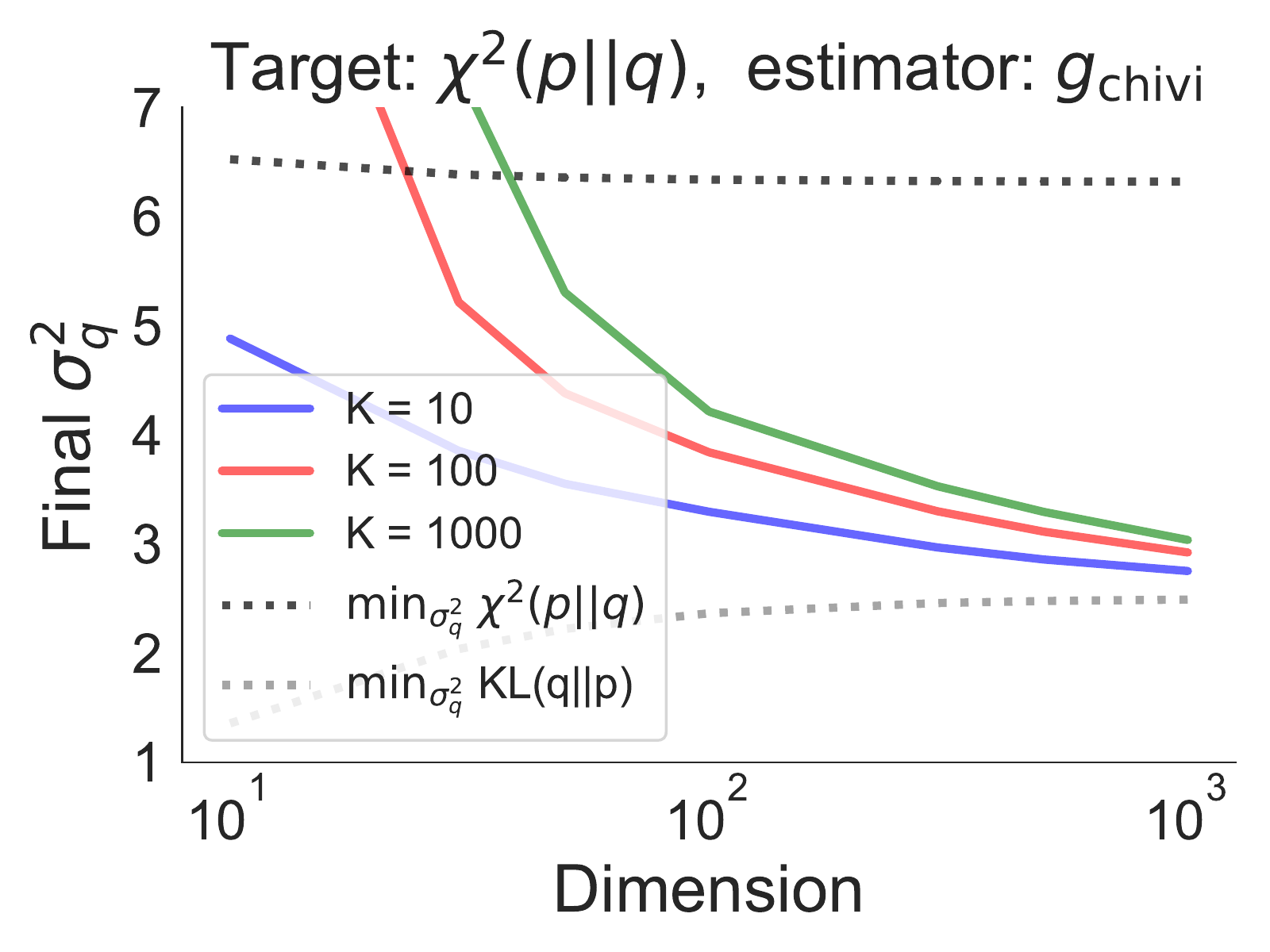}
  }
\end{figure}

We believe that the suboptimality of the solutions returned by biased methods in high dimensions is related to the weight collapse effect (also known as weight degeneracy) suffered by self normalized importance sampling \citep{bengtsson2008curse}. To verify this empirically, we plot the magnitude of the normalized importance weights obtained for different dimensionalities $d$ and number of samples $K$. We observe that the pairs $(d, K)$ for which solutions are highly biased correspond to the cases for which the weight collapse effect is observed (details in Appendix \ref{app:3} and Fig.~\ref{fig:wc} therein). 

\subsection{Evaluation II: Logistic Regression}

\noindent\textbf{Model:} Bayesian logistic regression with two datasets: \textit{sonar} ($d= 61$) and \textit{a1a} ($d = 120$).\vspace{0.1cm}

\noindent\textbf{Variational distribution:} We set $q$ to be a diagonal Gaussian, with mean $\mu_q$ and variance $\sigma_q^2$ (vectors of dimension $d$), with components initialized to $\mu_{q_i} = 0$ and $\sigma_{q_i}^2 = 9$. (We parameterize the variance using the log-scale parameters.)\vspace{0.1cm}

\noindent\textbf{Optimization details:} We attempt to optimize alpha-divergences by running Adam (step-size $\eta = 0.01$) for $5000$ steps using each of the gradient estimators introduced in Section \ref{sec:estimators}. We repeat this for estimators obtained using $K$ samples, with $K \in \{10, 100, 1000\}$.\vspace{0.1cm}

\noindent\textbf{Baselines:} We compare against the optimal parameters $(\mu_q^*, \sigma_q^*)$ that minimize $\mathrm{KL}(p||q)$. While these cannot be computed in closed form, we approximate them by minimizing $\mathrm{KL}(p||q)$ using the algorithm proposed by \citet{MarkovianSC}\footnote{The algorithm's main idea involves minimizing $\mathrm{KL}(p||q)$ using samples from $p$ obtained via MCMC. In our case we use Stan \citep{carpenter2017stan} to get reliable samples, making sure to run multiple chains and checking several convergence criteria, such as the value of $\hat R$.}. Again, having these parameters provides a clear way of visualizing the effect of using biased gradient estimates.\vspace{0.1cm}

\noindent\textbf{Results:} Fig.~\ref{fig:opt_log_reg_drep} shows optimization results for the estimator $g_\mathrm{stl}$, which targets $\mathrm{KL}(p\Vert q)$. It can be observed that, for the \textit{sonar} dataset ($d = 61$), distributions that attain near-optimal performance are obtained using gradient estimates computed with $K \geq 100$ samples. In contrast, for the \textit{a1a} dataset ($d = 120$), all values of $K$ tested lead to significantly biased and suboptimal solutions. (Though, as expected, increasing the number of samples $K$ reduces the suboptimality gap.)

\begin{figure}[htbp]
\floatconts
  {fig:opt_log_reg_drep}
  {\caption{\textbf{For the dataset with higher dimensionality the algorithm returns biased solutions that are suboptimal regardless of the number of samples $K$ used.} The plots show optimization results for minimizing $\mathrm{KL}(p\Vert q)$ for the logistic regression model using the estimator $g_{stl}$ obtained with $K$ samples. The y-axis in the plots show the true $\mathrm{KL}(p\Vert q)$ (up to the additive constant $c = \log p(x)$ -- which can be estimated using samples from $p(z|x)$, obtained using Stan \citep{carpenter2017stan}). The seemingly strange behavior of optimization traces is not due to bad optimization hyperparameters, but to the bias of the gradient estimator.} }
  {
  \includegraphics[scale=0.35, trim = {0 0 0 0}, clip]{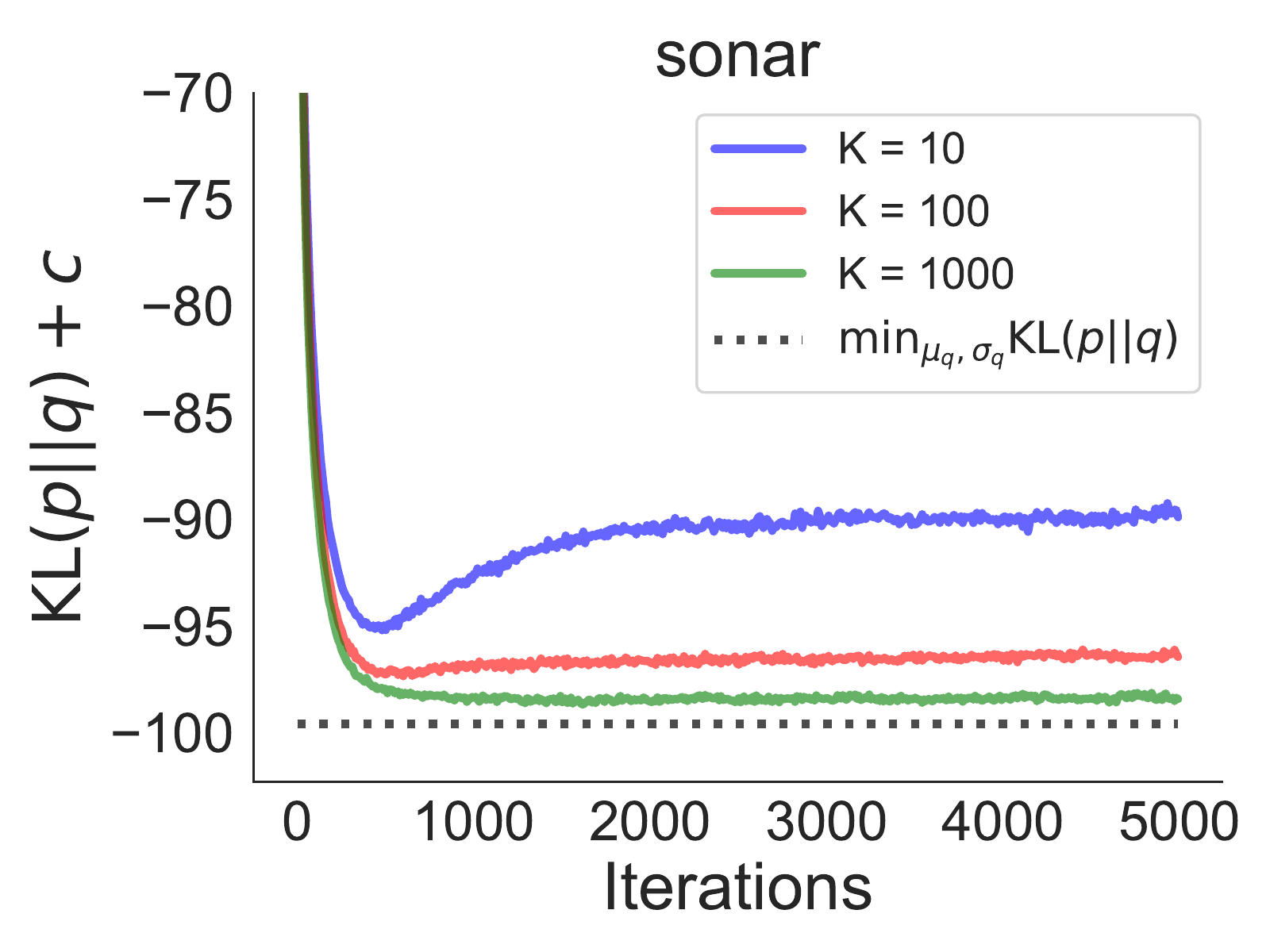}\hspace{0.5cm}
  \includegraphics[scale=0.35, trim = {1.5cm 0 0 0}, clip]{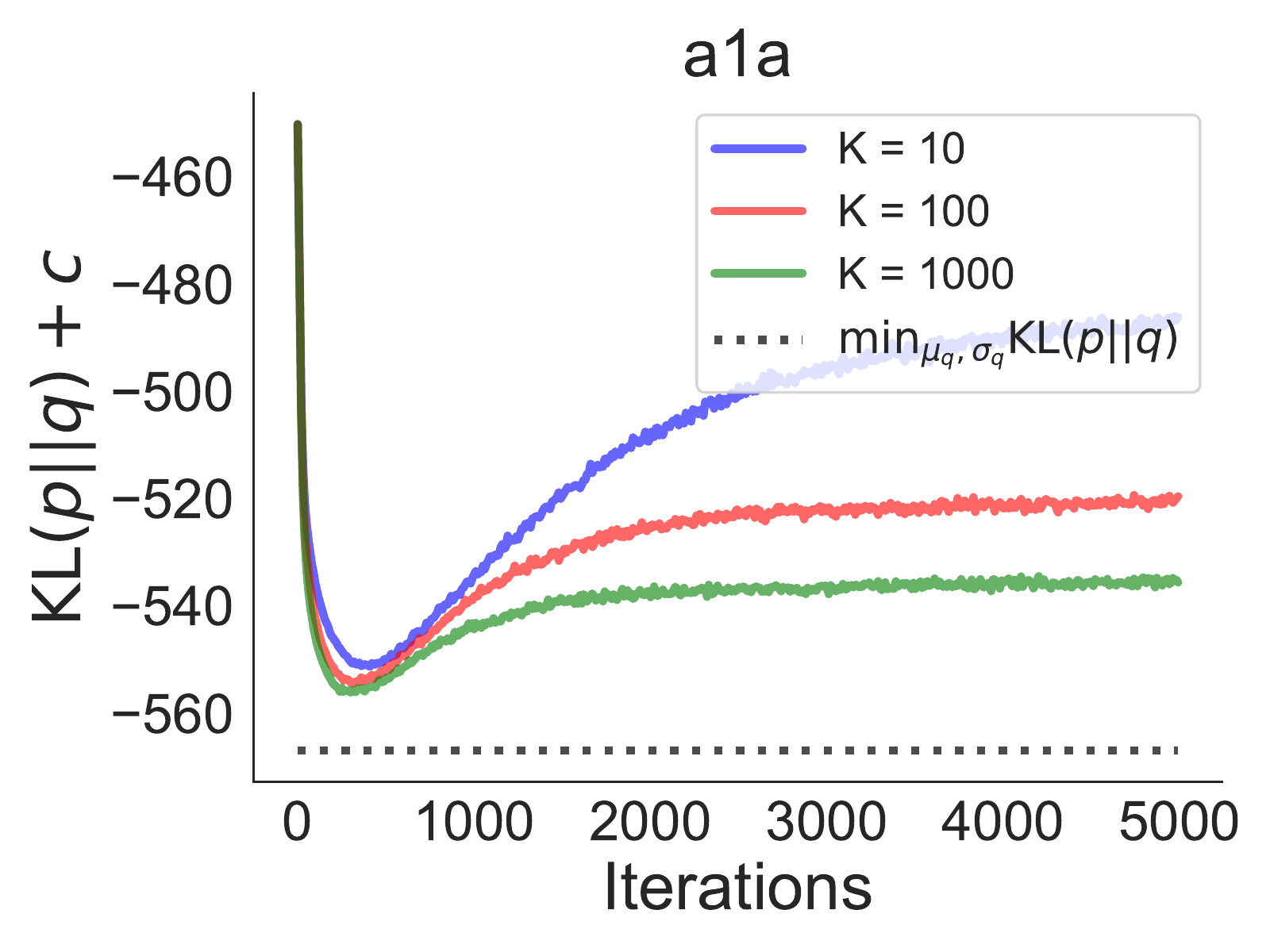}
  }
\end{figure}



Fig.~\ref{fig:params_stl_logreg} shows how the optimal parameters $(\mu_q^*, \sigma_q^*)$ compare against the parameters obtained by optimizing using the biased gradient estimator $g_\mathrm{stl}$. We observe two things. First, the optimal mean parameters are well-recovered for both datasets regardless of the number of samples $K$ used to estimate gradients\footnote{This is probably because optimizing $\mathrm{KL}(p||q)$ and $\mathrm{KL}(q||p)$ gives nearly the same mean parameters on these problems.}. Second, the scale parameters recovered are biased towards minimizers of $\mathrm{KL}(q||p)$. For the \textit{sonar} dataset $(d=61)$, this bias can be removed by increasing the number of samples $K$ used to estimate gradients. However, for the \textit{a1a} dataset $(d = 120)$, increasing $K$ to 1000 provides only a tiny improvement, suggesting a huge value for $K$ would be needed.

Results for all other estimators are similar to the ones shown in this section for $\g_\mathrm{stl}$. We show them in Appendix \ref{app:2}.

\begin{figure}[htbp]
\floatconts
  {fig:params_stl_logreg}
  {\caption{\textbf{In high dimensions optimizing with the biased estimator leads to solutions strongly biased towards minimizers of $\mathrm{KL}(q\Vert p)$, and an impractically large $K$ is needed to mitigate this effect.} Results for the logistic regression model with both datasets, \textit{sonar} $(d = 61)$ and \textit{a1a} $(d = 120)$. The plots show the mean and variance of each component of the variational distribution $q$ obtained by optimizing with the gradient estimator $g_\mathrm{stl}$ with $K$ samples. Components are sorted to facilitate visualization.} }
  {\includegraphics[scale=0.363, trim = {0 2cm 0 0}, clip]{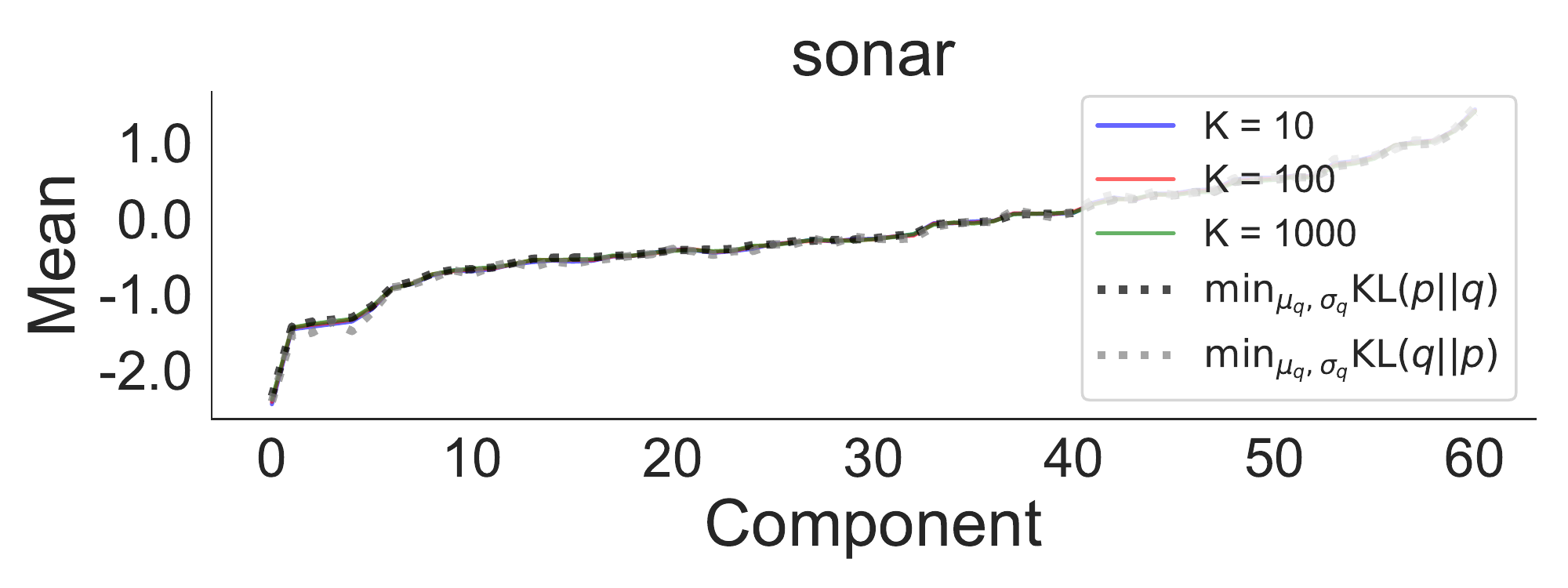}\hspace{0.5cm}
  \includegraphics[scale=0.35, trim = {1.3cm 2cm 0 0}, clip]{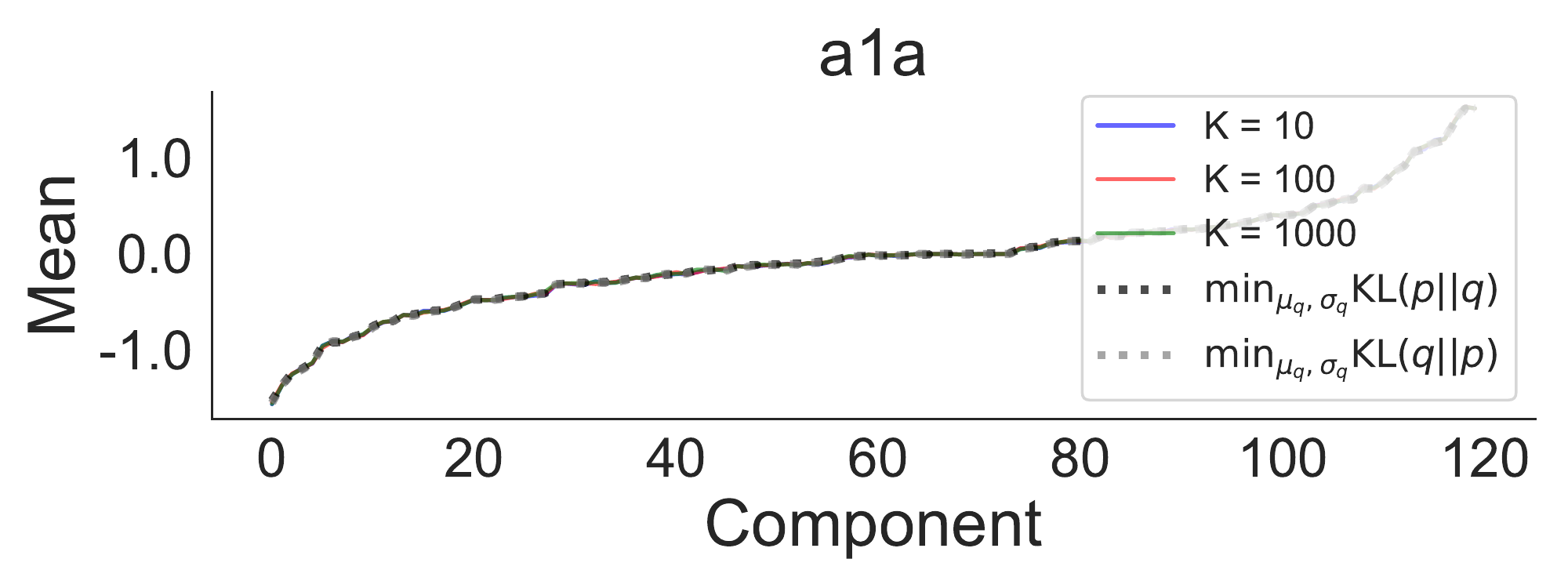}\vspace{0.25cm}
  
  \includegraphics[scale=0.363, trim = {0 0 0 1.15cm}, clip]{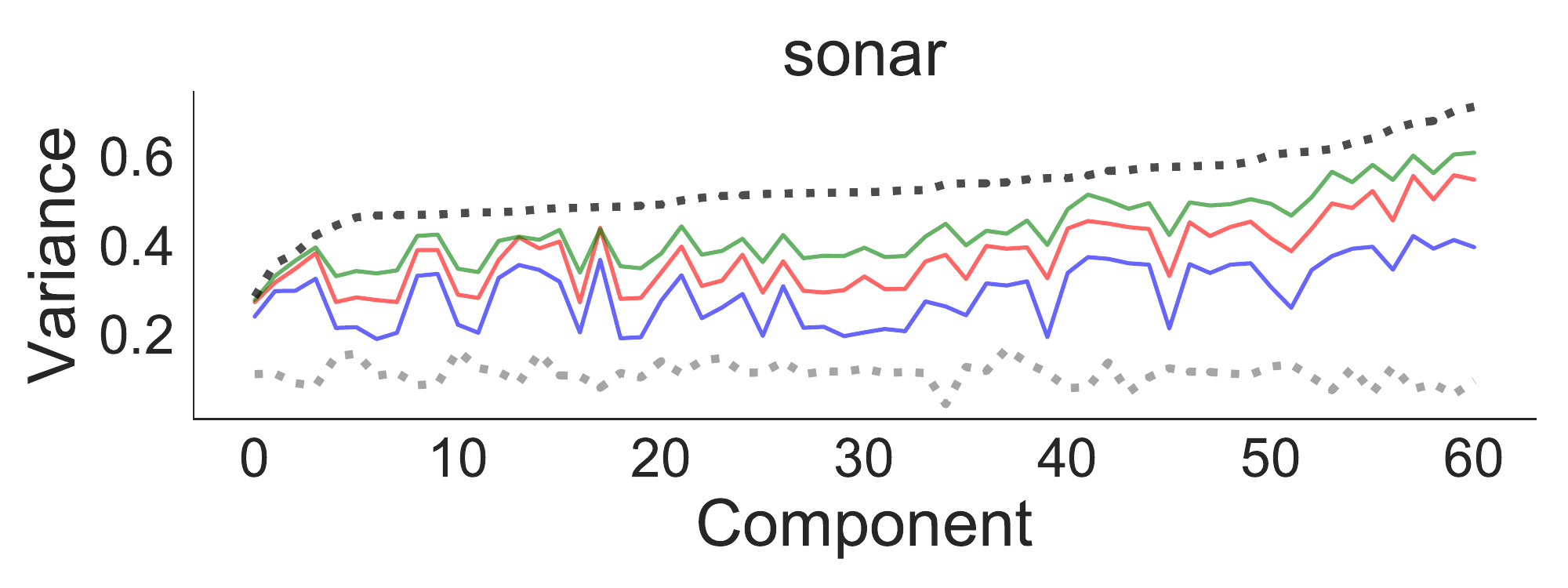}\hspace{0.5cm}
  \includegraphics[scale=0.35, trim = {1.2cm 0 0 1.15cm}, clip]{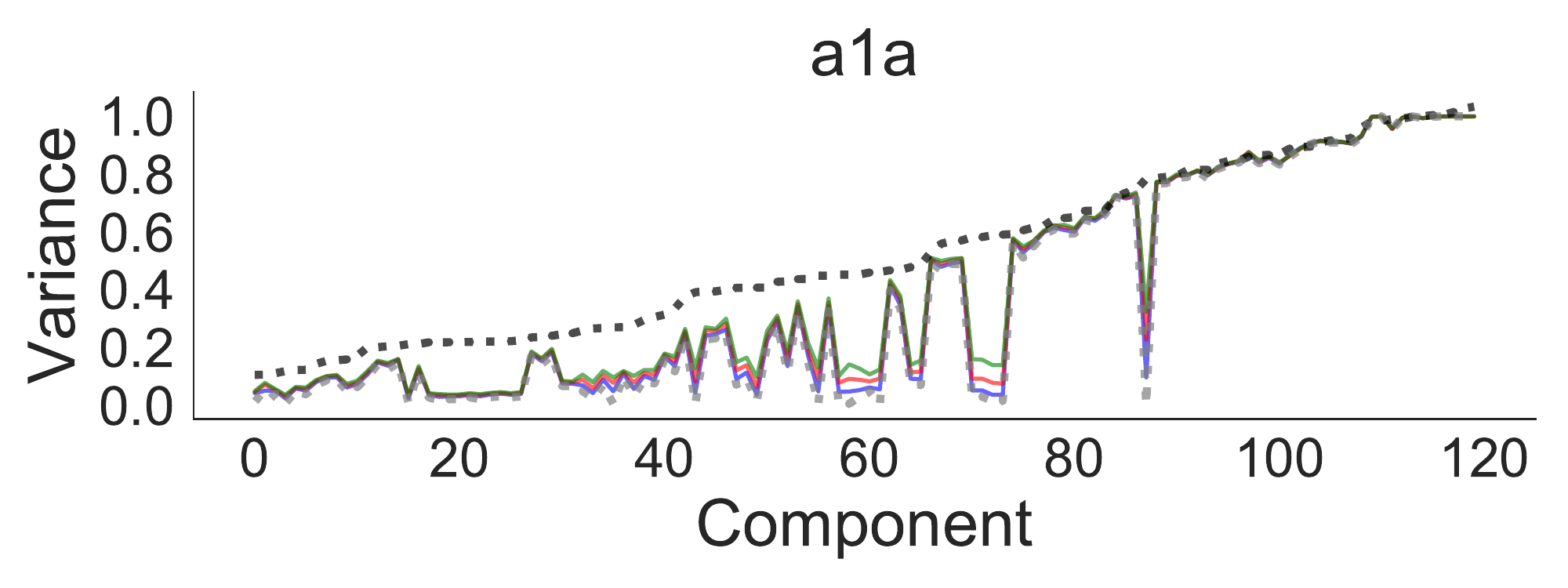}
  }
\end{figure}

\section{Conclusions}

All gradient estimators analyzed are asymptotically unbiased (except $g_\mathrm{chivi}$). This means that, if a large enough number of samples $K$ is used to estimate gradients, these methods are guaranteed to return near-optimal solutions. In practice, however, we observe that even for very simple problems, the value of $K$ needed is typically very large.

Interestingly, solutions returned by these methods appear to be biased towards minimizers of $\mathrm{KL}(q||p)$. Upon close examination, it is not obvious why this should be true and to the best of our knowledge no theoretical support for this behavior is known. We find this surprising and consider it to be an appealing property of these methods: Even when they fail to minimize the target alpha-divergence, they do something ``reasonable'', i.e. minimize the traditional divergence $\mathrm{KL}(q||p)$. 

\bibliography{jmlr-sample}

\appendix

\clearpage
\newpage

\section{Optimization results for all estimators with Gaussian model}\label{apd:1}

\begin{figure}[htbp]
\floatconts
  {fig:target_method_gauss_end}
  {\caption{Optimization results for all estimators.}}
  {
  \includegraphics[scale=0.28, trim = {0 0 0 0}, clip]{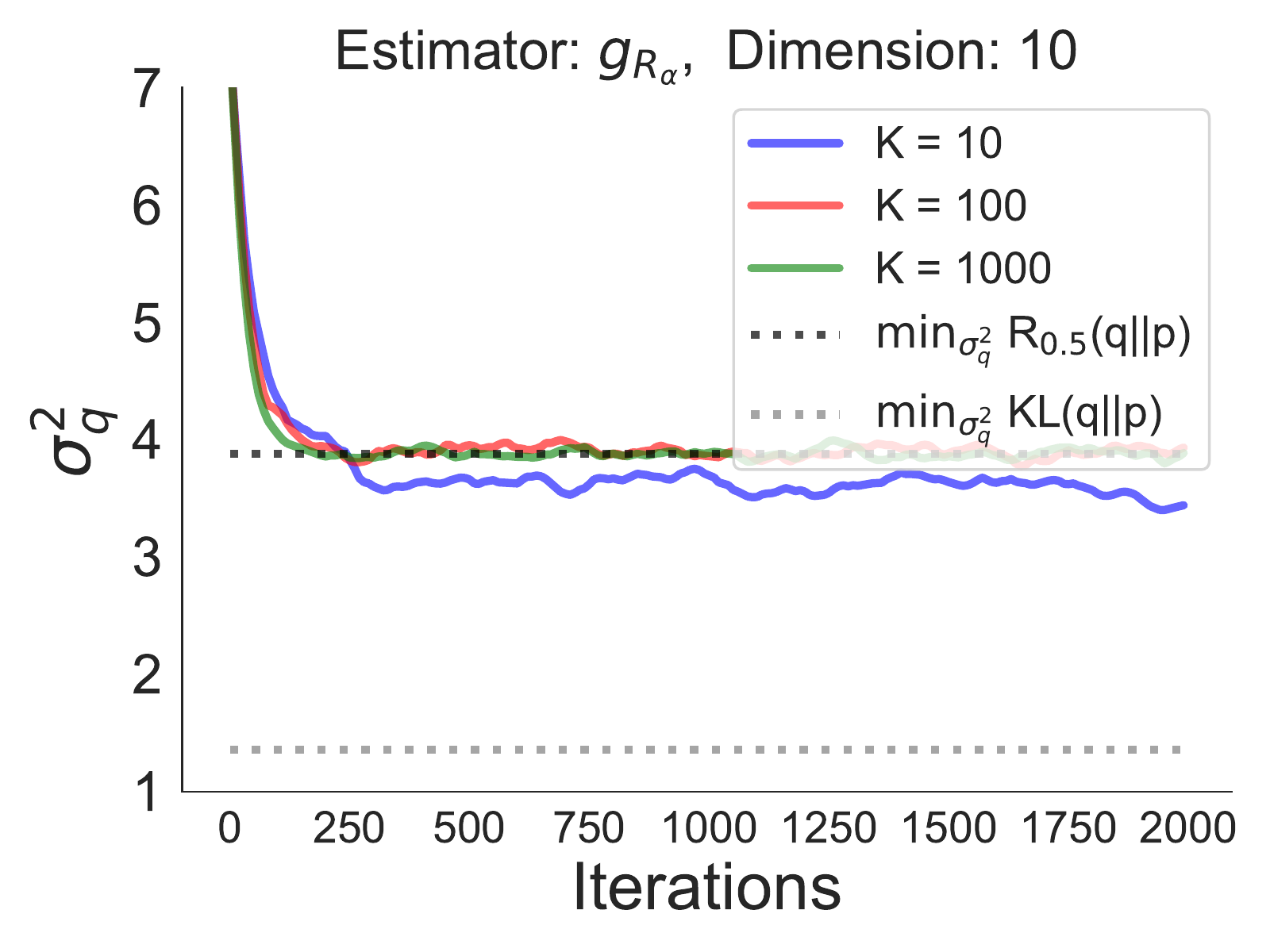}\hspace{0.5cm}
  \includegraphics[scale=0.28, trim = {2.1cm 0 0 0}, clip]{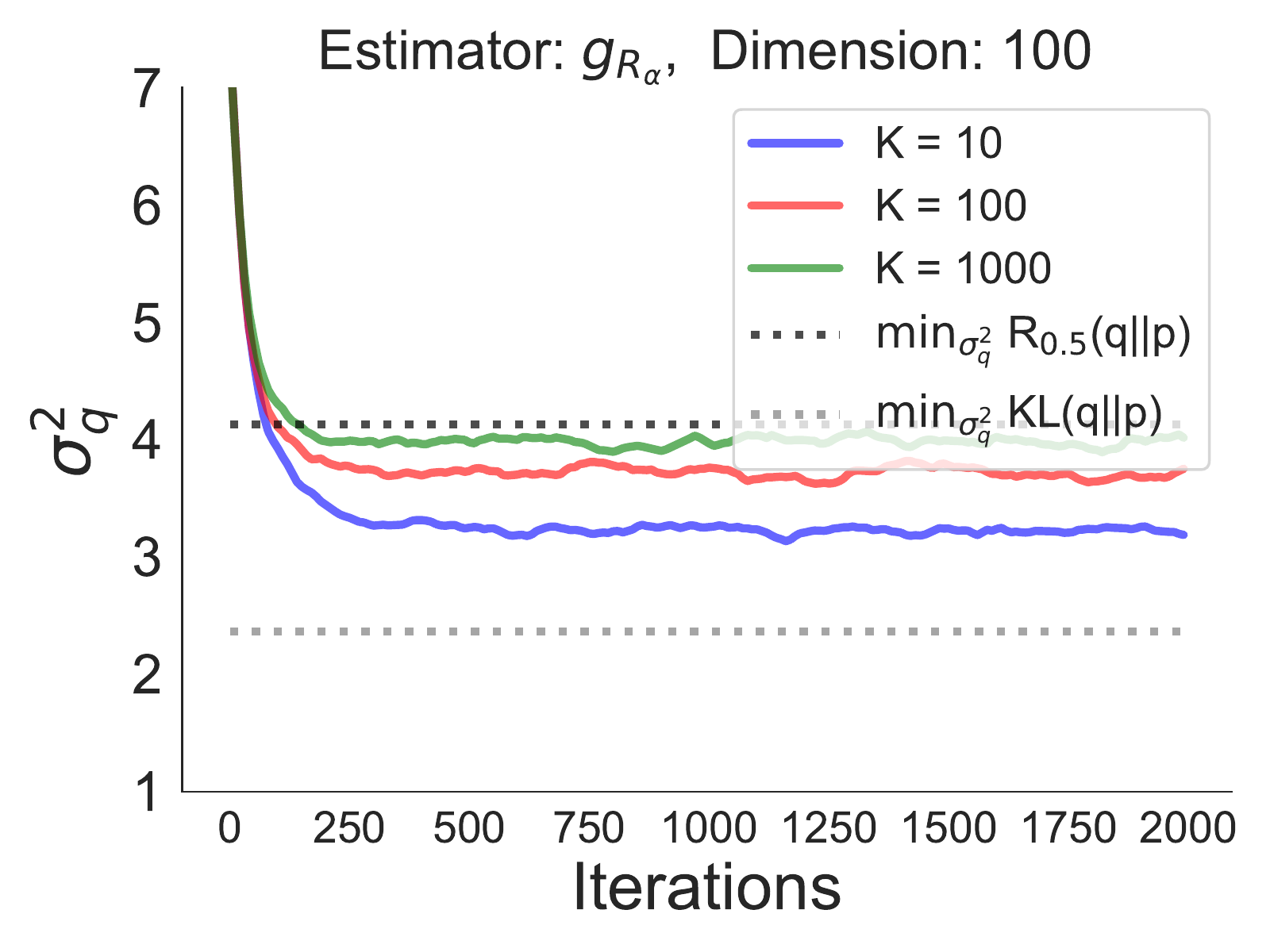}\hspace{0.5cm}
  \includegraphics[scale=0.28, trim = {2.1cm 0 0 0}, clip]{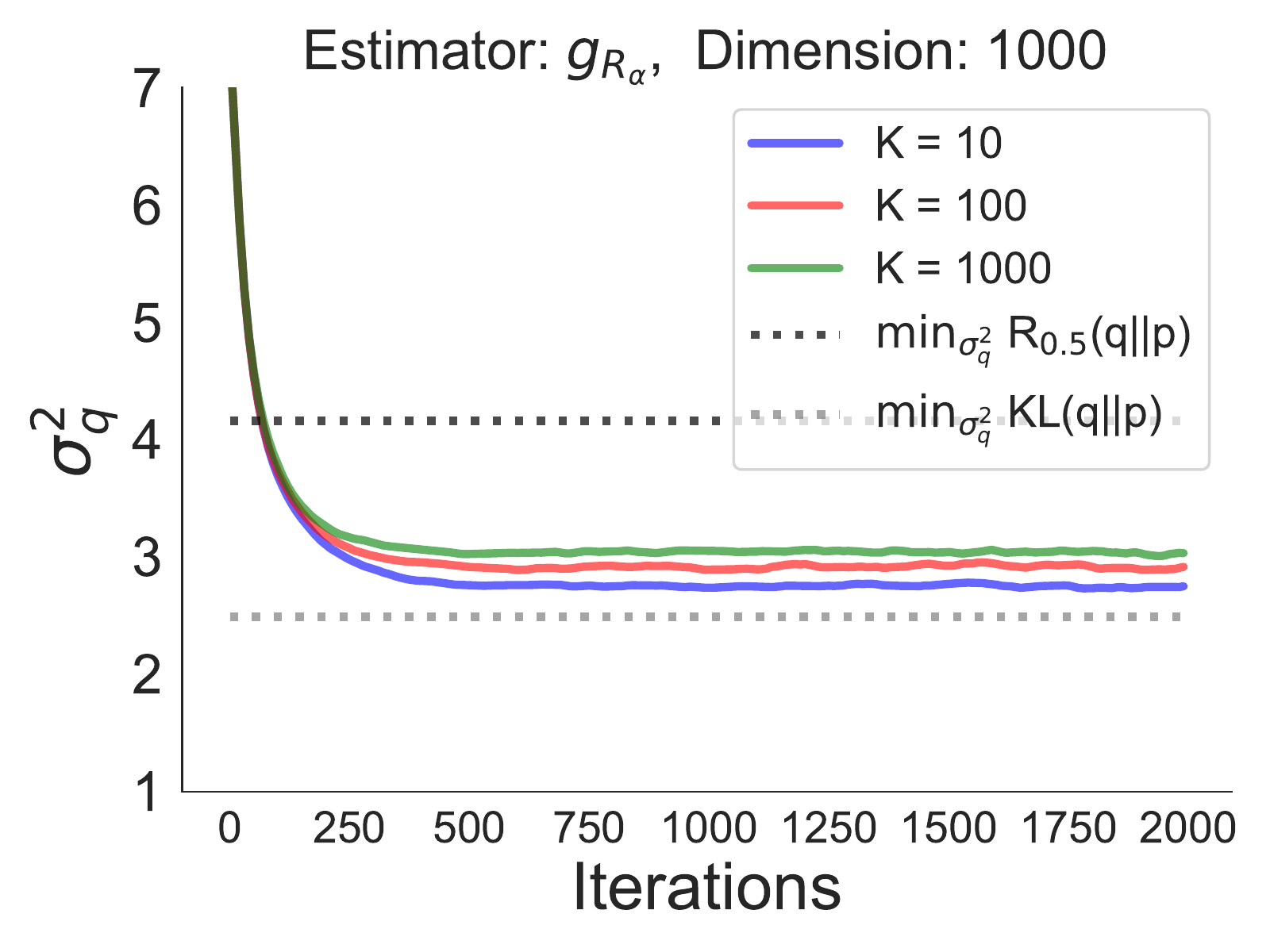}
  
  \vspace{0.3cm}
  
  \includegraphics[scale=0.28, trim = {0 0 0 0}, clip]{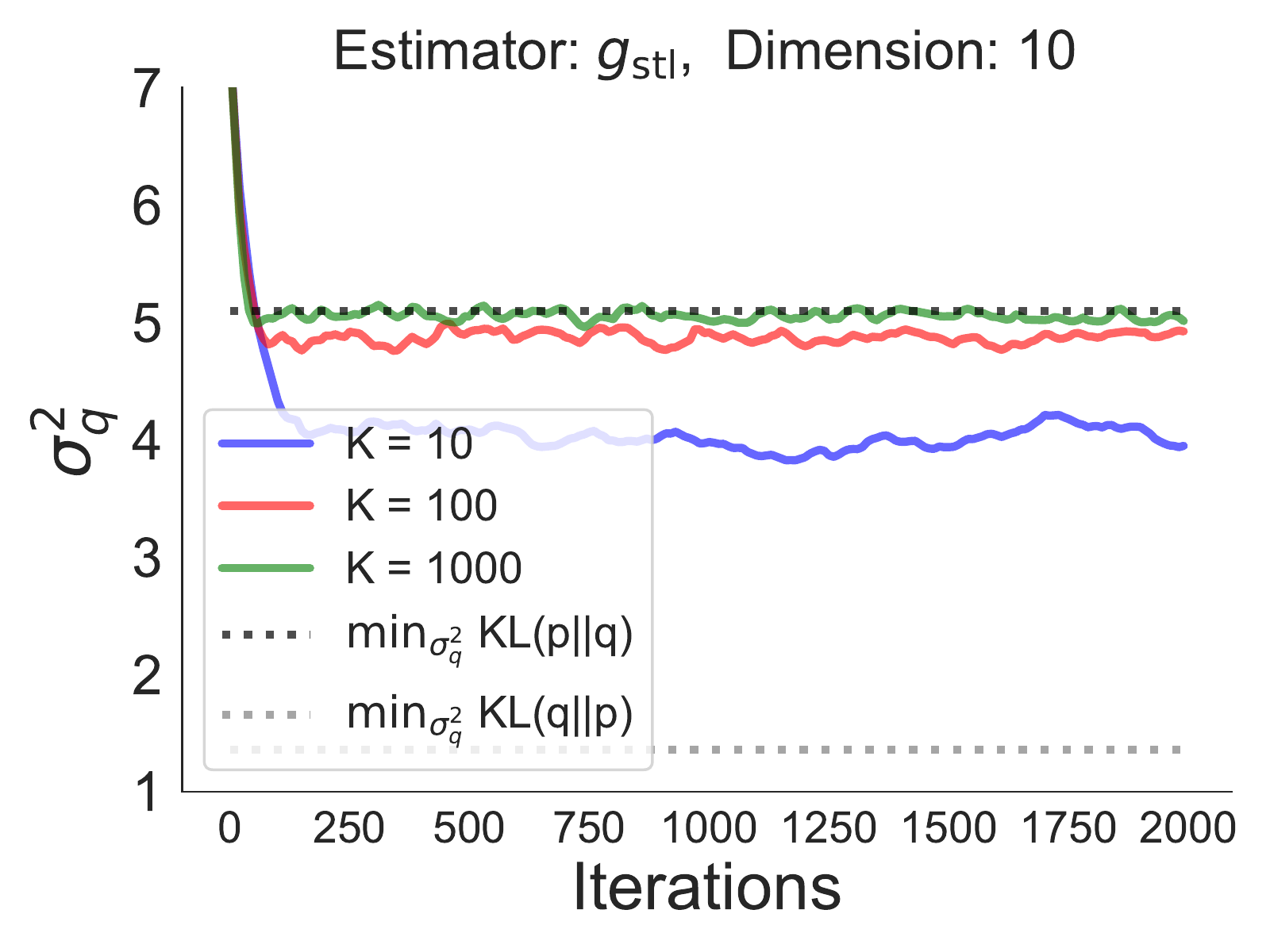}\hspace{0.5cm}
  \includegraphics[scale=0.28, trim = {2.1cm 0 0 0}, clip]{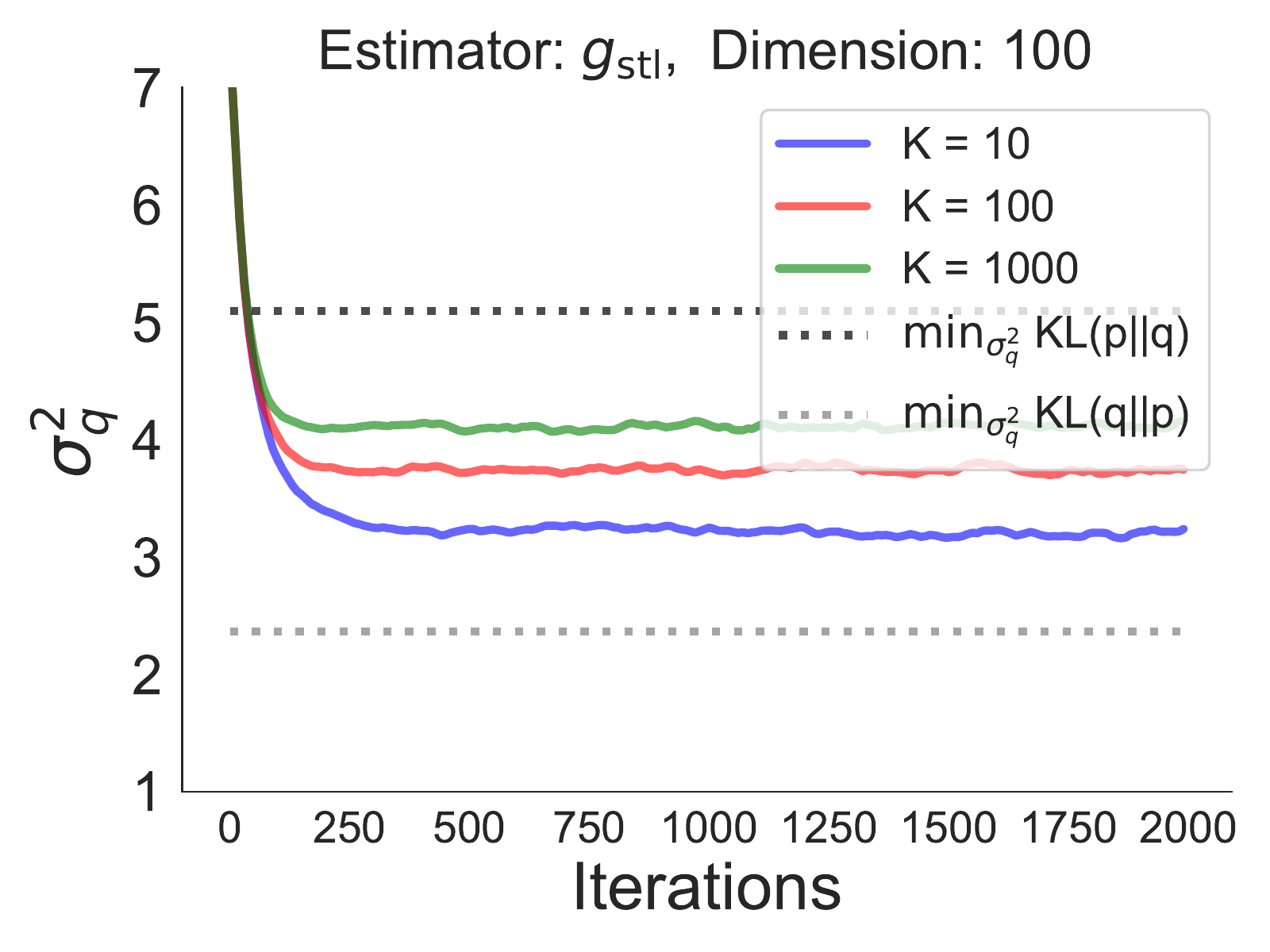}\hspace{0.5cm}
  \includegraphics[scale=0.28, trim = {2.1cm 0 0 0}, clip]{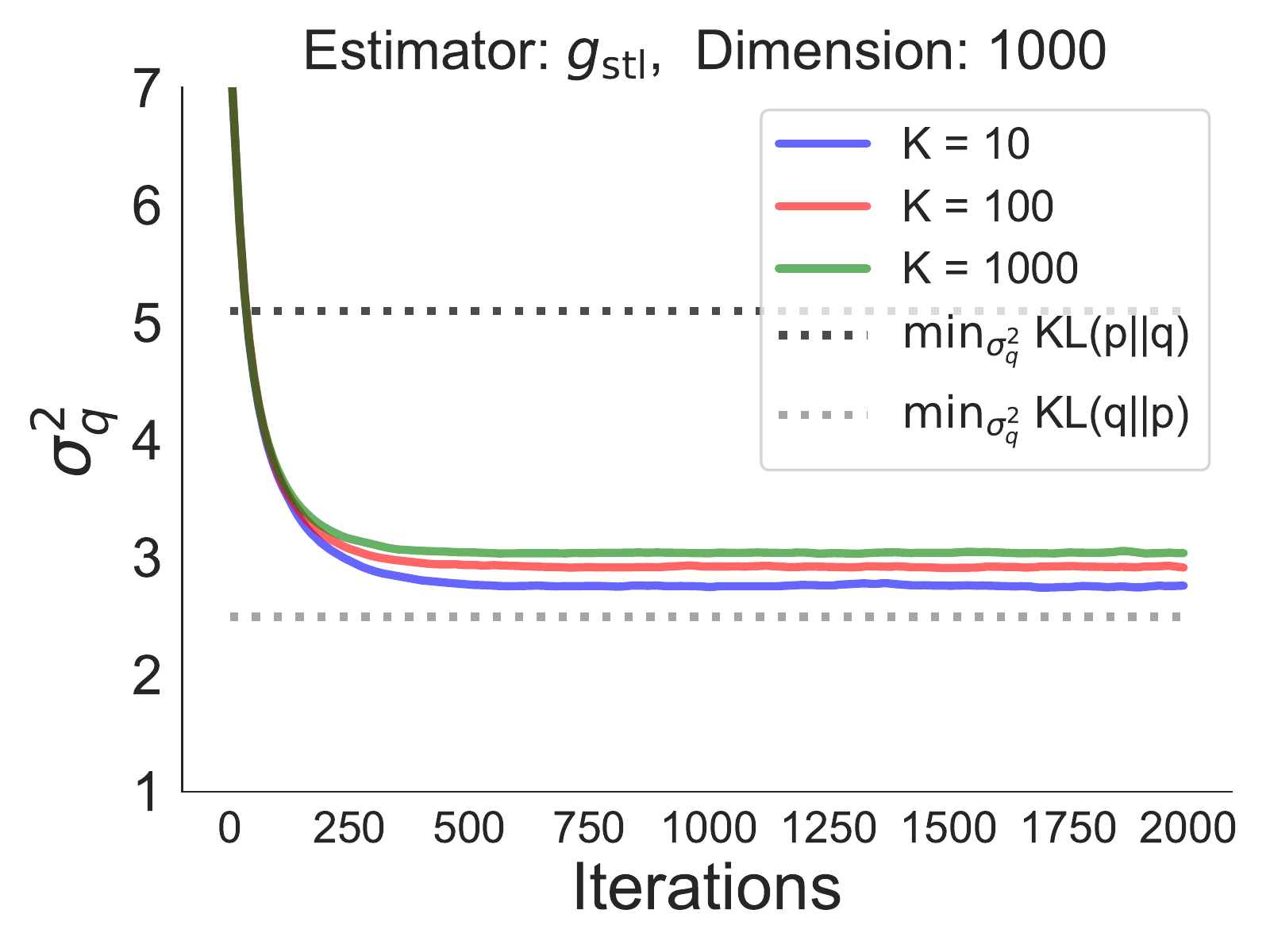}
  
  \vspace{0.3cm}
  
  \includegraphics[scale=0.28, trim = {0 0 0 0}, clip]{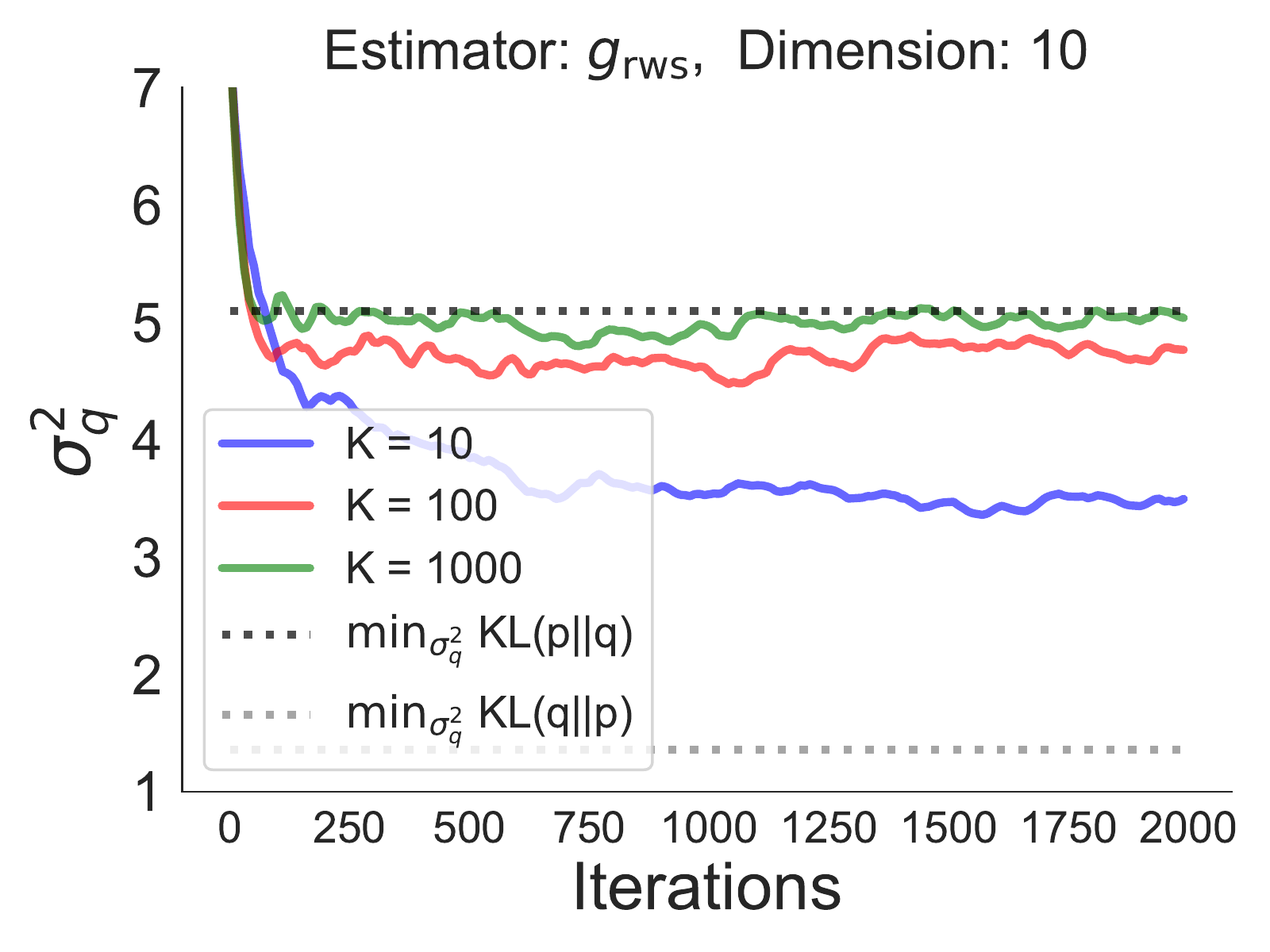}\hspace{0.5cm}
  \includegraphics[scale=0.28, trim = {2.1cm 0 0 0}, clip]{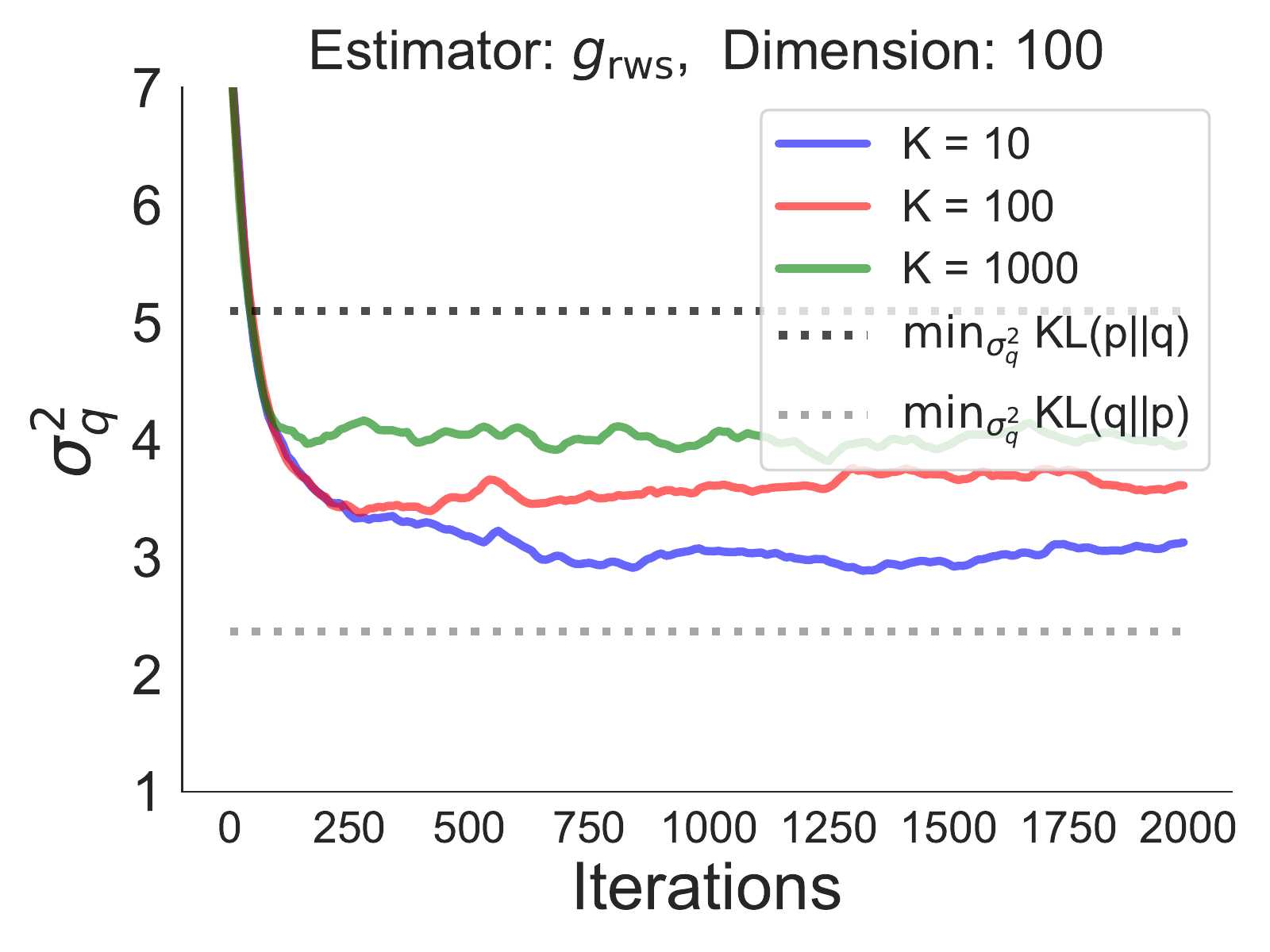}\hspace{0.5cm}
  \includegraphics[scale=0.28, trim = {2.1cm 0 0 0}, clip]{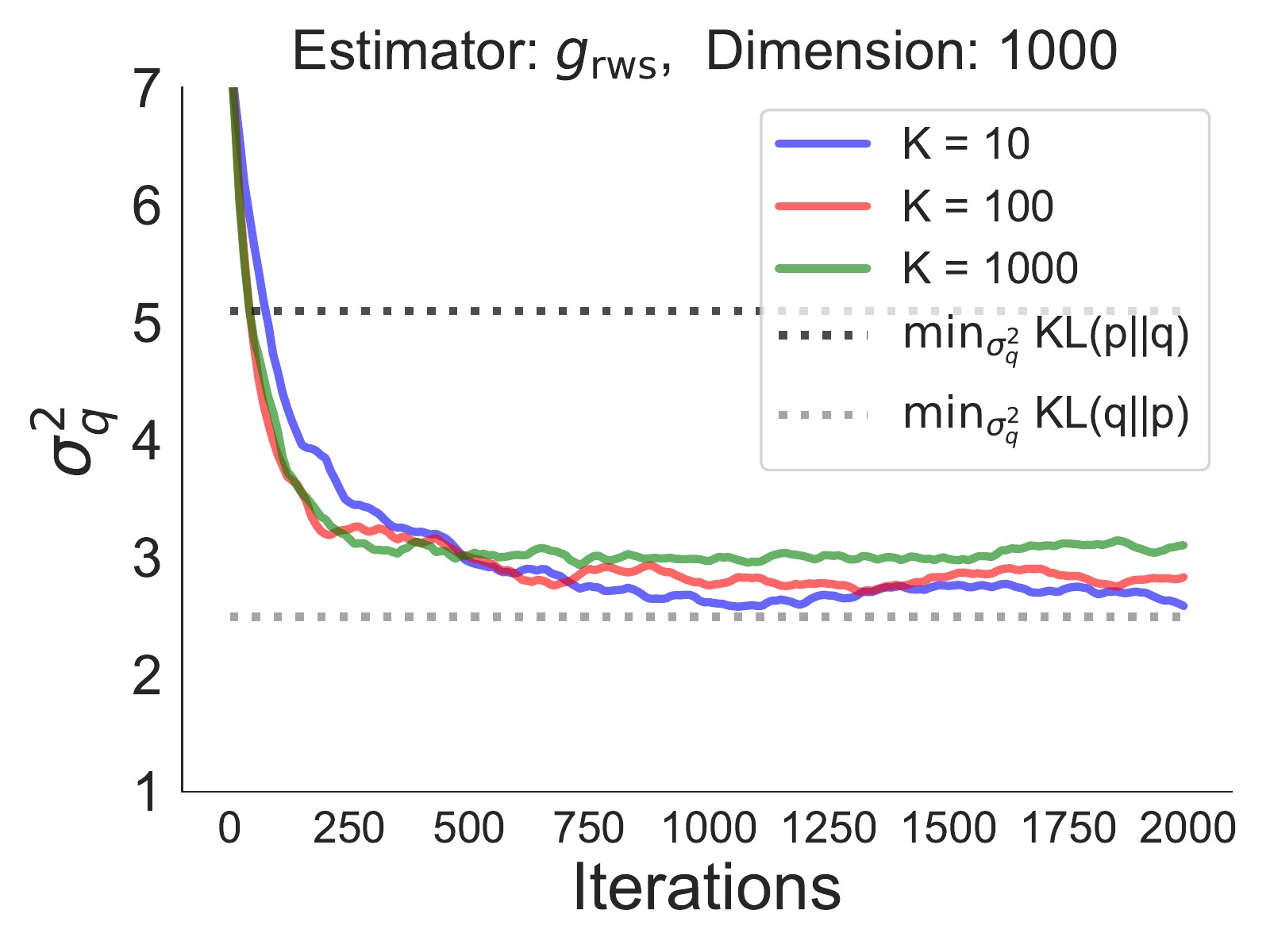}
  
  \vspace{0.3cm}
  
  \includegraphics[scale=0.28, trim = {0 0 0 0}, clip]{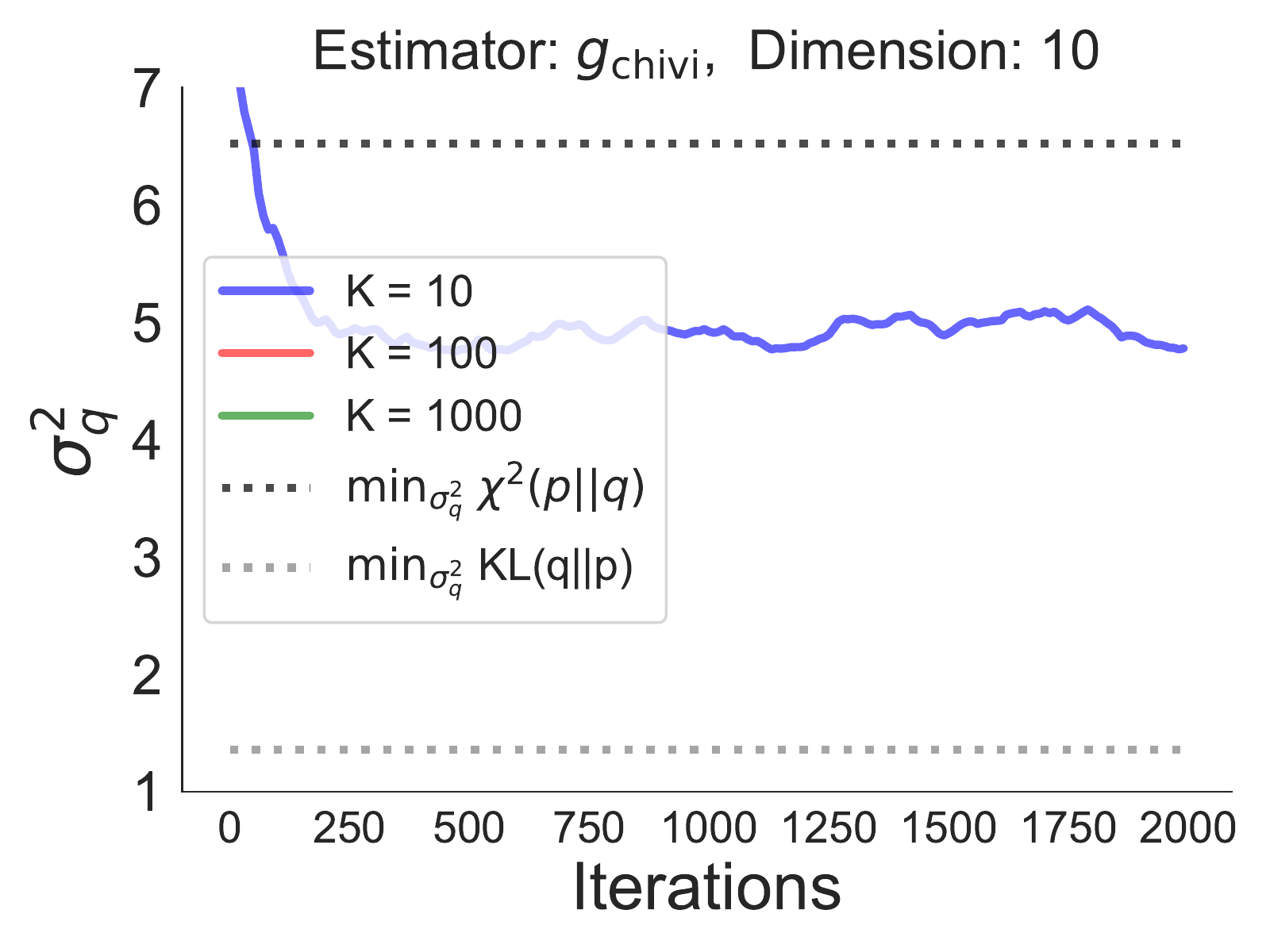}\hspace{0.5cm}
  \includegraphics[scale=0.28, trim = {2.1cm 0 0 0}, clip]{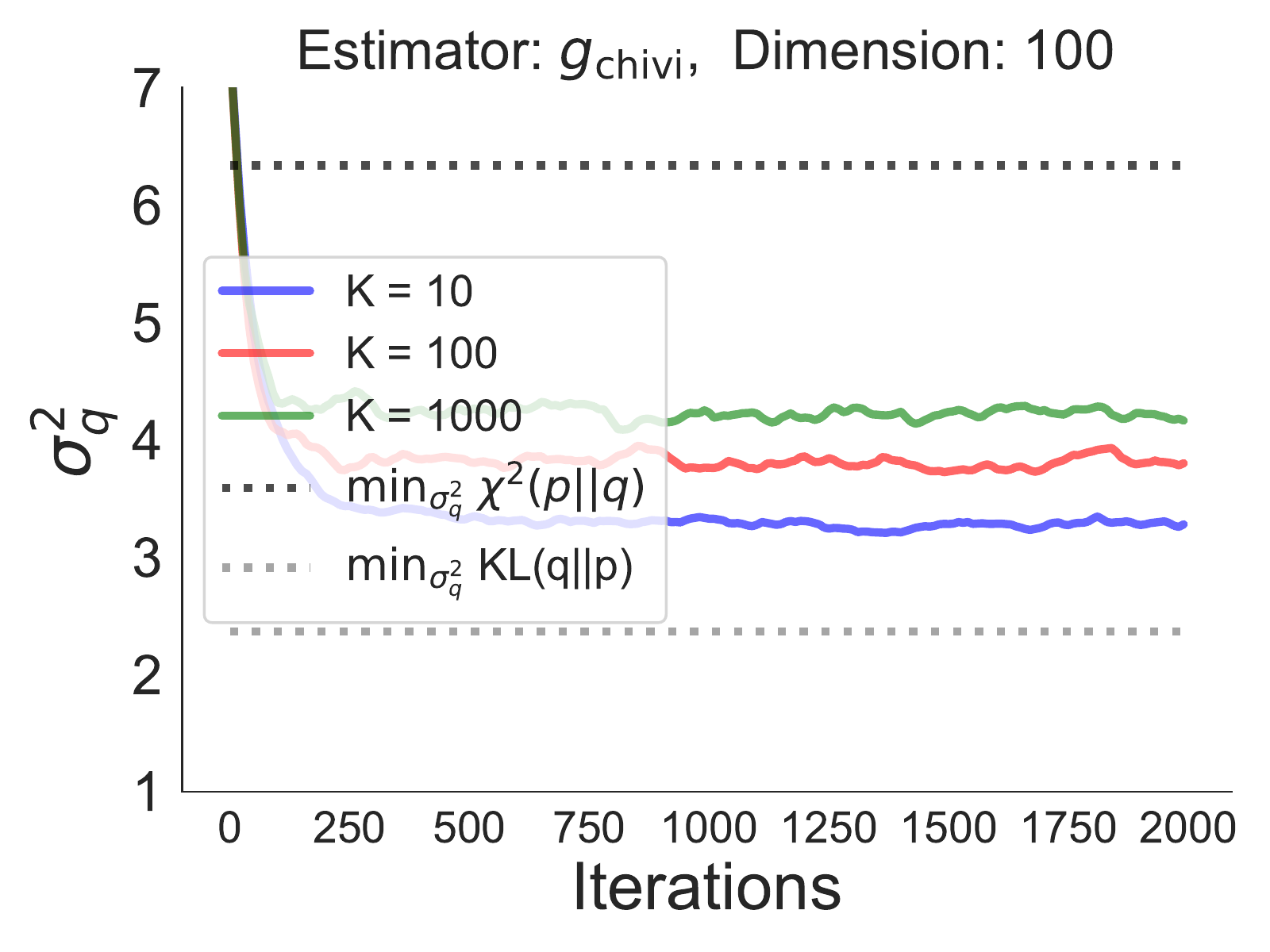}\hspace{0.5cm}
  \includegraphics[scale=0.28, trim = {2.1cm 0 0 0}, clip]{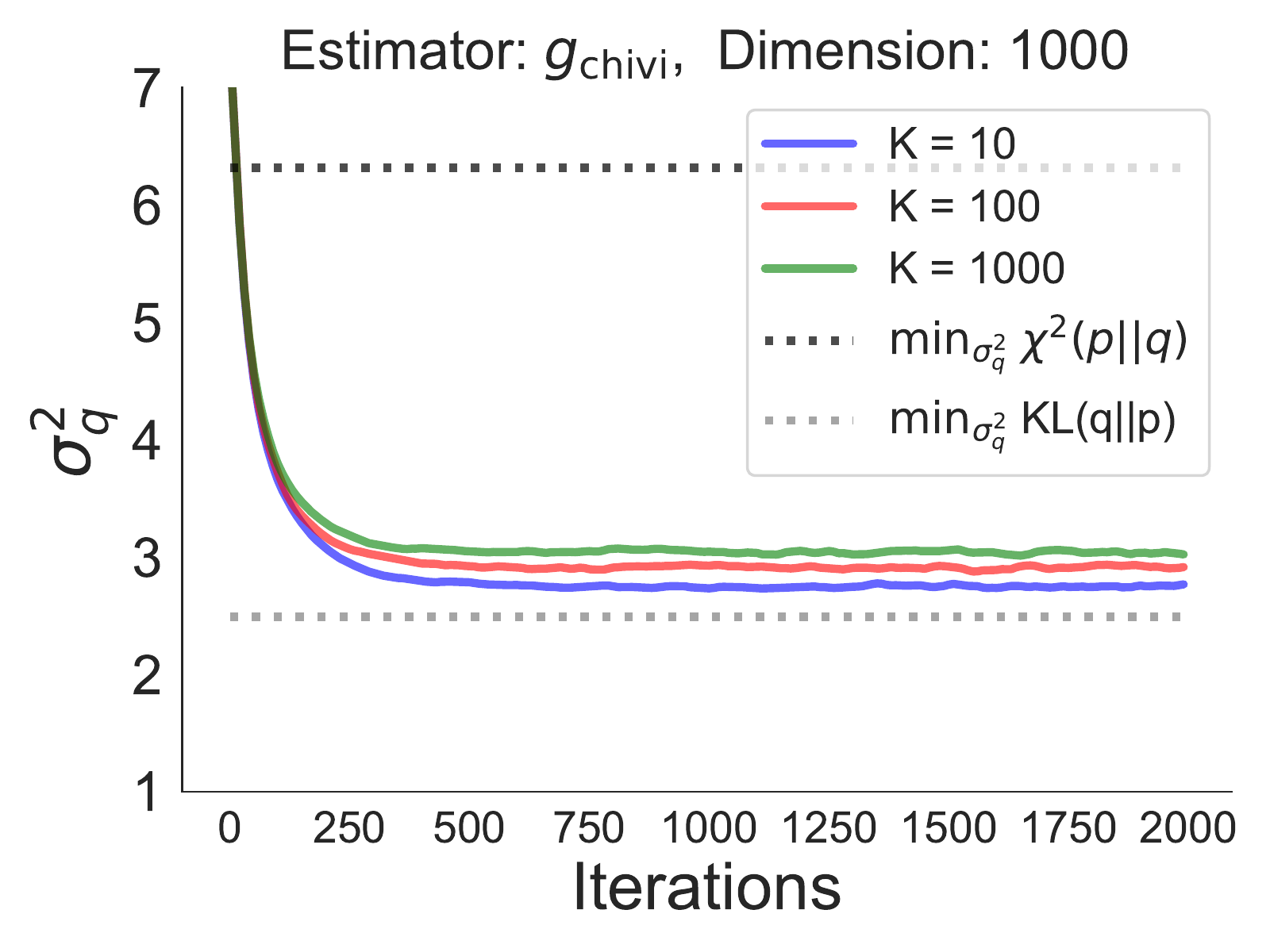}
  
  \vspace{0.3cm}
  
  \includegraphics[scale=0.28, trim = {0 0 0 0}, clip]{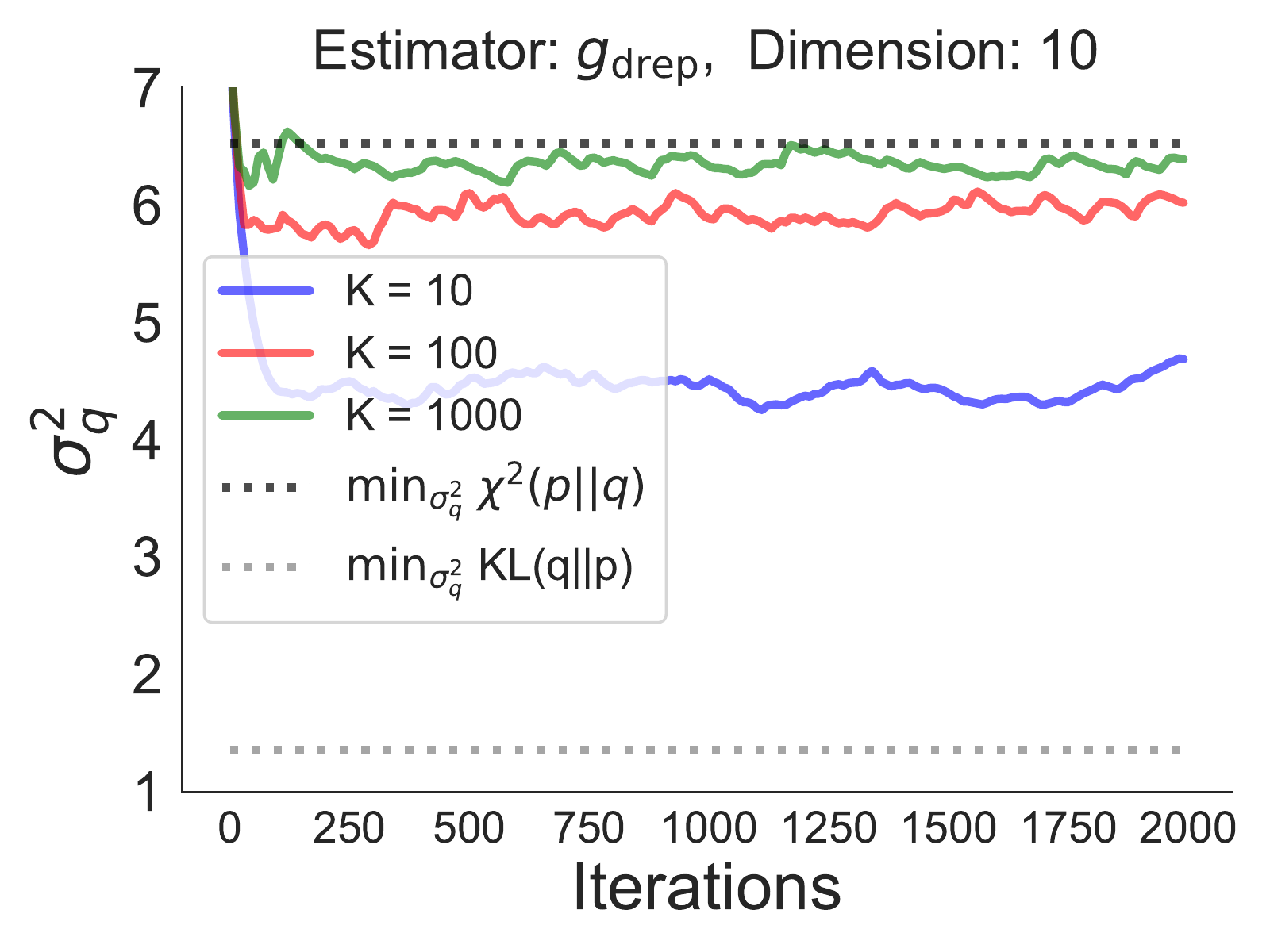}\hspace{0.5cm}
  \includegraphics[scale=0.28, trim = {2.1cm 0 0 0}, clip]{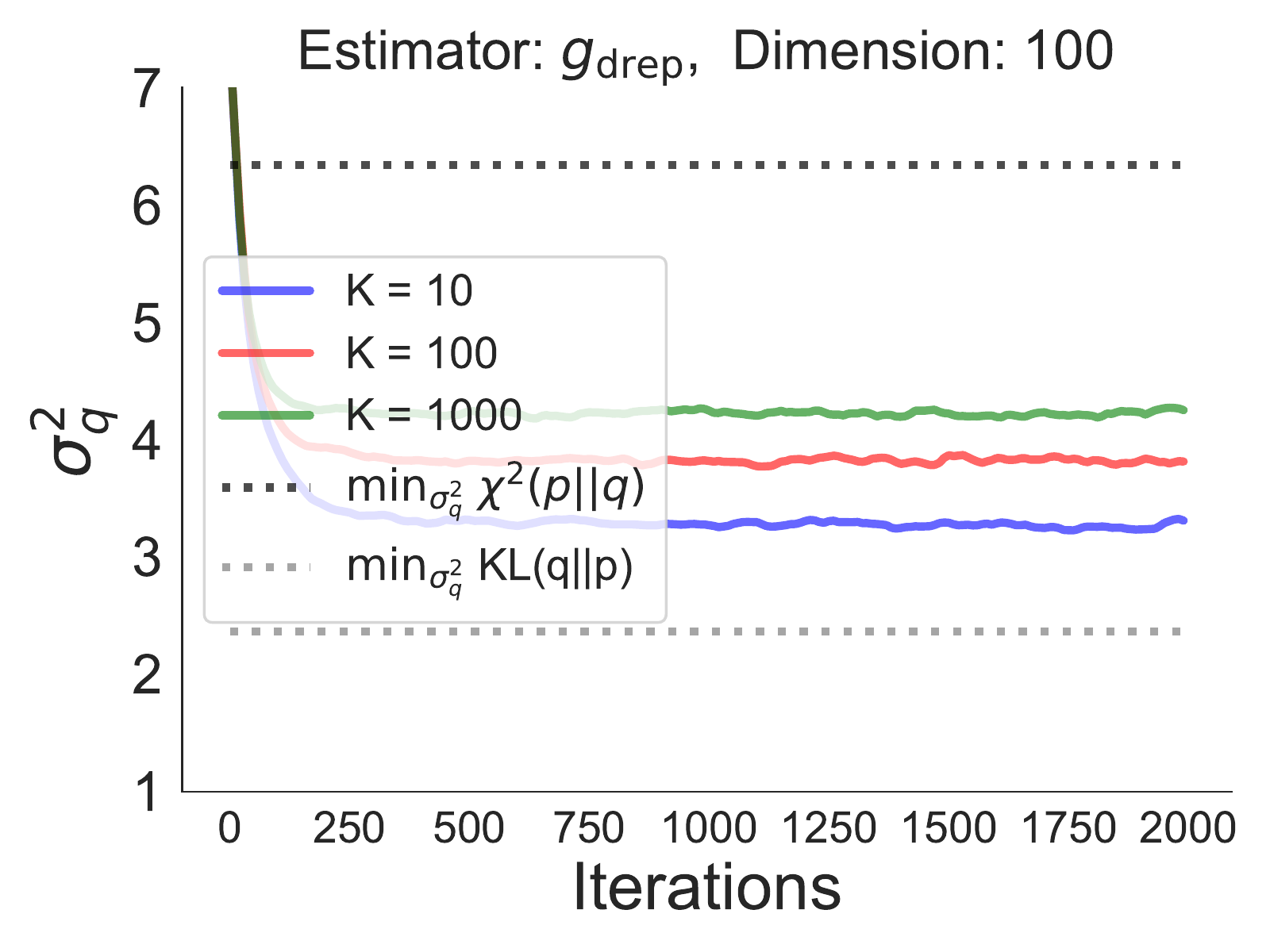}\hspace{0.5cm}
  \includegraphics[scale=0.28, trim = {2.1cm 0 0 0}, clip]{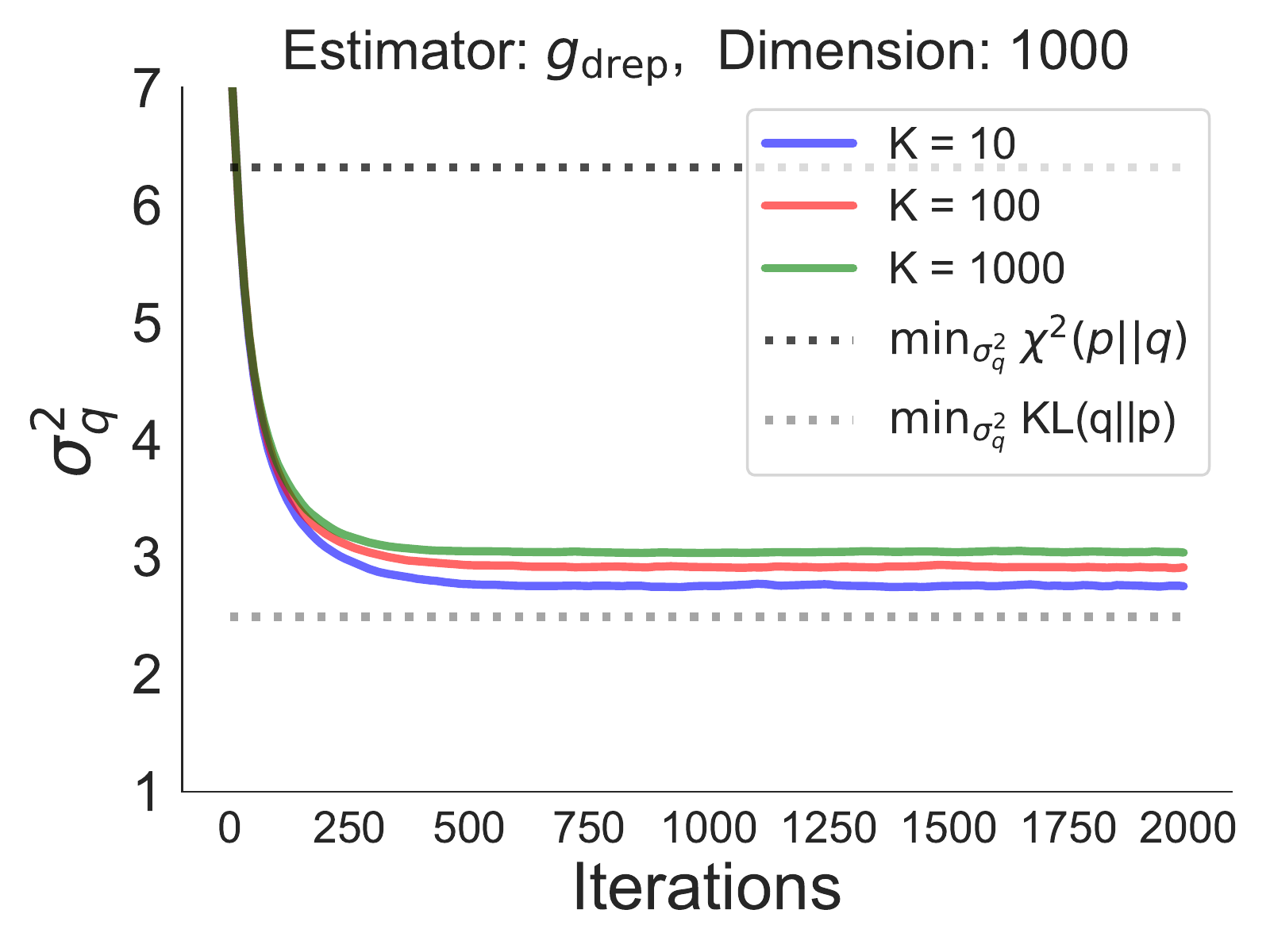}
  }
\end{figure}

\clearpage
\newpage

\section{Optimization results for all estimators with logistic regression model}\label{app:2}

This section shows the results obtained for the logistic regression model using all of the estimators introduced in Section \ref{sec:estimators}. Fig.~\ref{fig:end_opt_sonar} shows the results for the \textit{sonar} dataset $(d=61)$, and Fig.~\ref{fig:end_opt_a1a} the results for the \textit{a1a} dataset $(d=120)$. As mentioned in the main text, results are similar for all methods. For both datasets they all recover the optimal mean parameters correctly. This is not the case for the scale parameters. For the dataset of lower dimensionality, \textit{sonar}, increasing the number of samples $K$ used to compute gradients leads to improved solutions. However, for the \textit{a1a} dataset, the solutions obtained tend to be close to minimizers of $\mathrm{KL}(q||p)$, and increasing the number of samples $K$ up to 1000 leads to only marginal gains.

\begin{figure}[htbp]
\floatconts
  {fig:end_opt_sonar}
  {\caption{Optimization results for the \textit{sonar} dataset ($d=61$) with all estimators.}}
  {
  \includegraphics[scale=0.35, trim = {0 2cm 0 0}, clip]{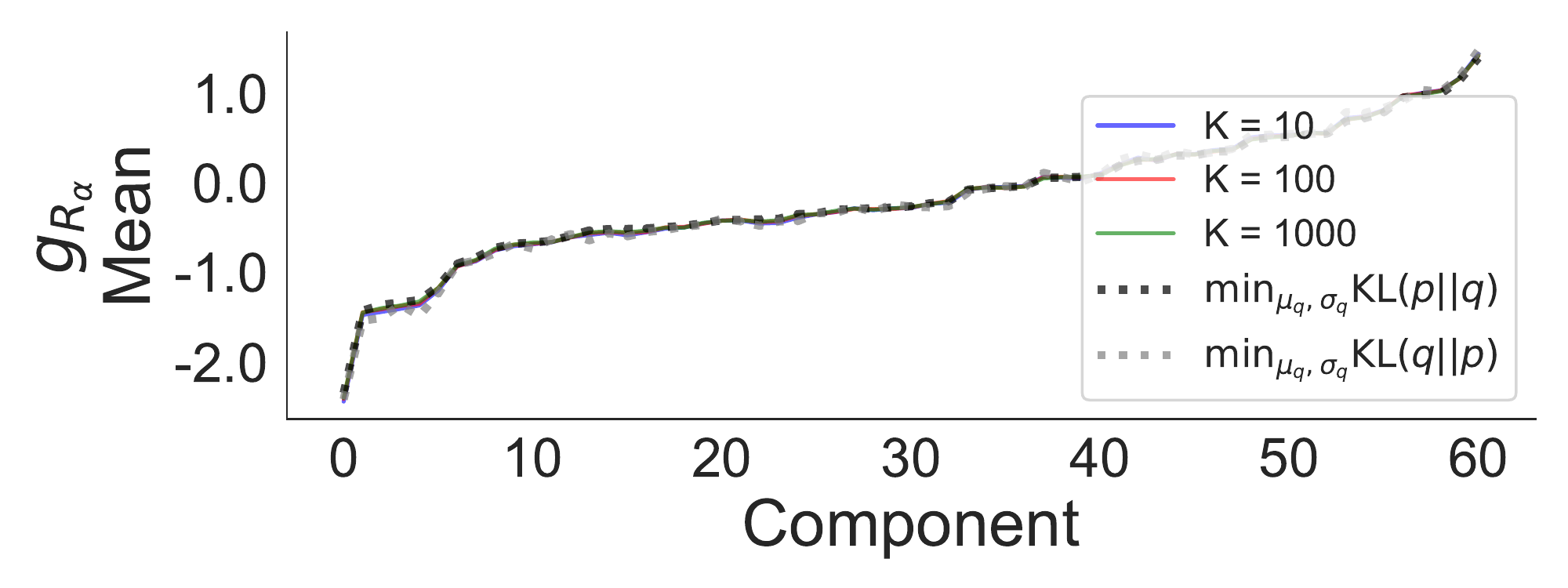}\hspace{0.25cm}
  \includegraphics[scale=0.35, trim = {0 2cm 0 0}, clip]{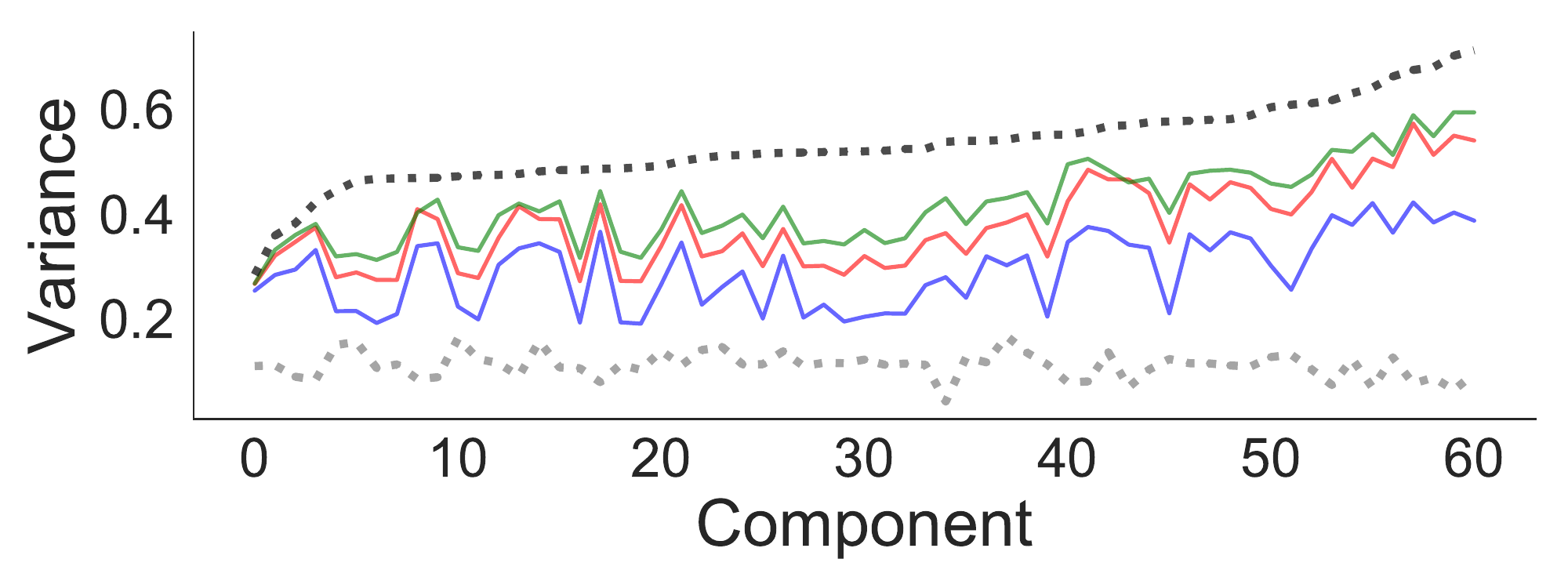}\vspace{0.1cm}
  
  \includegraphics[scale=0.35, trim = {0 2cm 0 0}, clip]{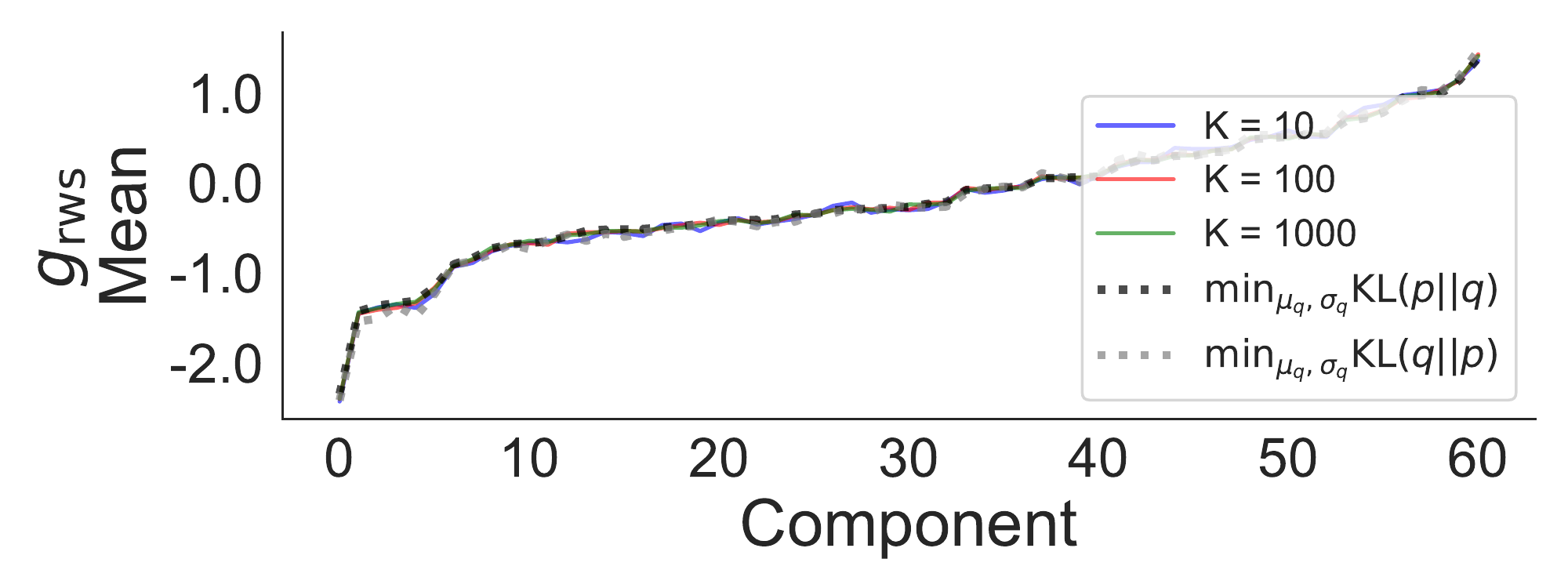}\hspace{0.25cm}
  \includegraphics[scale=0.35, trim = {0 2cm 0 0}, clip]{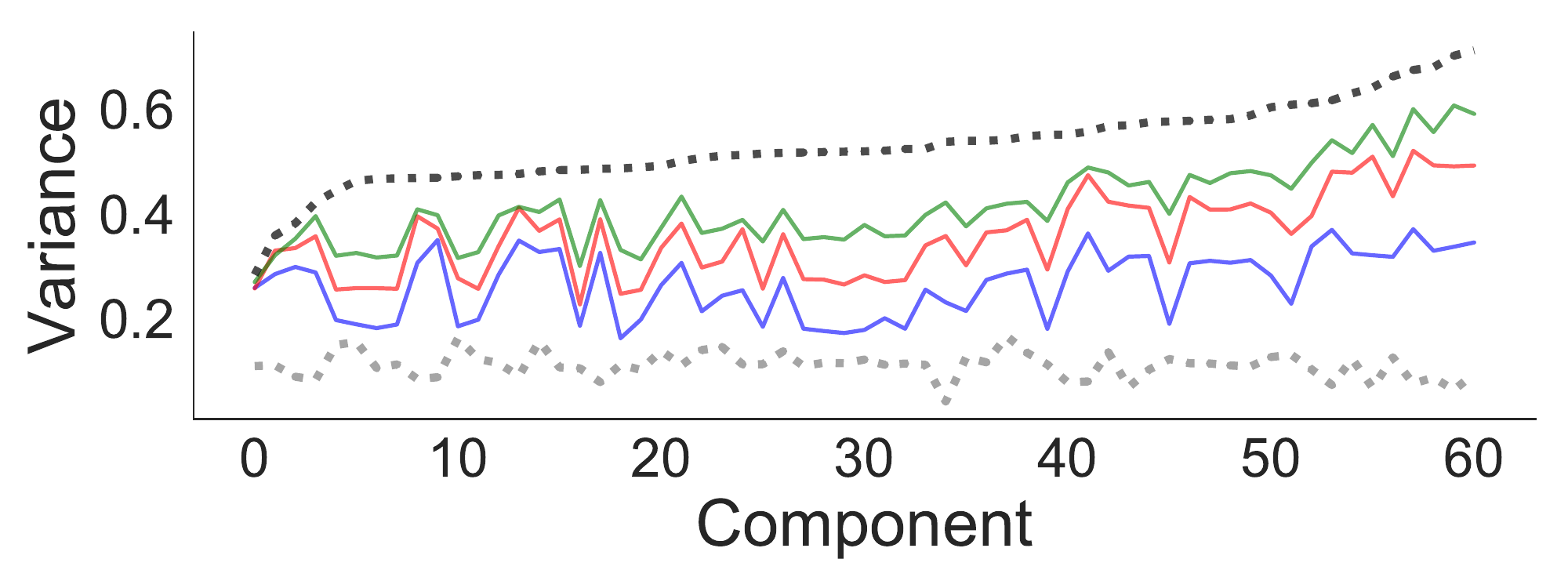}\vspace{0.1cm}
  
  \includegraphics[scale=0.35, trim = {0 2cm 0 0}, clip]{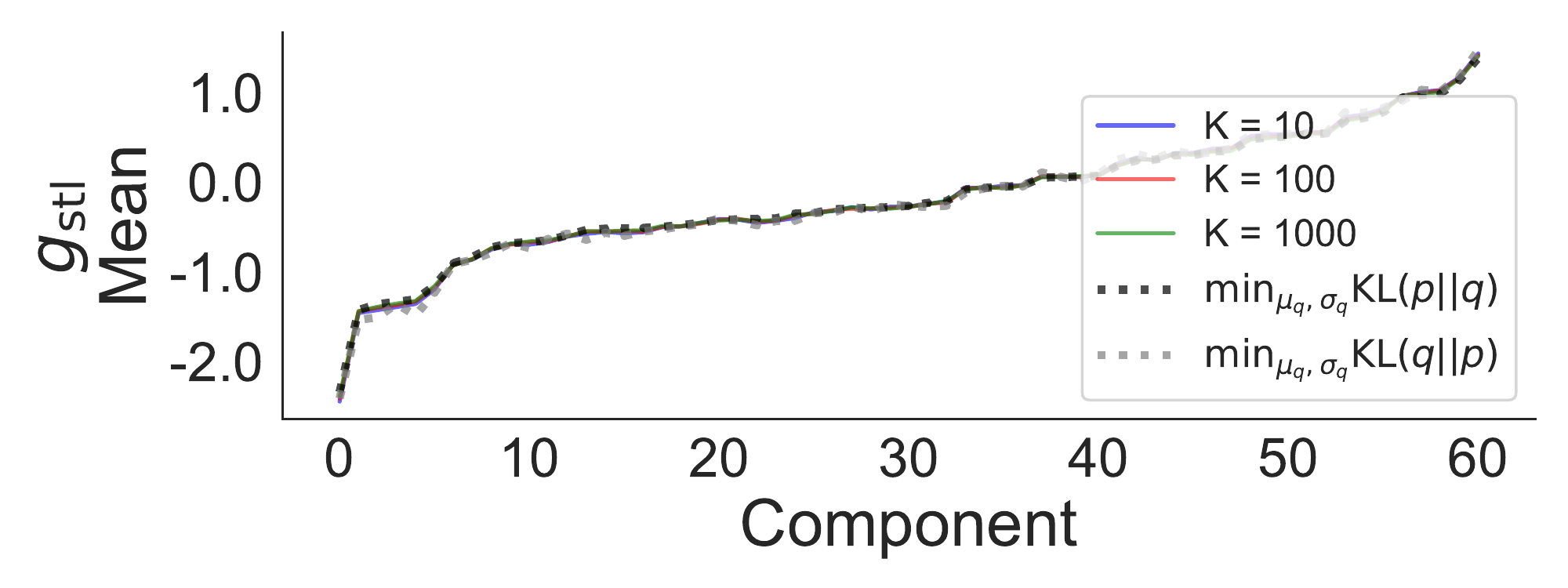}\hspace{0.25cm}
  \includegraphics[scale=0.35, trim = {0 2cm 0 0}, clip]{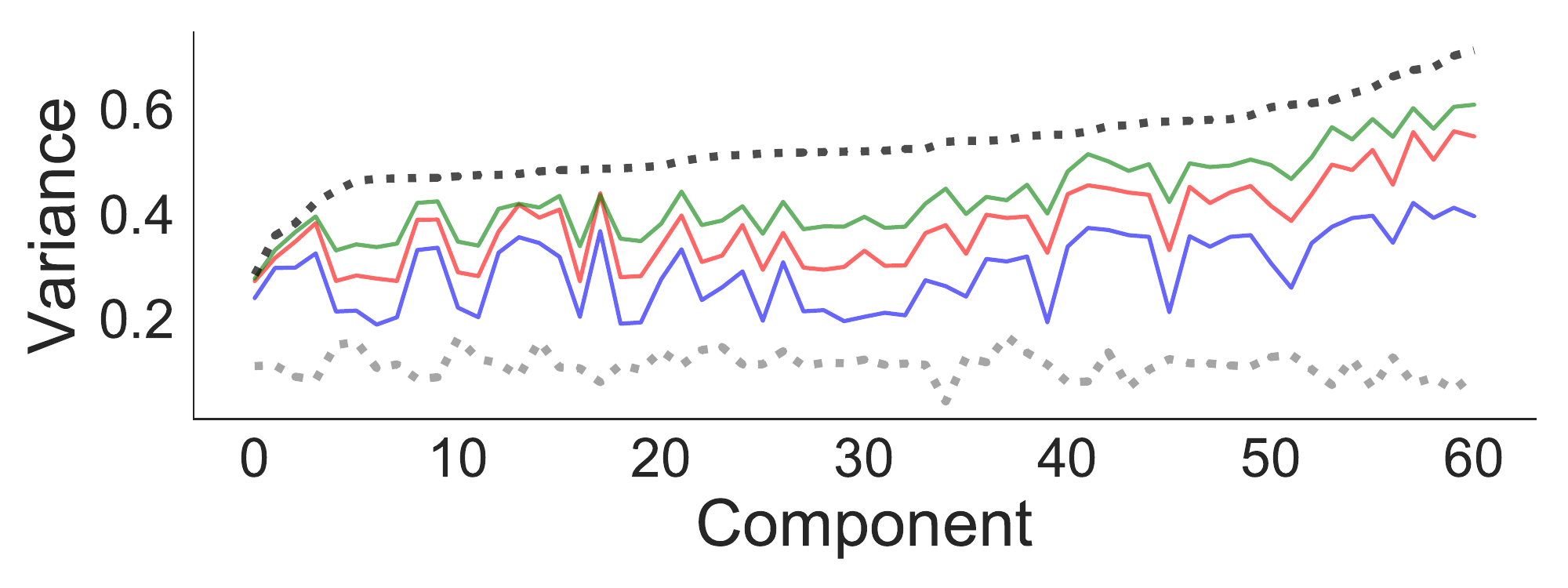}\vspace{0.1cm}
  
  \includegraphics[scale=0.35, trim = {0 2cm 0 0}, clip]{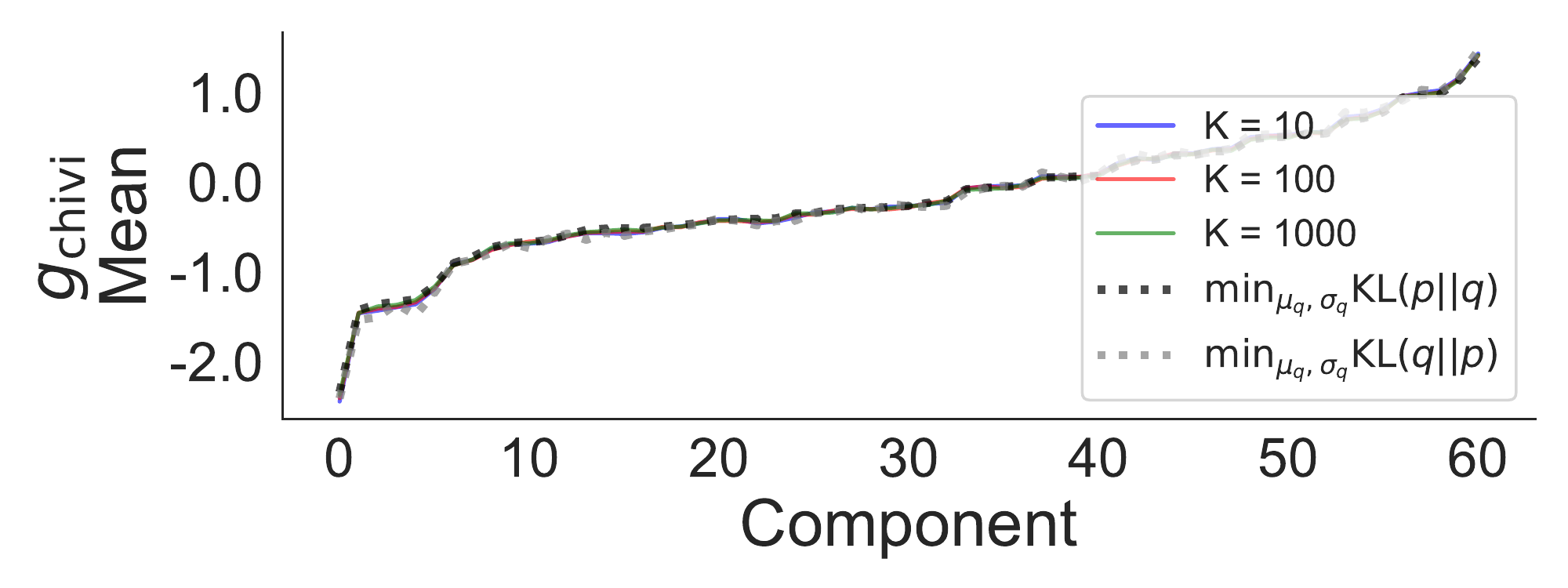}\hspace{0.25cm}
  \includegraphics[scale=0.35, trim = {0 2cm 0 0}, clip]{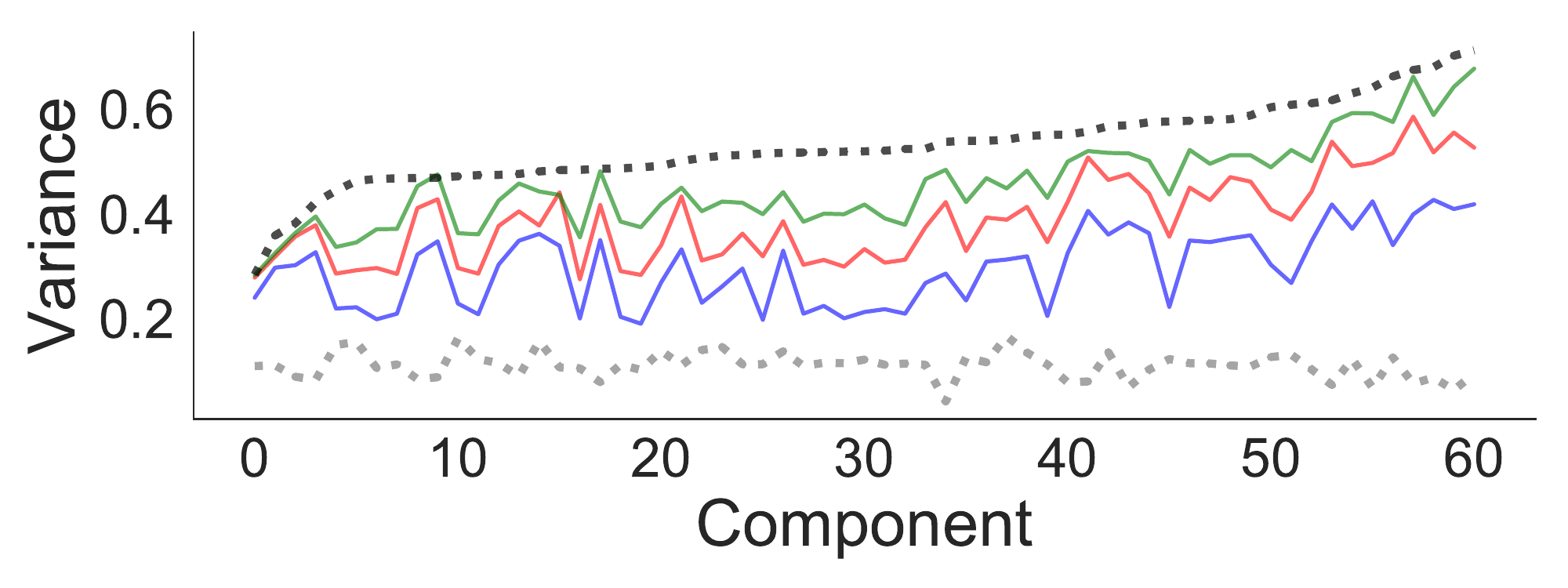}\vspace{0.1cm}
  
  \includegraphics[scale=0.35, trim = {0 0 0 0}, clip]{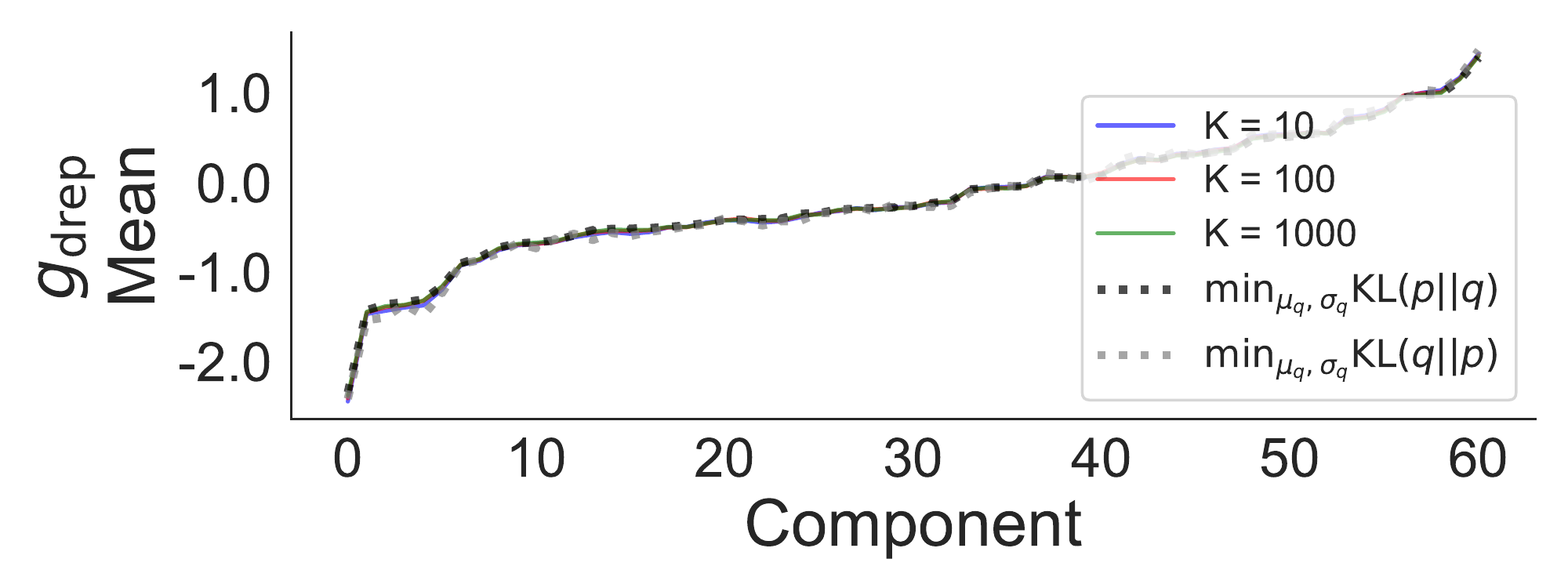}\hspace{0.25cm}
  \includegraphics[scale=0.35, trim = {0 0 0 0}, clip]{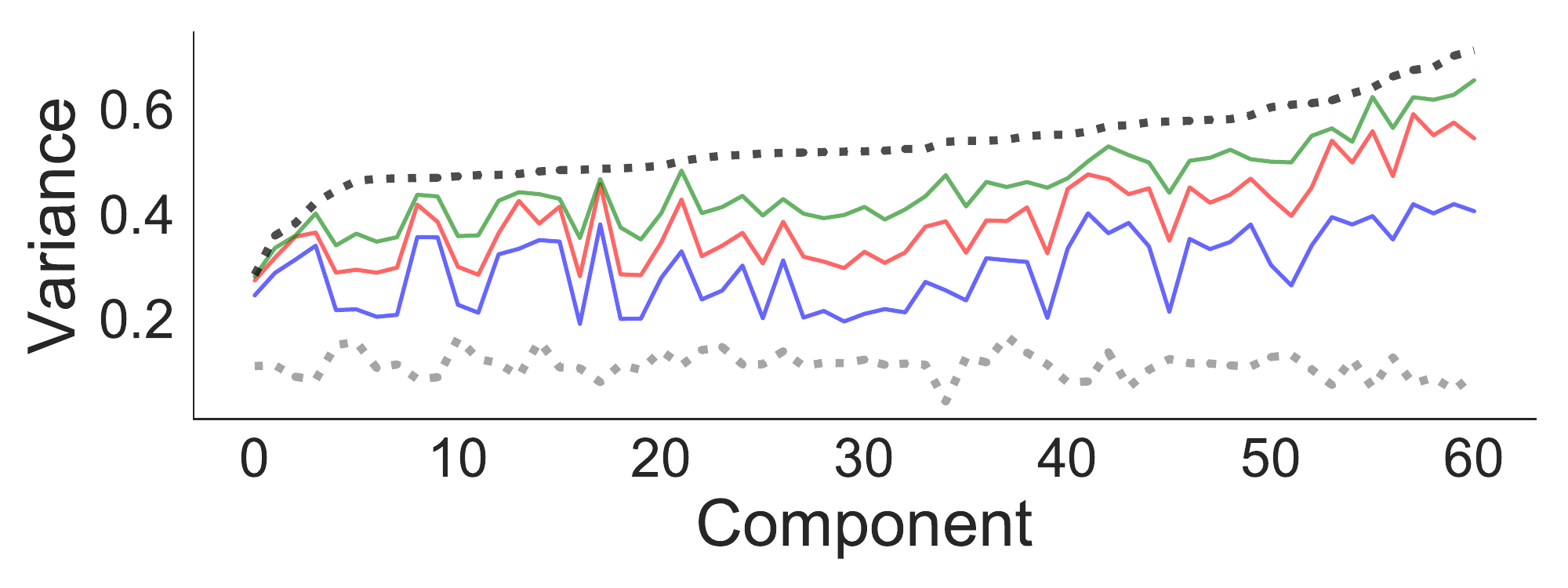}\vspace{0.1cm}
  }
\end{figure}

\begin{figure}[htbp]
\floatconts
  {fig:end_opt_a1a}
  {\caption{Optimization results for the \textit{a1a} dataset ($d=120$) with all estimators.}}
  {
  \includegraphics[scale=0.35, trim = {0 2cm 0 0}, clip]{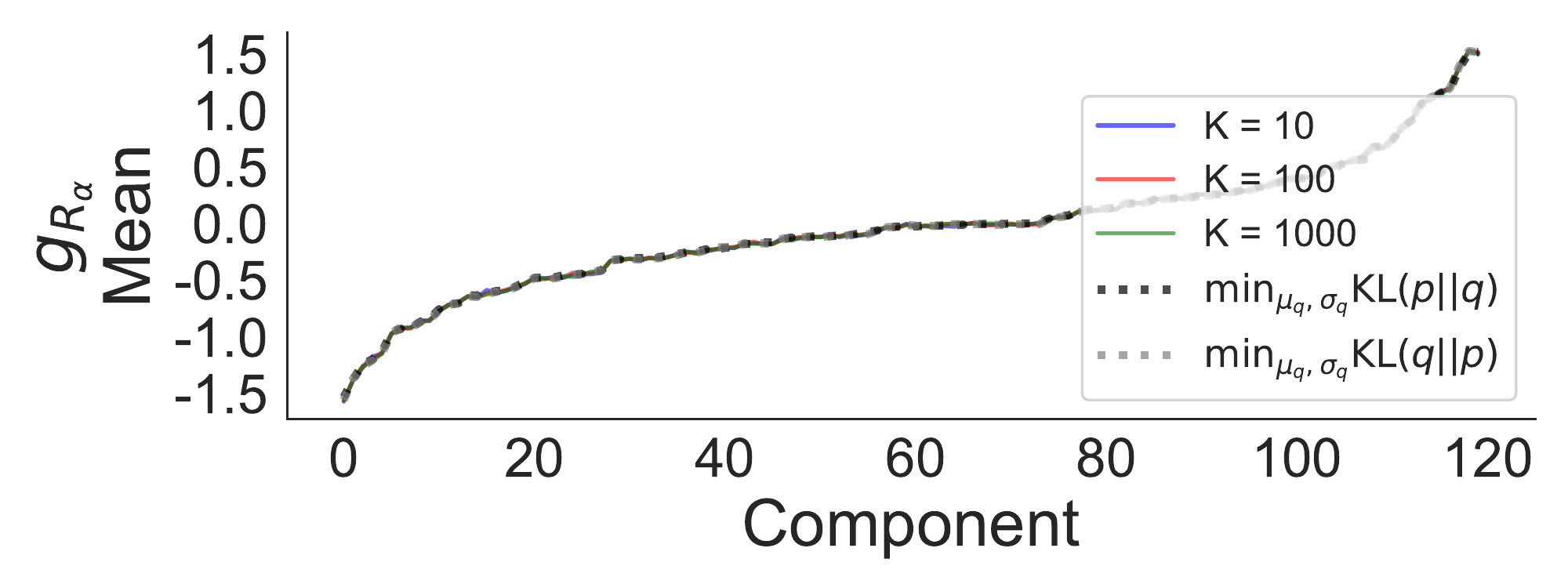}\hspace{0.25cm}
  \includegraphics[scale=0.35, trim = {0 2cm 0 0}, clip]{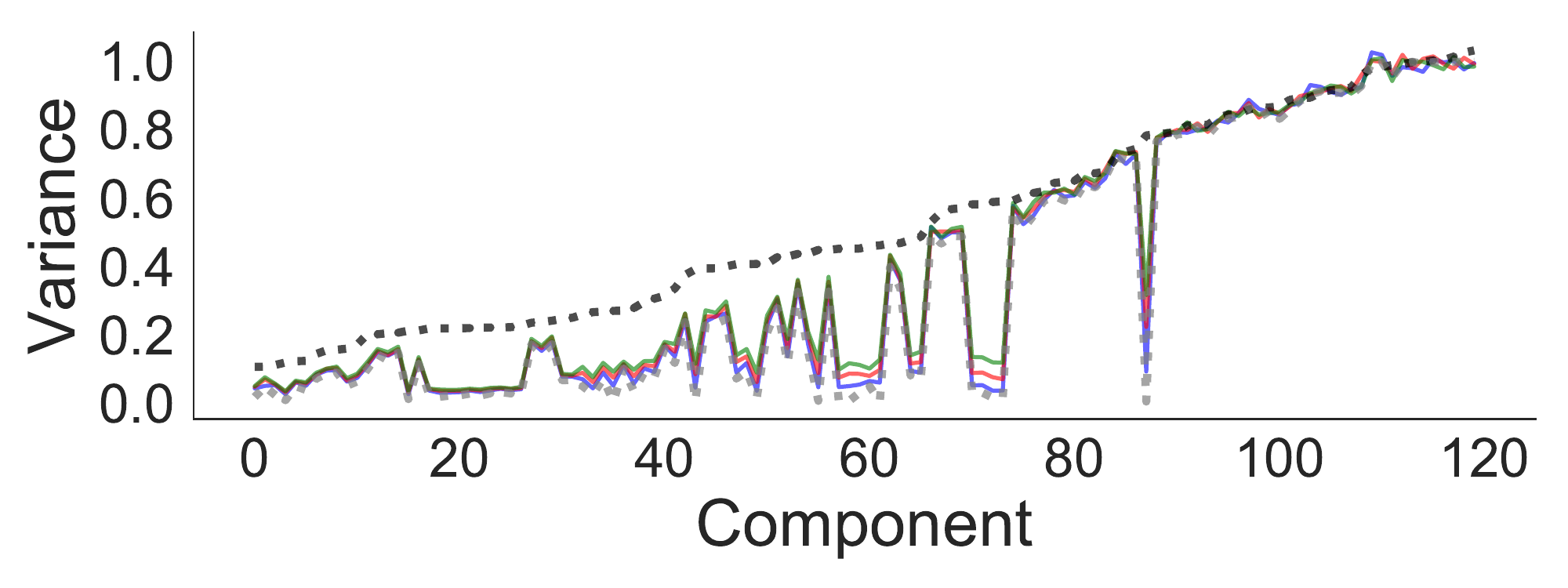}\vspace{0.1cm}
  
  \includegraphics[scale=0.35, trim = {0 2cm 0 0}, clip]{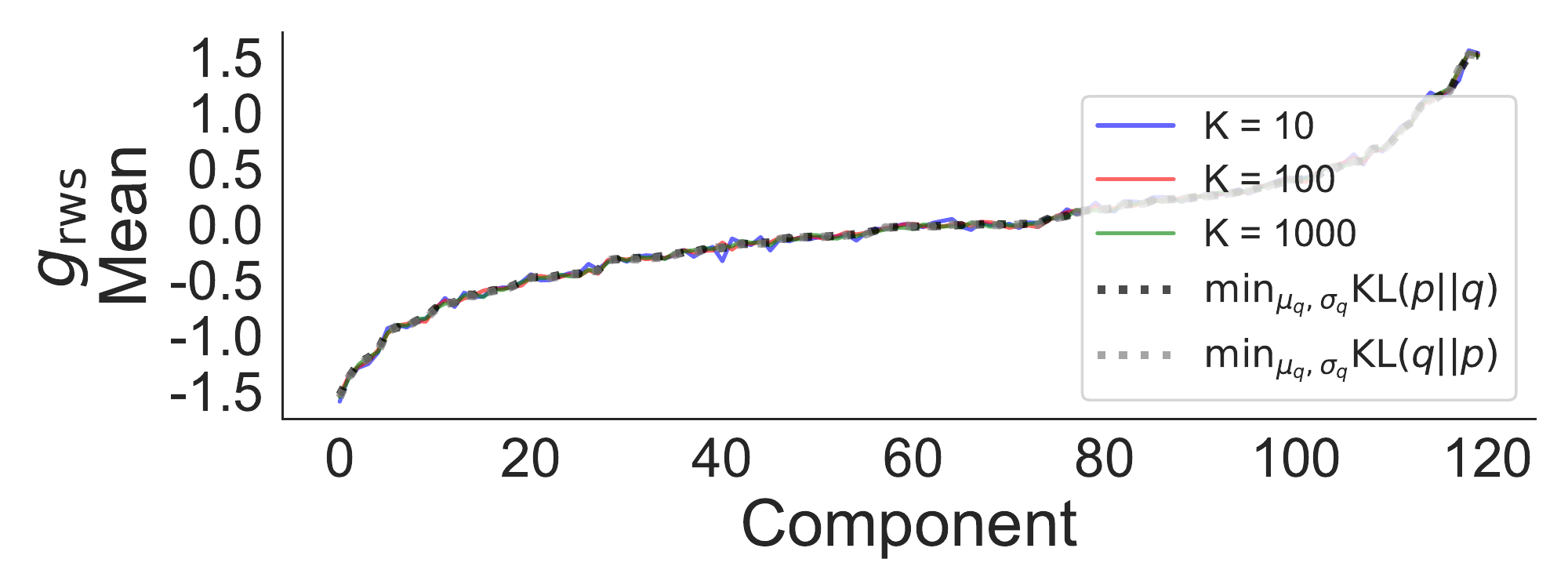}\hspace{0.25cm}
  \includegraphics[scale=0.35, trim = {0 2cm 0 0}, clip]{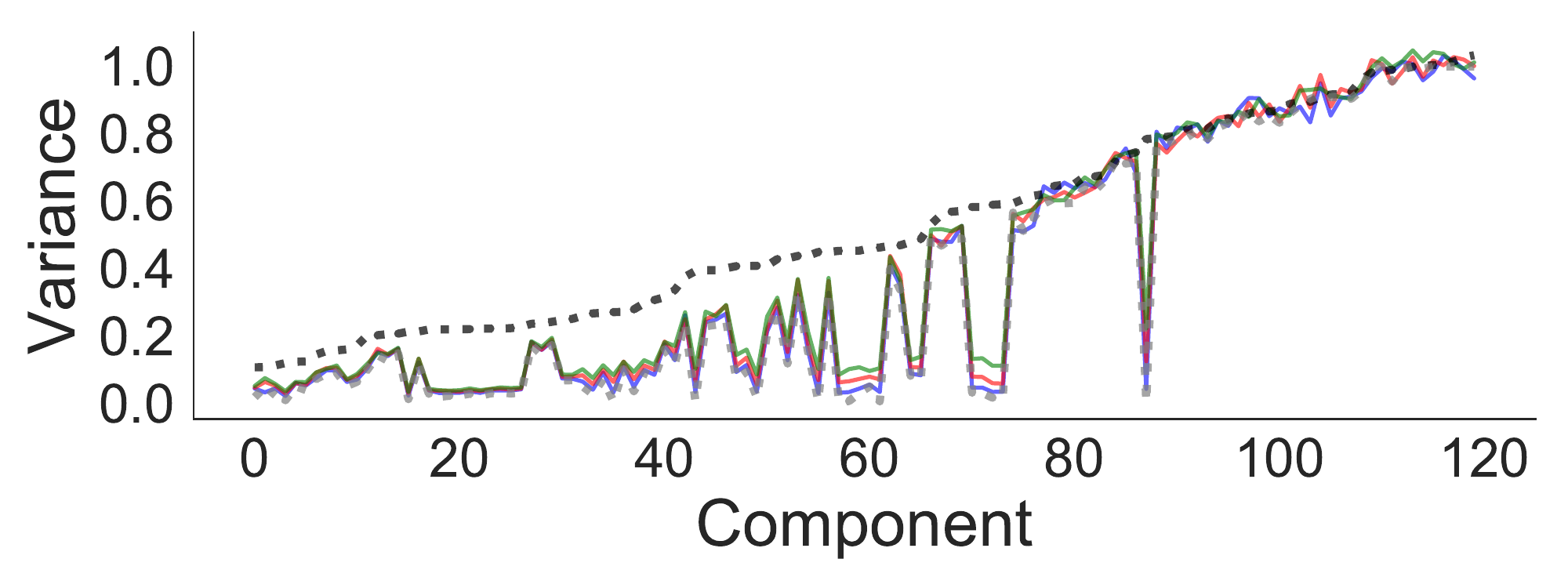}\vspace{0.1cm}
  
  \includegraphics[scale=0.35, trim = {0 2cm 0 0}, clip]{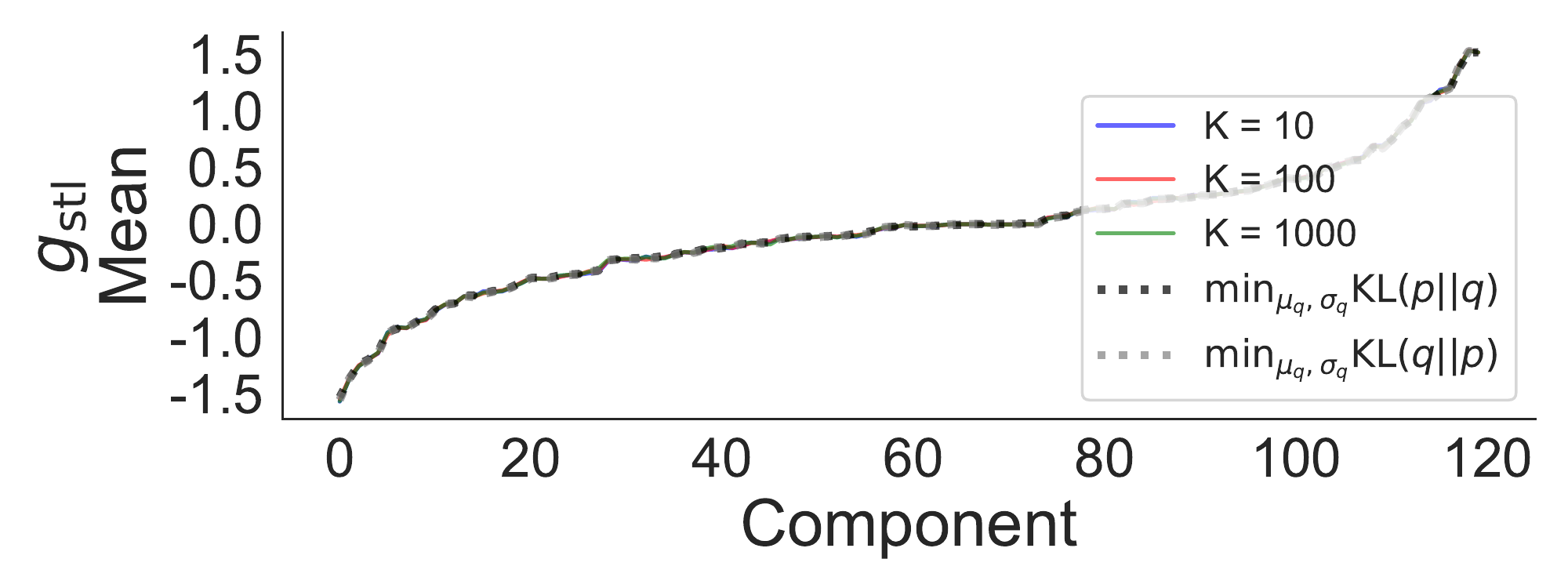}\hspace{0.25cm}
  \includegraphics[scale=0.35, trim = {0 2cm 0 0}, clip]{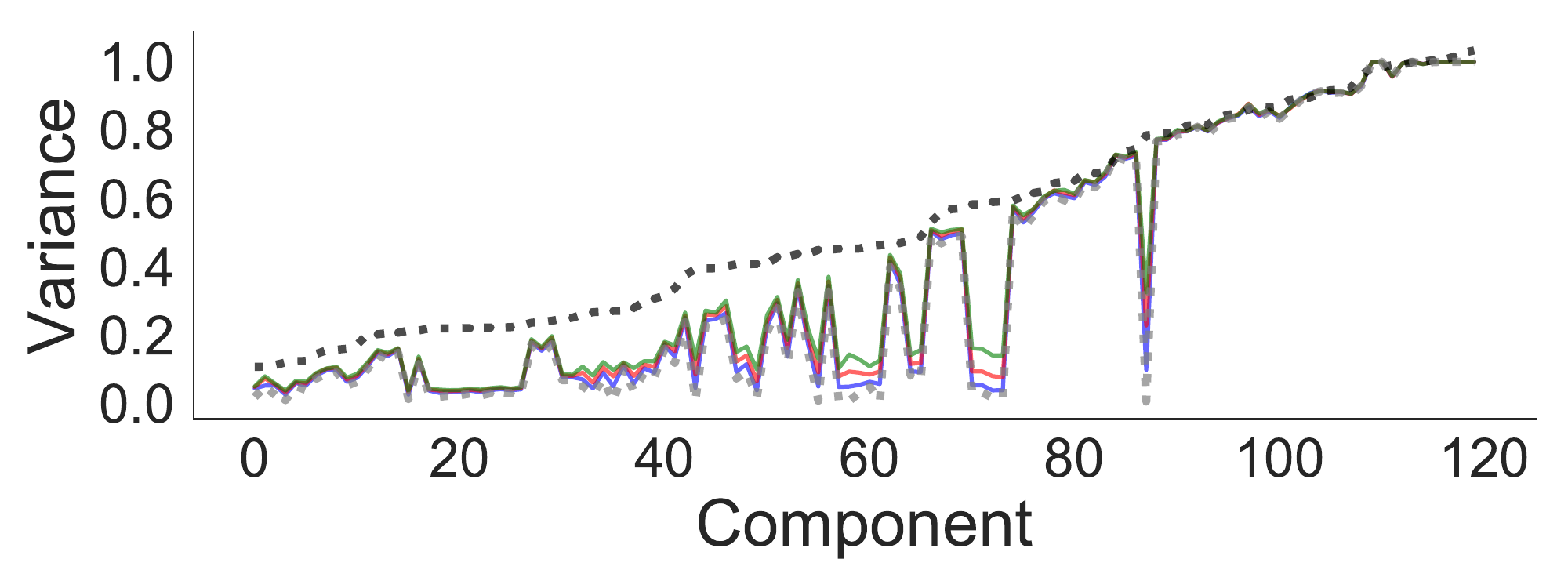}\vspace{0.1cm}
  
  \includegraphics[scale=0.35, trim = {0 2cm 0 0}, clip]{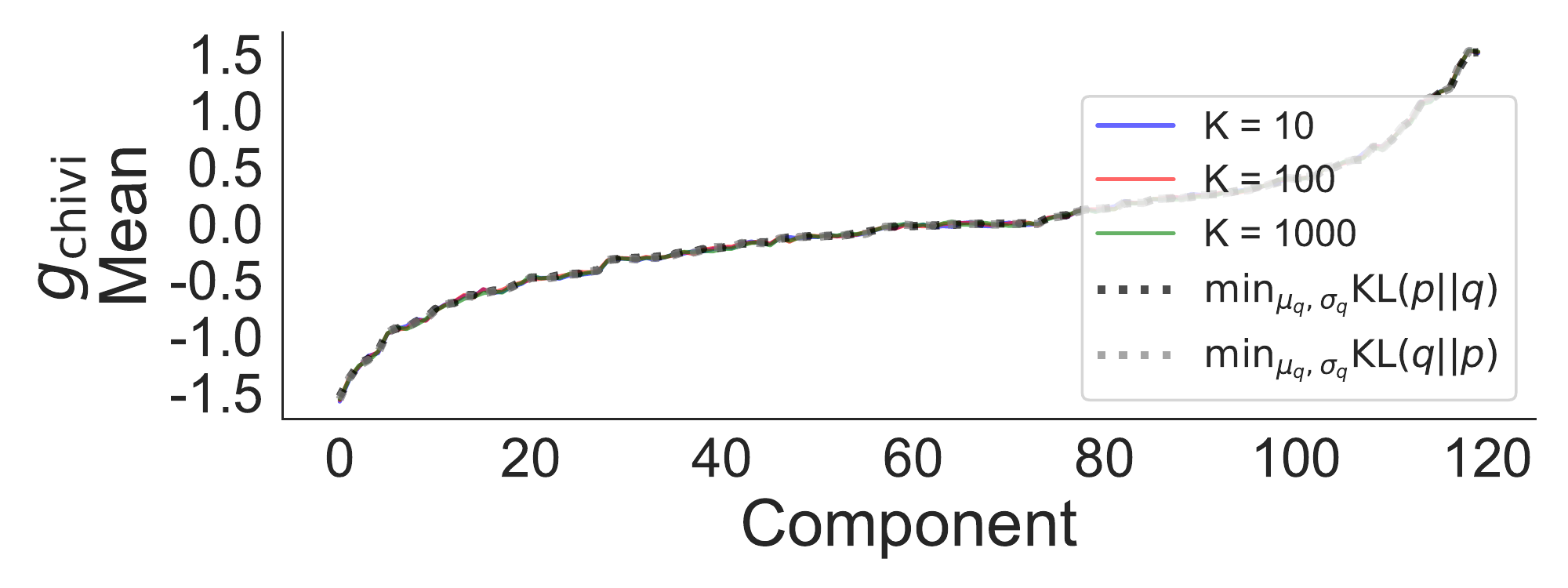}\hspace{0.25cm}
  \includegraphics[scale=0.35, trim = {0 2cm 0 0}, clip]{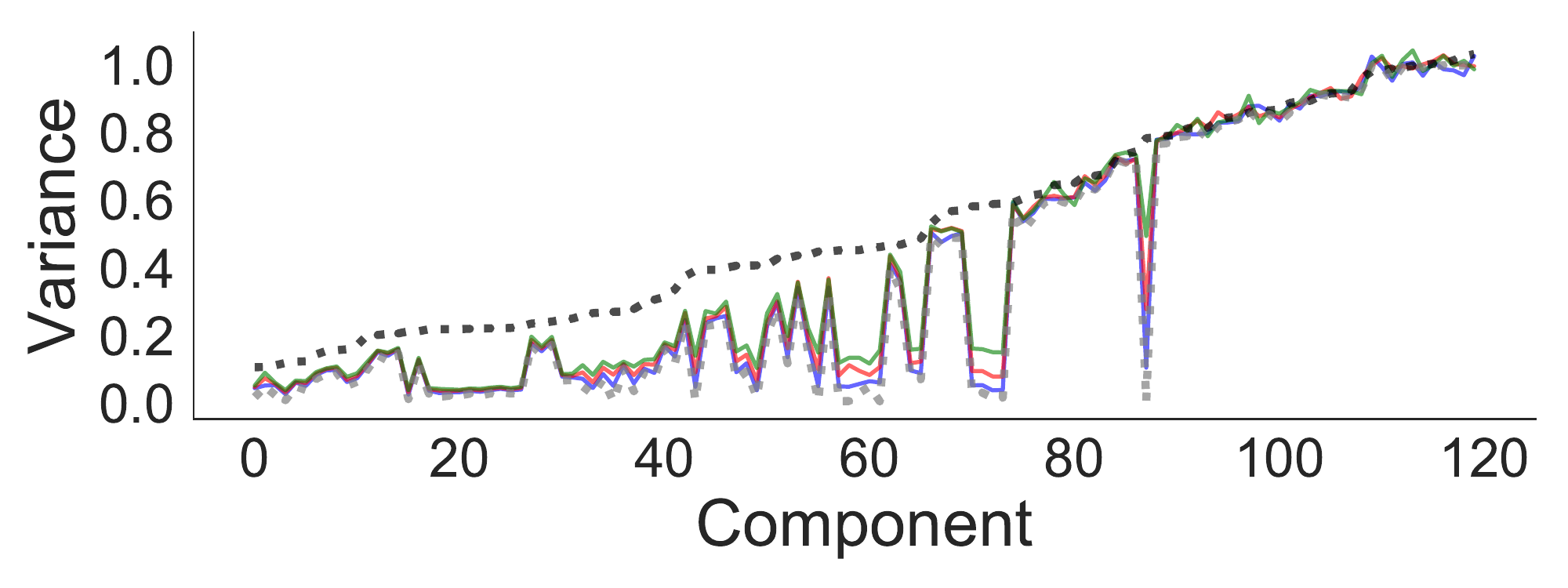}\vspace{0.1cm}
  
  \includegraphics[scale=0.35, trim = {0 0 0 0}, clip]{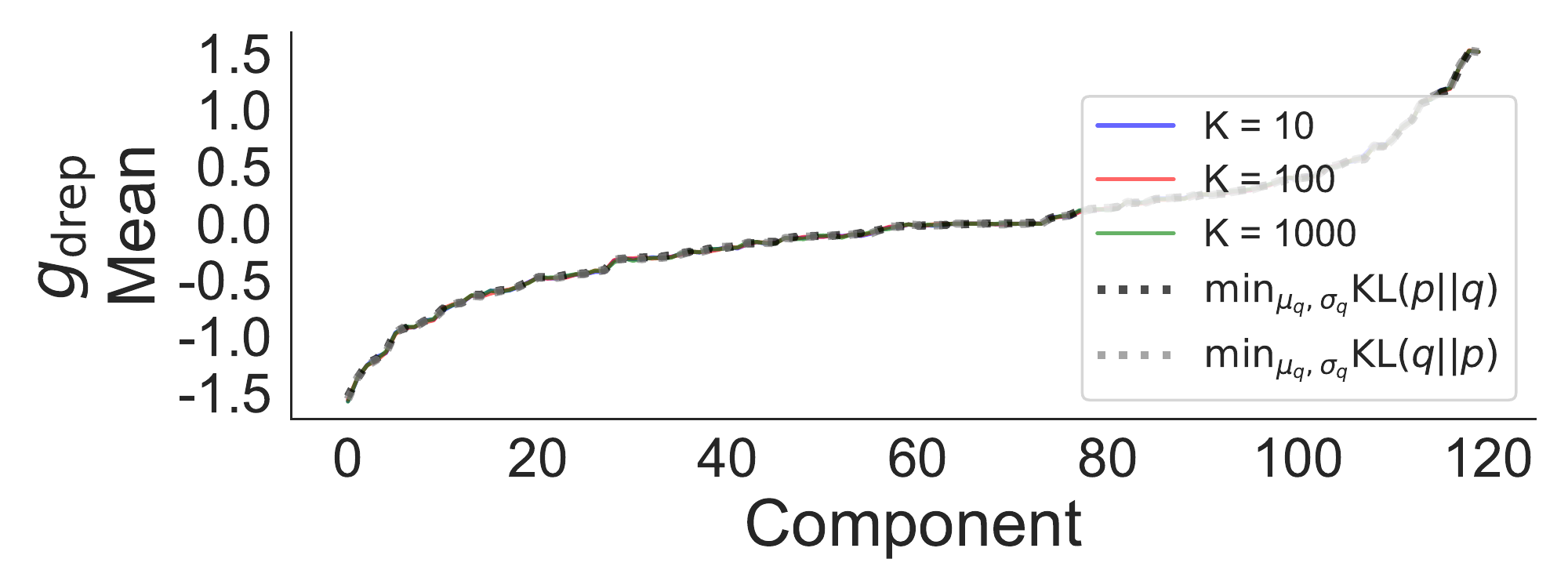}\hspace{0.25cm}
  \includegraphics[scale=0.35, trim = {0 0 0 0}, clip]{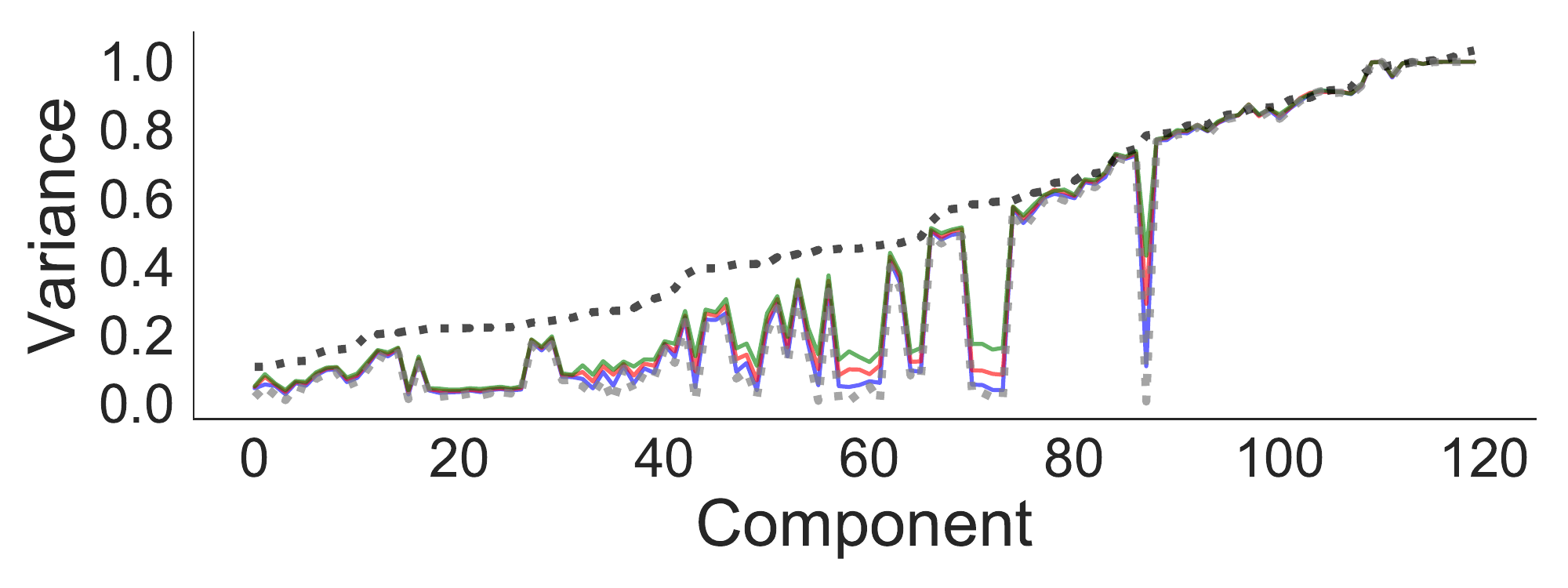}\vspace{0.1cm}
  }
\end{figure}

\clearpage
\newpage

\section{Weight collapse in self normalized importance sampling}\label{app:3}

This section shows the weight collapse (also known as weight degeneracy) effect of self normalized importance sampling. Simply put, the degeneracy of the normalized importance weights refers to the scenario where only a small number of samples have significant importance weight, and thus completely dominate the value of the approximations. This is known to be an inefficiency of self normalized importance sampling, since most of the samples have almost no contribution at all in the value of the estimates \citep{bengtsson2008curse}.

We consider the same setting as the one in Section \ref{sec:gaussian}. We set $p$ to be a diagonal $d$-dimensional Gaussian with mean zero and variances $\sigma_{p_i}^2 = 0.2 + 9.8 \frac{i}{d}$ ($i = 1,...,d$), and we set $q$ to be an isotropic Gaussian with mean zero and covariance $\sigma_q^2 I$, with $\sigma_q^2 = 9$ (its value at initialization). The normalized importance weights are computed as 
\[\tilde w_k = \frac{\frac{p(x,z_k)}{q(z_k)}}{\sum_{j=1}^K \frac{p(x,z_j)}{q(z_j)}} \qquad \mbox{where} \qquad z_k\sim q.\]

We compute the normalized importance weights for several dimensions $d$ and number of samples $K$. Fig.~\ref{fig:wc} shows the results. Specifically, it shows the values of the ten largest normalized weights for all the pairs $(d, K)$ considered. It can be observed that, for a dimensionality $d\geq 100$, almost all of the mass is concentrated in the largest two weights, regardless of the value of $K$ used (i.e. the weights ``collapse''). This is exactly aligned with the failures observed in Section \ref{sec:gaussian}.

\begin{figure}[htbp]
\floatconts
  {fig:wc}
  {\caption{Visualization of the weight collapse effect of self normalized importance sampling.}}
  {
  \includegraphics[scale=0.3, trim = {0 2cm 0 0}, clip]{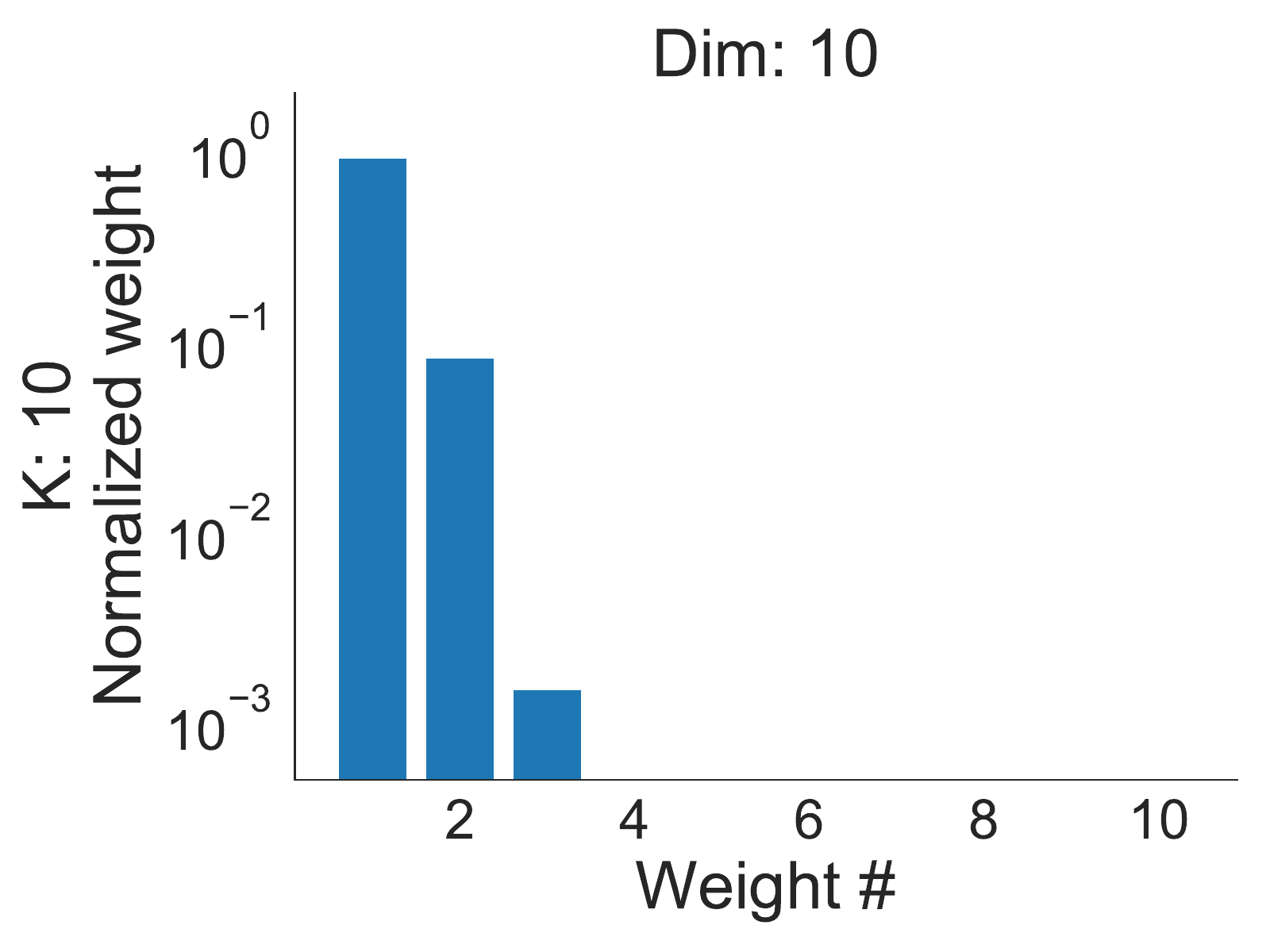}\hspace{0.5cm}
  \includegraphics[scale=0.3, trim = {3.5cm 2cm 0 0}, clip]{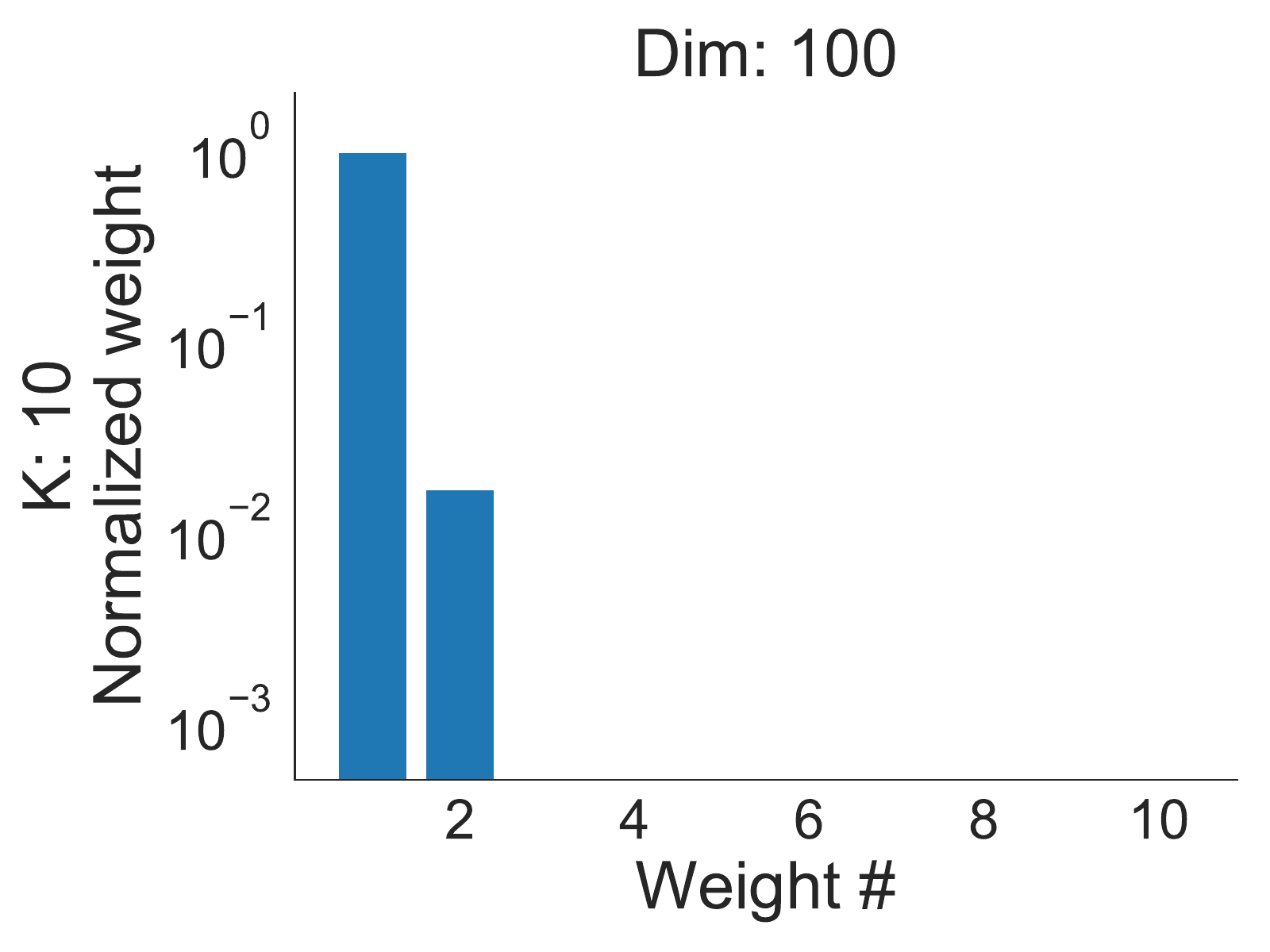}\hspace{0.5cm}
  \includegraphics[scale=0.3, trim = {3.5cm 2cm 0 0}, clip]{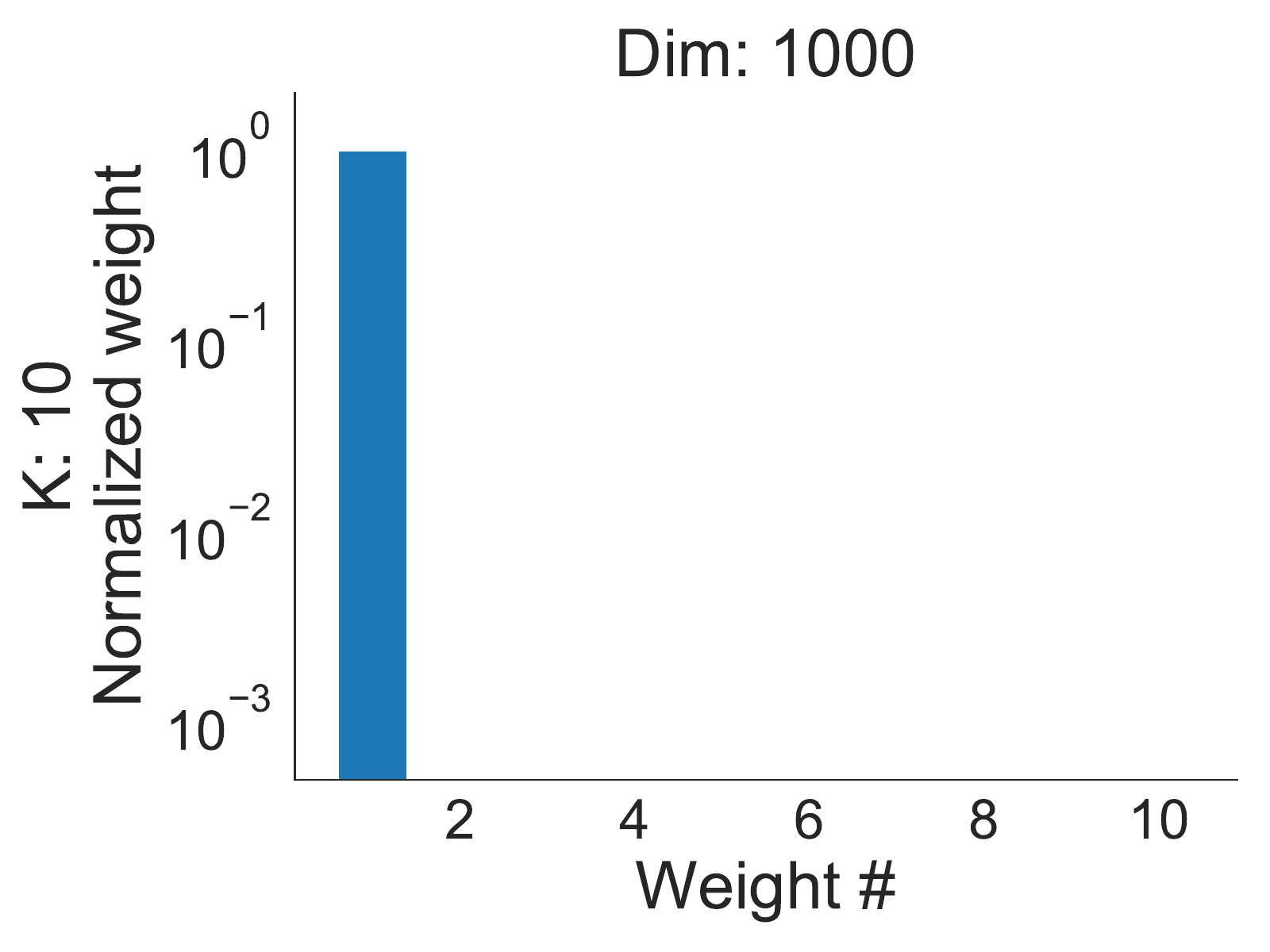}
  
  \includegraphics[scale=0.3, trim = {0 2cm 0 1.0cm}, clip]{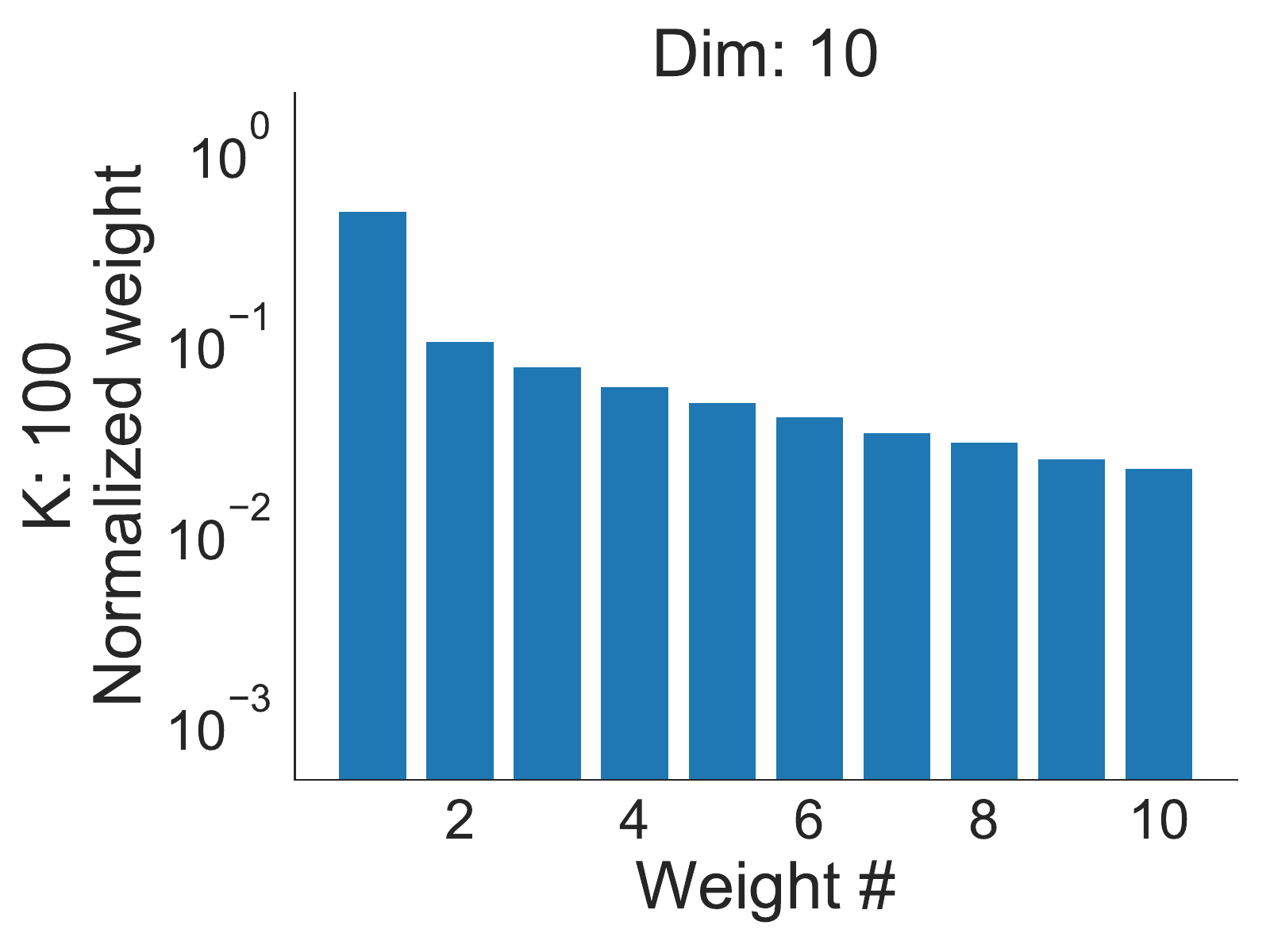}\hspace{0.5cm}
  \includegraphics[scale=0.3, trim = {3.5cm 2cm 0 1.0cm}, clip]{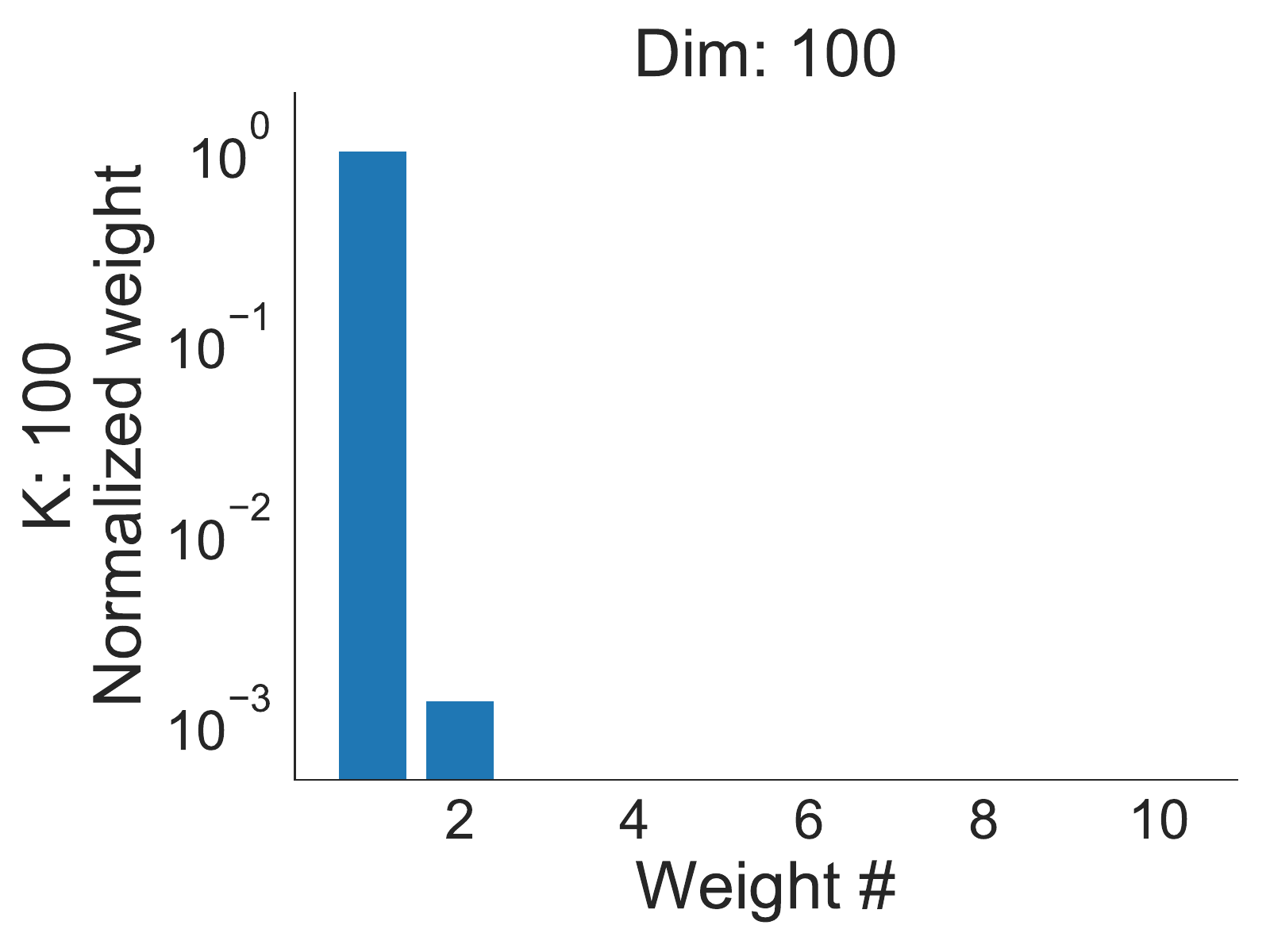}\hspace{0.5cm}
  \includegraphics[scale=0.3, trim = {3.5cm 2cm 0 1.0cm}, clip]{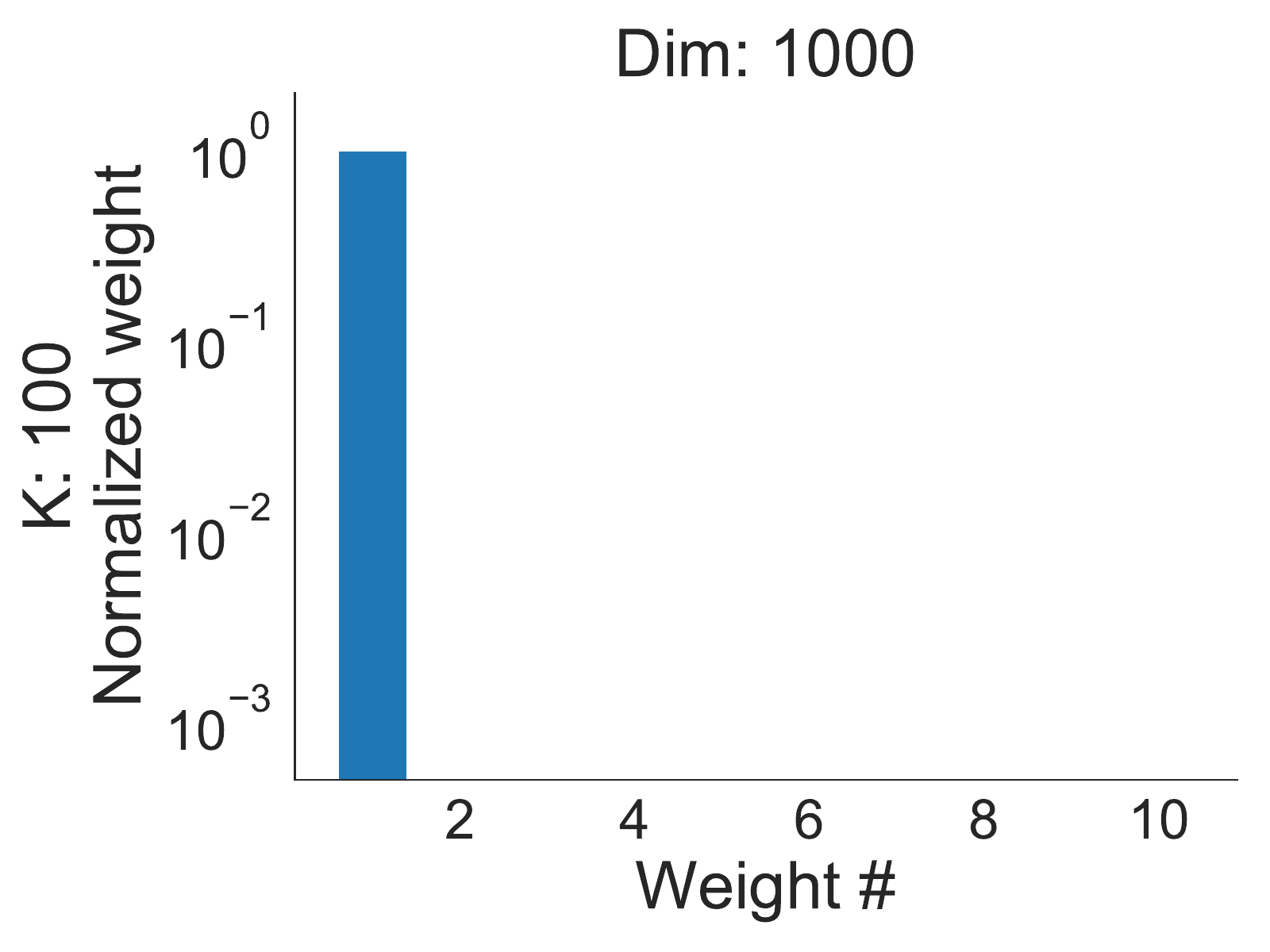}
  
  \includegraphics[scale=0.3, trim = {0 2cm 0 1.0cm}, clip]{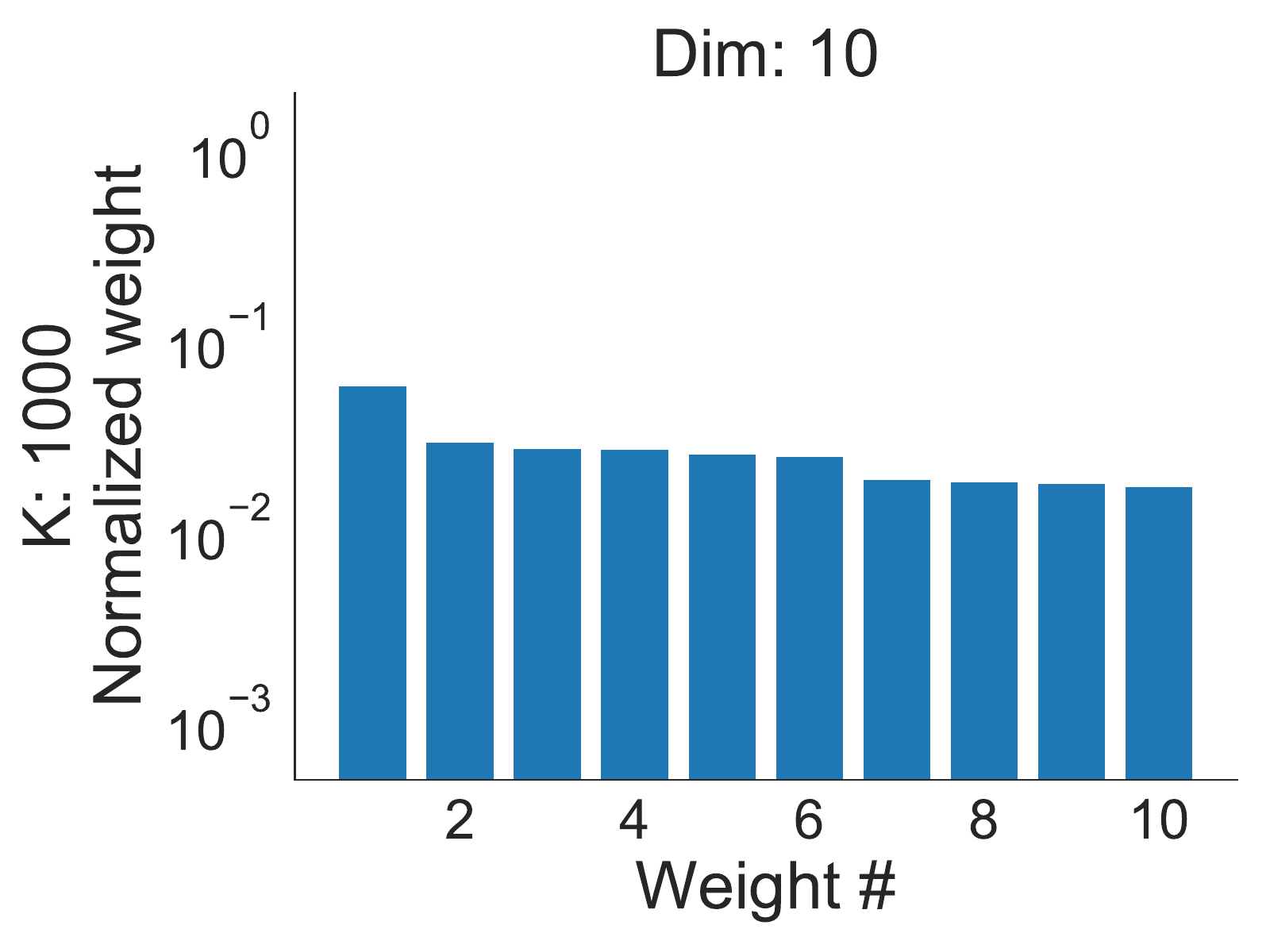}\hspace{0.5cm}
  \includegraphics[scale=0.3, trim = {3.5cm 2cm 0 1.0cm}, clip]{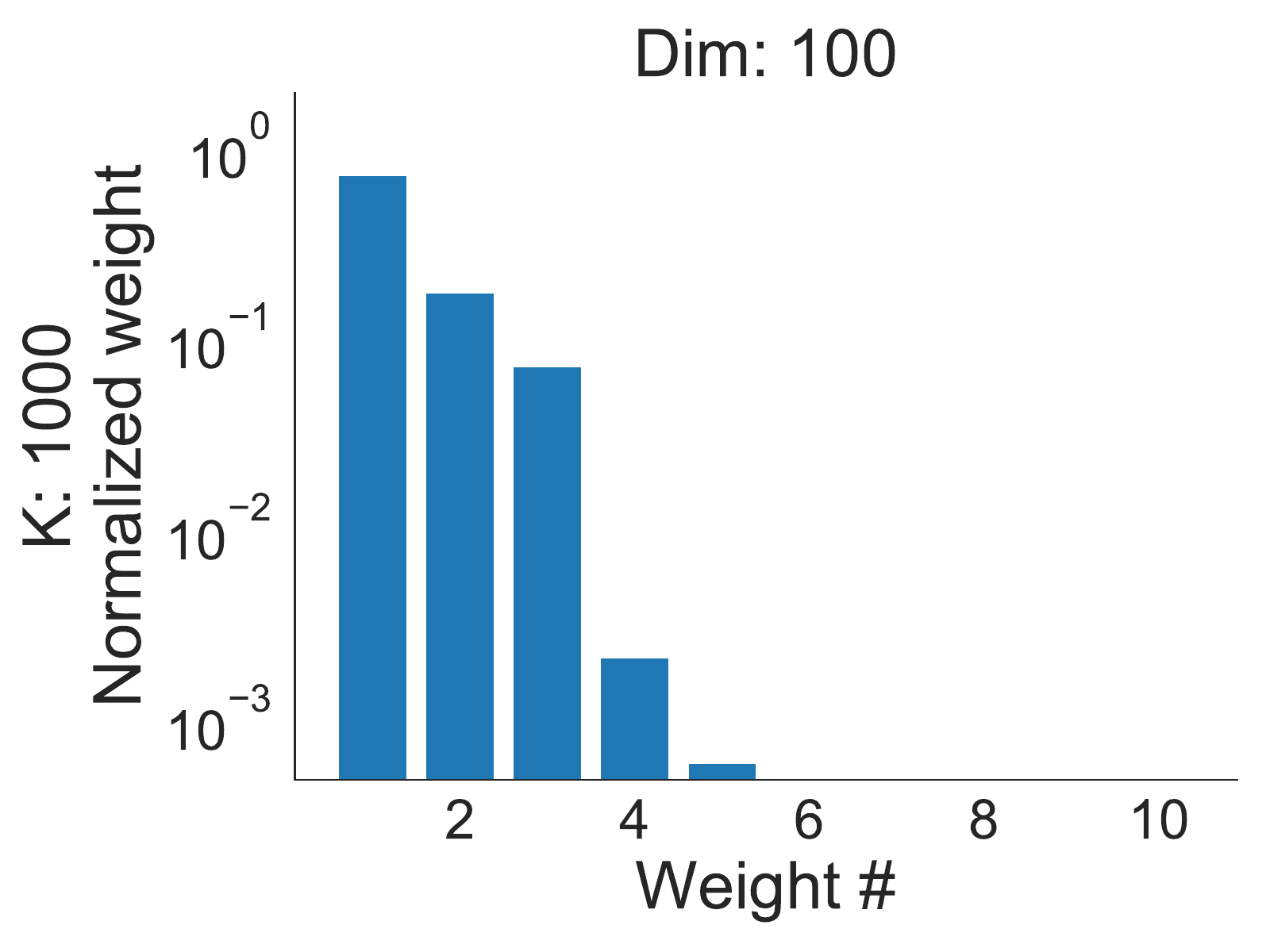}\hspace{0.5cm}
  \includegraphics[scale=0.3, trim = {3.5cm 2cm 0 1.0cm}, clip]{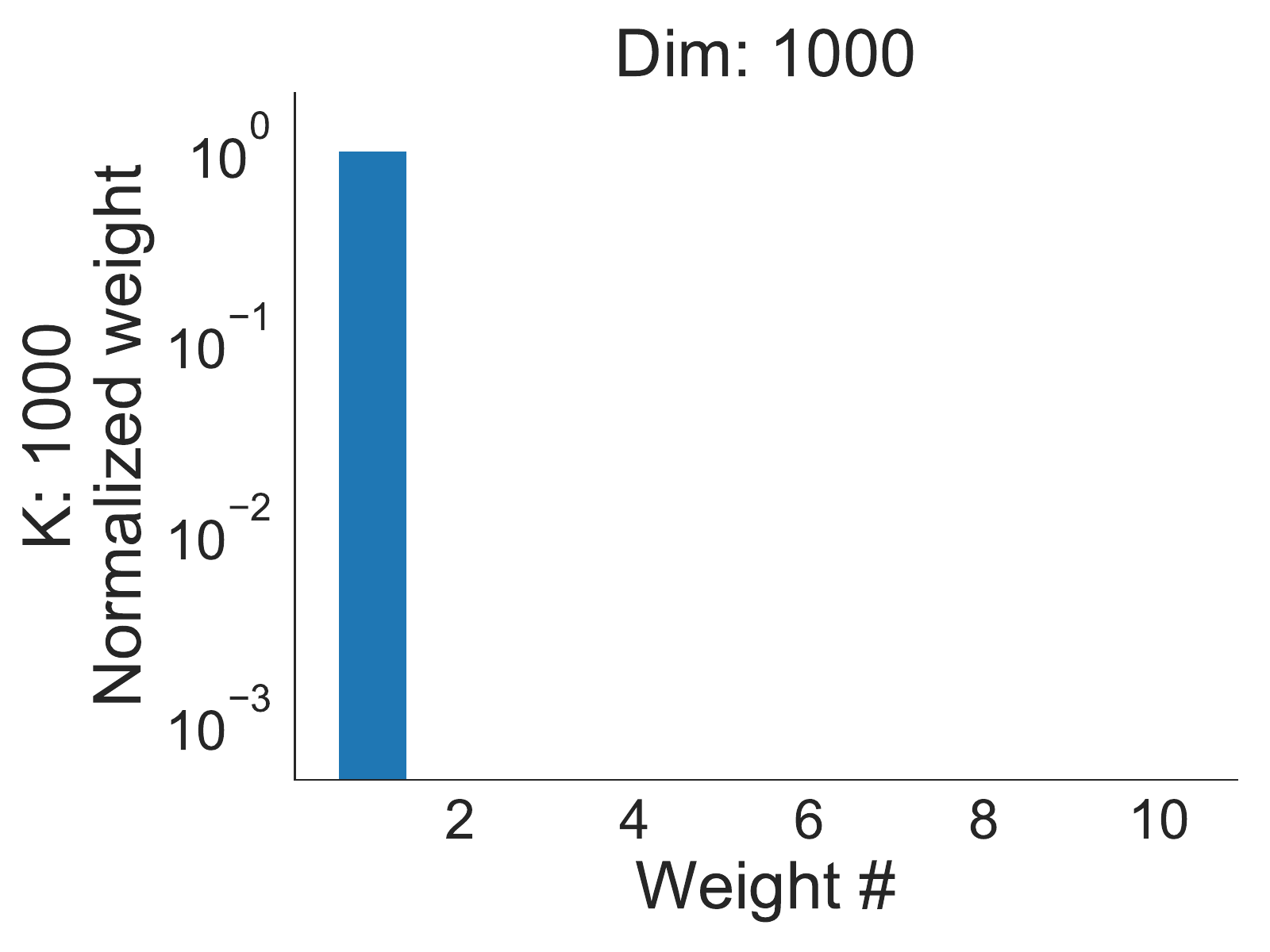}
  
  \includegraphics[scale=0.3, trim = {0 0 0 1.0cm}, clip]{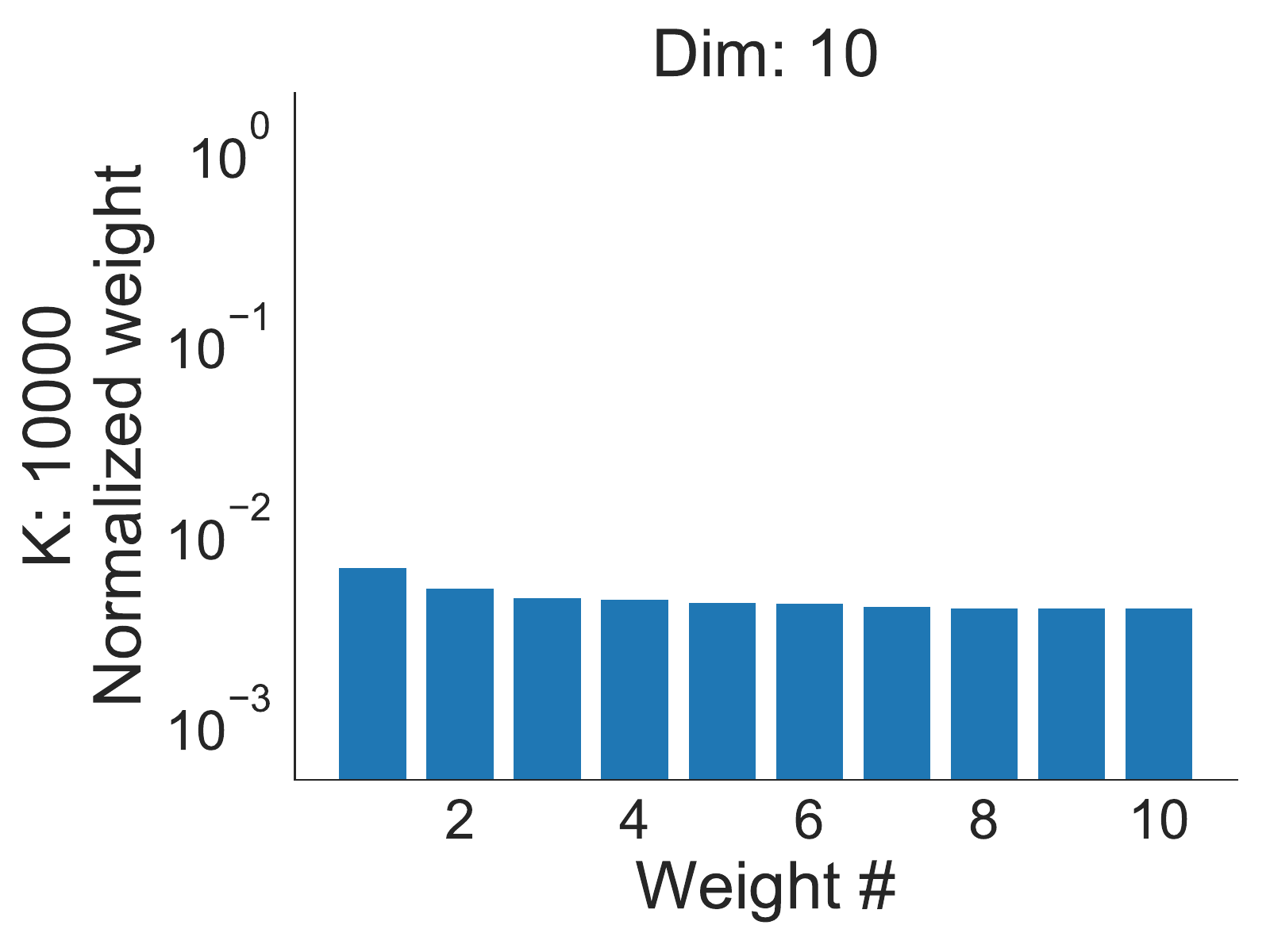}\hspace{0.5cm}
  \includegraphics[scale=0.3, trim = {3.5cm 0 0 1.0cm}, clip]{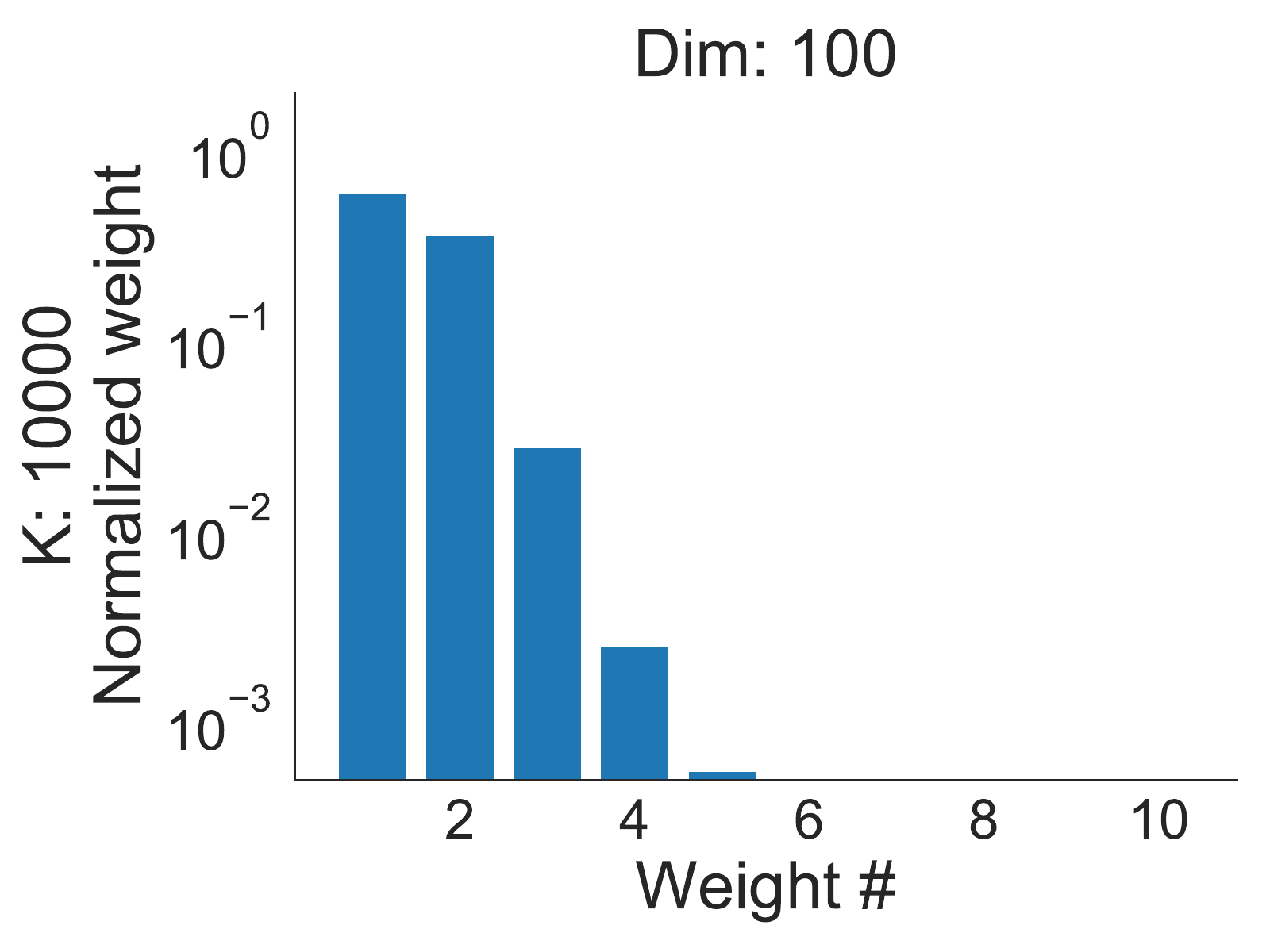}\hspace{0.5cm}
  \includegraphics[scale=0.3, trim = {3.5cm 0 0 1.0cm}, clip]{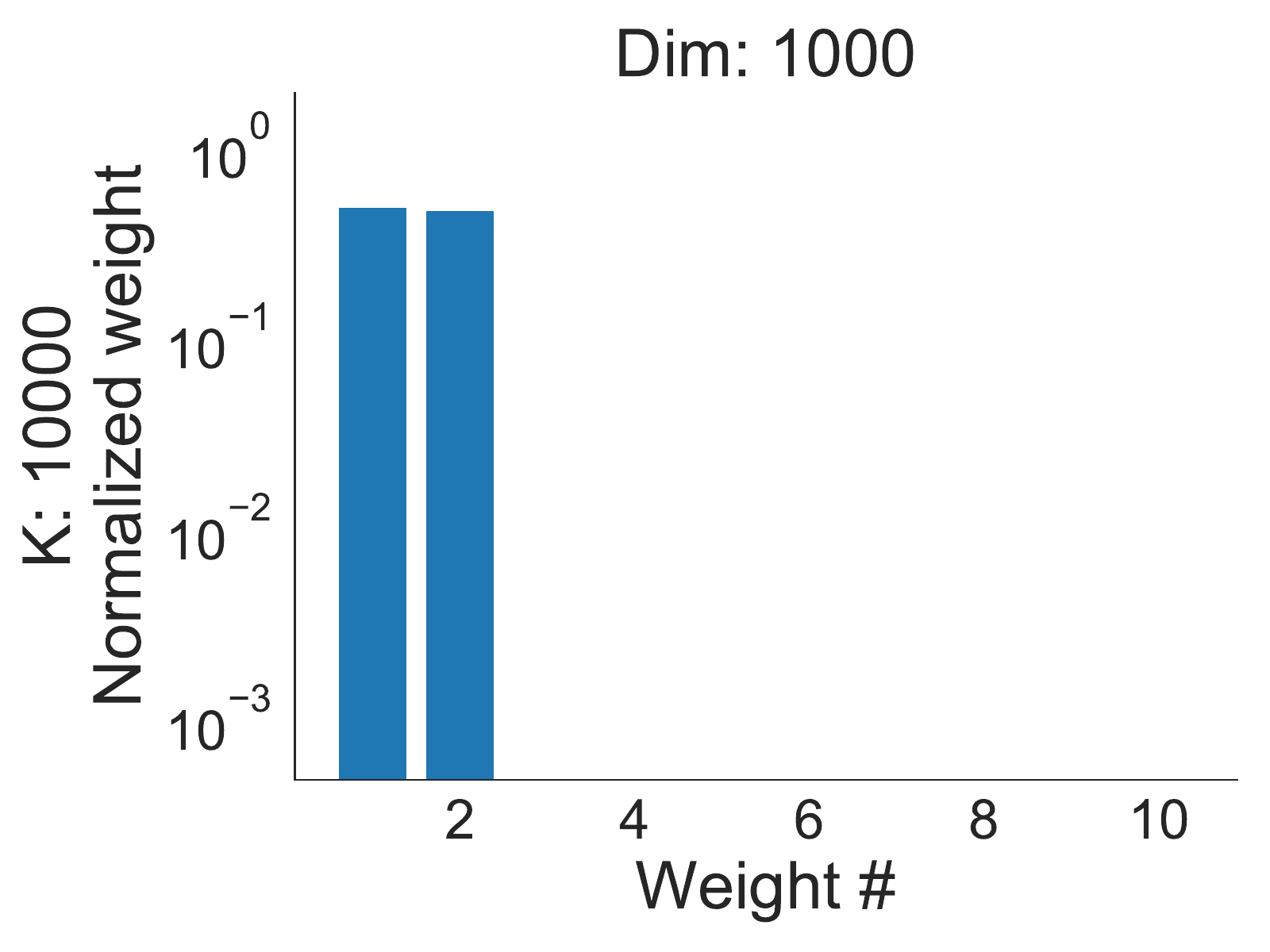}
  }
\end{figure}

\end{document}